\pdfoutput=1
\documentclass[10pt,twocolumn,letterpaper]{article}
\usepackage[pagenumbers]{cvpr}
\usepackage[utf8]{inputenc}
\usepackage[T1]{fontenc}
\usepackage{amsfonts}
\usepackage{amsthm}
\usepackage{multirow}
\usepackage{nicefrac}
\usepackage{microtype}
\usepackage{arydshln}
\usepackage{algorithm}
\usepackage{algorithmic}
\usepackage{needspace}

\newtheorem{corollary}{Corollary}
\newtheorem{lemma}{Lemma}
\newtheorem{proposition}{Proposition}
\newtheorem{theorem}{Theorem}

\newtheorem{assumption}{Assumption}
\newtheorem*{theorem*}{Theorem}
\newtheorem*{proposition*}{Proposition}

\newcommand{\pbar}{\bar{\mathbf{p}}}
\newcommand{\pdag}{\mathbf{p}^\dagger}
\newcommand{\bpi}{\boldsymbol{\pi}}
\newcommand{\KL}{\mathrm{KL}}
\newcommand{\TV}{\mathrm{TV}}
\newcommand{\Ibatch}{I_{\mathrm{batch}}}

\definecolor{cvprblue}{rgb}{0.21,0.49,0.74}
\usepackage[pagebackref,breaklinks,colorlinks,allcolors=cvprblue]{hyperref}

\title{CAMA: Calibrated Anti-Collapse Marginal Alignment for\\Test-Time Adaptation of Vision--Language Models}
\author{Jinho Kang \qquad Yunhui Guo\thanks{Corresponding author.}\\
The University of Texas at Dallas\\
{\tt\small \{jinho.kang, yunhui.guo\}@utdallas.edu}
}

\begin{document}
\maketitle
\begin{abstract}
Test-time adaptation (TTA) updates a model using unlabeled test data to help it handle distribution shifts. However, prior work has found that image corruption causes a vision--language model (VLM) to produce more similar features for different images, a phenomenon known as feature compression. Under compression, entropy minimization, the most common TTA objective, can cause the model to favor a few classes and perform worse than before adaptation. Despite these observations, a detailed analysis of how these failures are connected and an effective strategy for preventing them are still lacking.
In this paper, we first analyze why entropy minimization fails for VLMs. We then propose CAMA, a new test-time adaptation objective that maximizes batch mutual information to encourage confident predictions for individual images while maintaining diverse predictions across the batch. CAMA also estimates a reference class distribution early in the test stream and aligns each batch's average prediction with this reference during adaptation. We prove that feature compression biases the predictions and that entropy minimization amplifies this bias into collapse, whereas CAMA is guaranteed to preserve every class. Across standard corruption benchmarks, we observe collapse on nearly every VLM backbone evaluated. CAMA prevents collapse in every observed case and, in turn, achieves the best accuracy on every backbone. Quantitative and qualitative analyses confirm that CAMA reverses the feature compression behind the collapse, and ablations show that each of its components contributes.
\end{abstract}

\vspace{-1.5em}
\section{Introduction}

Test-time adaptation (TTA), which updates a pre-trained model on test data, faces two main challenges, the lack of labels and the shift between training and test distributions. Because entropy minimization ~\citep{wang2020tent} addresses both, it is a key component in many widely used TTA methods~\citep{wang2020tent,niu2023towards,shu2022test}. Entropy minimization uses only the model's own predictions, so it needs no labels, and by sharpening these predictions it restores the confidence that the shift lowers~\citep{grandvalet2004semi,wang2020tent}. Due to its effectiveness, entropy minimization has been widely applied for improving model performance in face of distribution shifts in application of image classification~\citep{wang2020tent,niu2022efficient,niu2023towards}, semantic segmentation~\citep{wang2020tent} and object detection~\citep{belal2026vlod}.

\begin{figure}[t]
\centering
\includegraphics[width=\linewidth]{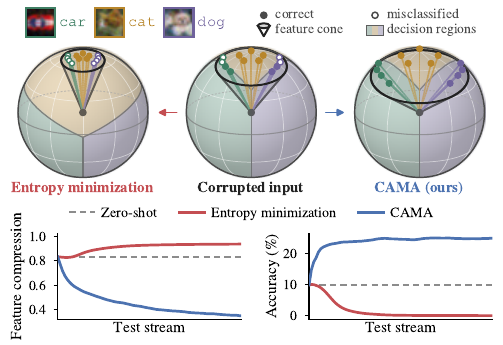}
\caption{\textbf{Corruption compresses the image features, entropy minimization grows the resulting bias into collapse, and CAMA reverses both}. Top, a schematic where the cone spans the features. It is compressed by corruption and further by entropy minimization, until every feature is predicted as \texttt{cat}, and expands again under CAMA. Bottom, feature compression, the mean pairwise cosine of the image features, and accuracy over an ImageNet-C test stream.}
\vspace{-1.5em}
\label{fig:teaser}
\end{figure}

In recent years, vision--language models (VLMs) such as CLIP~\citep{radford2021learning} have emerged as powerful models with rich knowledge acquired through image--text pre-training~\citep{radford2021learning,zhang2024vision}. However, VLMs also suffer from performance degradation under distribution shifts~\citep{shu2022test}. To address this, entropy minimization is increasingly being extended to TTA for VLMs~\citep{shu2022test,li2025generalizing}.

However, when entropy minimization is applied directly to VLMs, limitations arise in both their predictions and image features. First, ClipTTA~\citep{lafon2025cliptta} found that under image corruption~\citep{hendrycks2019benchmarking}, a standard evaluation setting for TTA, entropy minimization concentrates CLIP's predictions on a few classes, a failure mode we refer to as \emph{prediction collapse}. Second, Mint~\citep{bao2025mint} found that under image corruption, CLIP's image features, which already lie in a narrow cone like those of other contrastive VLMs~\citep{liang2022mind,mistretta2025cross}, undergo \emph{feature compression}, moving even closer together. However, each observation captures only one aspect of the failure.

Our analysis reveals that feature compression further drives prediction collapse. Corruption first compresses the image features, biasing predictions toward the same few classes. Entropy minimization then reinforces this bias until the predictions collapse. We refer to this process as the \emph{compression--collapse phenomenon}, illustrated in Figure~\ref{fig:teaser}. Addressing this failure therefore requires correcting the prediction bias introduced by feature compression while preserving the adaptation benefits of entropy minimization without allowing it to amplify that bias.

To this end, we propose \textbf{CAMA}, a new test-time adaptation objective for VLMs. Rather than maximizing prediction confidence alone, as entropy minimization does, CAMA maximizes batch mutual information to encourage confident predictions for individual images while maintaining diverse predictions across the batch. This objective preserves the adaptation benefits of entropy minimization while discouraging predictions from collapsing to the same class. In addition, CAMA corrects the prediction bias introduced by feature compression by aligning its predictions with a reference class distribution estimated with reduced influence from this bias. This alignment further helps prevent prediction collapse (Figure~\ref{fig:teaser}). Finally, if confidence maximization dominates, it amplifies bias and leads to collapse; if alignment dominates, it hinders adaptation. CAMA therefore balances the terms so that their gradients have comparable magnitudes, preventing either from dominating the update.

Theoretically, we establish three results. (i) When the image features lie close to one direction, we derive a lower bound on the average probability assigned to a single class. This bound tightens as the features become more compressed. (ii) The lower the loss of entropy minimization, the larger this class's share must be, and it converges to one as the loss is minimized. (iii) In the limit of strong compression, where all predictions coincide, every optimum of entropy minimization assigns all probability to a single class, whereas CAMA's objective has a unique optimum that assigns positive probability to every class.

Empirically, across corruptions, stronger compression correlates with a larger gain of CAMA. Visualizations further show that corruption compresses the class clusters together, whereas CAMA separates them again. Countering this phenomenon also improves model accuracy under distribution shifts. On seven contrastive VLMs that differ in training objective, data, recipe, and architecture, and on three corruption benchmarks, CAMA is the best method in all but one of the 21 backbone--benchmark pairs, by up to 6.5\% in accuracy over the strongest baseline. It remains the best on continual and mixed streams, and ablations show that each component contributes.

Our contributions are as follows.
\begin{itemize}[leftmargin=1.1em,itemsep=1pt,topsep=2pt,parsep=0pt]
    \item We prove, and verify empirically, that feature compression biases predictions and entropy minimization amplifies this bias into collapse.
    \item We propose CAMA, which keeps the adaptation power of entropy minimization with a theoretical guarantee against collapse.
    \item We show that CAMA prevents the collapse wherever entropy minimization suffers it and outperforms prior methods in all but one backbone--benchmark pair.
\end{itemize}
\vspace{-0.5em}
\section{Related Work}\label{sec:related}

\noindent\textbf{Test-Time Adaptation.}~~TTA updates a pre-trained model on unlabeled test data. For unimodal classifiers, TENT~\citep{wang2020tent} minimizes entropy through the BatchNorm parameters, EATA~\citep{niu2022efficient} and SAR~\citep{niu2023towards} filter and stabilize its updates, and RPL~\citep{rusak2022rpl} uses robust pseudo-labels instead. For VLMs, methods tune prompts~\citep{shu2022test,abdul2023align}, cache features~\citep{karmanov2024efficient}, average weights across prompt templates~\citep{osowiechi2024watt}, fine-tune the visual projector~\citep{fahes2026clip}, build pseudo-labels from image and text similarities~\citep{a2025clipartt}, or adapt the LayerNorm parameters of both encoders~\citep{maharana2025batclip}. Like most recent methods that update the model~\citep{a2025clipartt,osowiechi2024watt,maharana2025batclip,lafon2025cliptta,bao2025mint}, we adapt only the affine parameters of the normalization layers. Our main evaluation is the episodic protocol, which resets the model at every corruption boundary, and as other studies do, we also evaluate continual adaptation without resets~\citep{wang2022cotta}, streams that mix domains~\citep{marsden2024universal}, and temporally correlated label streams~\citep{yuan2023rotta}.

\noindent\textbf{Embedding Geometry of VLMs.}~~The image features of a contrastive VLM lie in a narrow cone~\citep{liang2022mind,mistretta2025cross} that forms already at initialization and that contrastive training keeps apart from the text features by a modality gap~\citep{liang2022mind}. Under corruption, the variance of the image features shrinks~\citep{bao2025mint}, which compresses this cone further. Mint~\citep{bao2025mint} counters this compression in feature space by maximizing the variance between its predicted classes. However, the class means around which Mint spreads the features come from its own predictions, which compression has already biased, and its objective has no term on the class distribution of the predictions, so Mint does not act on the collapse directly. CAMA instead aligns this distribution, where compression turns into collapse, with a reference estimated with the bias reduced (Section~\ref{sec:method}).

\noindent\textbf{Prediction Collapse.}~~Several TTA methods counter concentrated predictions with an inequity penalty~\citep{niu2025adapt}, a diversity term over spectral experts~\citep{baek2026imse}, a batch-aware contrastive loss~\citep{lafon2025cliptta}, or an external semantic signal~\citep{wang2026ced}. Analyses of the failure of entropy minimization explain it by general properties of the model and its adaptation, such as the mismatch between entropy and contrastive pre-training~\citep{lafon2025cliptta} or embedding drift over long adaptation~\citep{press2024entropy}, rather than by compression under corruption. Moreover, the observations of Mint and ClipTTA both come from CLIP alone, and to show that they reflect a property common to contrastive VLMs, we measure both the compression and the collapse on seven different VLM backbones.

Information maximization (IM), on which the mutual information term of CAMA builds, counters such collapse by maximizing the mutual information between inputs and predictions~\citep{krause2010discriminative,liang2020we,boudiaf2020transductive}, and some variants target a given class prior~\citep{krause2010discriminative,boudiaf2020transductive}. Under strong compression, however, the predictions become alike, so this mutual information vanishes and no longer resists the collapse. CAMA keeps it as its adaptation signal and, unlike these variants, sets the target and strength of its alignment for each stream.
\vspace{-0.5em}
\section{Setup and Notation}\label{sec:setup}

\noindent\textbf{Contrastive Vision--Language Classification.}~~A contrastive VLM has an image encoder $f_\theta$ and a text encoder $g_\theta$, where $\theta$ denotes the parameters adapted at test time. For $K$ classes, each class prompt gives a text feature $\mathbf{t}_k = g_\theta(\mathrm{prompt}_k)$, and each image $x_i$ gives an image feature $\mathbf{f}_i = f_\theta(x_i)$. The model scores class $k$ by the cosine similarity $\mathrm{sim}(\mathbf{f}_i, \mathbf{t}_k) = \hat{\mathbf{f}}_i^\top \hat{\mathbf{t}}_k$ of the $L^2$-normalized features $\hat{\mathbf{f}}_i, \hat{\mathbf{t}}_k \in \mathcal{S}^{d-1}$, the unit sphere of $\mathbb{R}^d$. The prediction $\mathbf{p}_i$ of sample $i$ is the softmax of these scores at the pre-training temperature $\tau$, $p_{ik} = \exp(\mathrm{sim}(\mathbf{f}_i, \mathbf{t}_k)/\tau) / \sum_j \exp(\mathrm{sim}(\mathbf{f}_i, \mathbf{t}_j)/\tau)$. We write $\Delta^{K-1}$ for the probability simplex, $\Delta^{K-1}_\circ$ for its interior, $H$ for the Shannon entropy, and $\KL$ for the Kullback--Leibler divergence.

\noindent\textbf{Online Test-Time Adaptation.}~~Test batches $X_t = \{x_1, \ldots, x_B\}$ of size $B$ arrive one after another from a corrupted stream. The model may update its parameters once per batch and never sees test labels or the type and severity of the corruption. We adapt the affine parameters $\theta$ of the normalization layers of both encoders and keep prompts and all other weights frozen. Our analysis tracks the average prediction of a batch because it shows collapse directly. A collapsed batch puts most of this average on a few classes.

\vspace{-0.5em}
\section{Compression-Induced Prediction Collapse}\label{sec:mechanism}

This section shows how the feature compression and the prediction collapse that Mint~\citep{bao2025mint} and ClipTTA~\citep{lafon2025cliptta} observed separately are connected, and how entropy minimization reinforces this connection.

\noindent\textbf{Notation.}~~The normalized image features of a batch are \emph{$\varepsilon$-concentrated} if they all lie within distance $\varepsilon$ of one direction $\tilde{\boldsymbol{\mu}} \in \mathcal{S}^{d-1}$, so a small $\varepsilon$ means strong compression. We write $\pbar = \frac{1}{B}\sum_i \mathbf{p}_i$ for the average prediction of a batch and $\mathcal{L}_{\mathrm{ent}} = \frac{1}{B}\sum_i H(\mathbf{p}_i)$ for the loss of entropy minimization. Proposition~\ref{prop:sink} shows that compression alone concentrates the predictions on one class.

\begin{proposition}[Compression Concentrates the Predictions]\label{prop:sink}
If the normalized image features are $\varepsilon$-concentrated around $\tilde{\boldsymbol{\mu}}$, then for $k^* = \arg\max_k \hat{\mathbf{t}}_k^\top \tilde{\boldsymbol{\mu}}$ with margin $\rho = \min_{j \neq k^*} (\hat{\mathbf{t}}_{k^*} - \hat{\mathbf{t}}_j)^\top \tilde{\boldsymbol{\mu}}$, provided $\rho > 2\varepsilon$,
\begin{equation*}
\bar{p}_{k^*} \geq 1 - (K-1) \cdot \exp\!\big(\!-(\rho - 2\varepsilon)/\tau\big).
\end{equation*}
\end{proposition}

Here $k^*$ is the class whose text feature is closest to the shared direction $\tilde{\boldsymbol{\mu}}$, and the margin $\rho$ is its lead over the runner-up class. The proposition says that every sample of a compressed batch ranks $k^*$ first, and that the average prediction concentrates on $k^*$ more strongly as $\varepsilon$ or $\tau$ gets smaller. Because each feature lies within $\varepsilon$ of $\tilde{\boldsymbol{\mu}}$, each cosine differs from its value at $\tilde{\boldsymbol{\mu}}$ by at most $\varepsilon$, so the lead of $k^*$ over any other class shrinks by at most $2\varepsilon$, and the softmax turns the remaining lead, divided by $\tau$, into the exponential bound. The $\varepsilon$-concentration assumption is reasonable because the image features of contrastive VLMs already lie in a narrow cone~\citep{liang2022mind,mistretta2025cross} that corruption compresses further~\citep{bao2025mint}. Feature compression therefore biases the predictions toward $k^*$.

Proposition~\ref{prop:amplify} shows what entropy minimization does to the bias.

\begin{proposition}[Entropy Minimization Amplifies the Bias]\label{prop:amplify}
If every sample of a batch ranks $k^*$ first, the loss of entropy minimization $\mathcal{L}_{\mathrm{ent}}$ bounds the share of $k^*$ from below,
\begin{equation*}
\bar{p}_{k^*} \geq \exp(-\mathcal{L}_{\mathrm{ent}}).
\end{equation*}
Hence $\bar{p}_{k^*} \to 1$ as entropy minimization drives $\mathcal{L}_{\mathrm{ent}} \to 0$, and the batch collapses onto $k^*$.
\end{proposition}

Proposition~\ref{prop:amplify} shows that, once every sample ranks $k^*$ first, the lower entropy minimization drives its loss, the larger the share of $k^*$ must be. This premise holds for a compressed batch by Proposition~\ref{prop:sink}, and it persists under entropy minimization, because its gradient with respect to the logits raises the top logit of each sample the most. Entropy minimization on biased predictions therefore reinforces the bias until the predictions collapse. Appendix~\ref{app:proofs} proves both propositions and the gradient property.

\begin{figure*}[t]
\centering
\includegraphics[width=\linewidth]{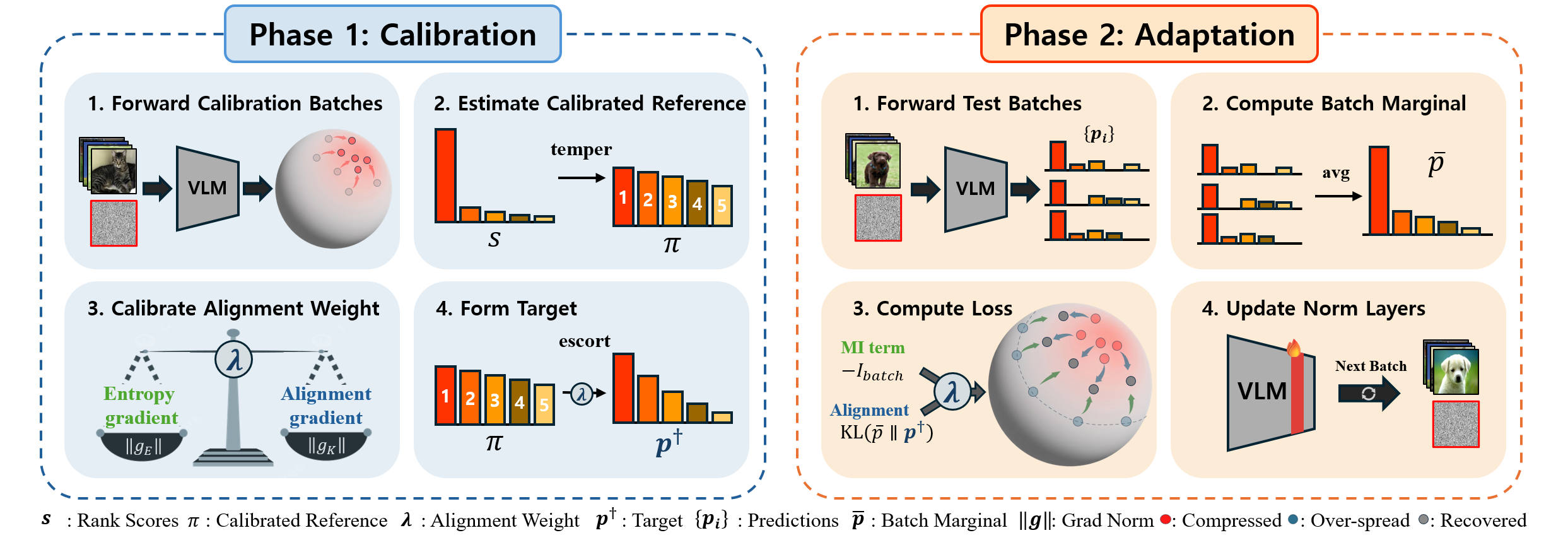}
\caption{CAMA calibrates once and then adapts with the calibrated quantities fixed. Phase 1 estimates the reference $\bpi$ from rank scores, sets the weight $\lambda$ by balancing the entropy and alignment gradients, and forms the target $\pdag$ as the escort of $\bpi$. Phase 2 minimizes the MI term and the alignment term on each batch. The alignment term spreads the compressed features (red) and the MI term pulls back the over-spread ones (blue), so both move toward the recovered features (grey).}
\label{fig:method}
\vspace{-1em}
\end{figure*}

\section{Method}\label{sec:method}

As Section~\ref{sec:mechanism} showed, compression creates a bias in the class distribution of the predictions, and entropy minimization amplifies it into collapse. Preventing the collapse therefore requires that adaptation proceed with this bias corrected, which prior methods do not achieve. Mint~\citep{bao2025mint} and BATCLIP~\citep{maharana2025batclip} adapt the features, by increasing the variance between the predicted classes or by aligning the image prototypes of these classes with their text features, but the classes are defined by the biased predicted labels, so their correction inherits the bias. ClipTTA~\citep{lafon2025cliptta} treats the collapse as a flaw of the entropy loss and replaces the loss, without accounting for the compression of the features from which the collapse begins. CAMA, which this section presents, performs adaptation and the correction of the bias in a single objective. It keeps the adaptation of entropy minimization and acts directly on the class distribution of the predictions, where the bias appears, so that the bias is undone as adaptation proceeds. Figure~\ref{fig:method} gives an overview of CAMA, and the rest of this section presents the objective, its weight, and the proof that it prevents collapse.

\subsection{Adaptation and Alignment}\label{sec:objective}

\noindent\textbf{Maximizing Batch Mutual Information.}~~Entropy minimization raises only the confidence of each prediction, so nothing keeps all predictions of a batch from moving to the same class. CAMA therefore uses the mutual information of the batch, which looks at the batch as a whole. We call the average prediction $\pbar$ the \emph{batch marginal}. The batch mutual information (MI), the mutual information between a sample of the batch and its predicted class, is $\Ibatch = H(\pbar) - \frac{1}{B}\sum_i H(\mathbf{p}_i)$, and CAMA maximizes it through the MI term $-\Ibatch$ of its loss. Lowering the mean entropy makes each prediction confident, as entropy minimization does, and this is the adaptation signal. Raising the entropy of the batch marginal makes the predictions of a batch diverse.

This term alone is not enough, however. The batch mutual information can also be written as $\Ibatch = \frac{1}{B}\sum_i \KL(\mathbf{p}_i \| \pbar)$, so it measures how much each prediction differs from the average prediction. Under the $\varepsilon$-concentration of Proposition~\ref{prop:sink}, for every class the cosines of any two samples each lie within $\varepsilon$ of the value at $\tilde{\boldsymbol{\mu}}$, so they differ by at most $2\varepsilon$ and their logits by at most $2\varepsilon/\tau$. Every $\mathbf{p}_i$ is then close to $\pbar$, and $\Ibatch$ vanishes as $\varepsilon/\tau \to 0$. In this limit the MI term has the same value whatever the batch marginal, so it cannot prevent collapse. This is the worst case for collapse, and we state it as an assumption for the analysis of Section~\ref{sec:theory}.

\begin{assumption}[Identical Predictions]\label{asm:a1}
$\mathbf{p}_1 = \cdots = \mathbf{p}_B = \mathbf{p}$, equivalently $\Ibatch = 0$.
\end{assumption}

Besides the diversity of the predictions, the objective therefore needs a term that corrects the bias itself.

\noindent\textbf{Aligning to a Calibrated Reference.}~~The alignment term fills this gap. It pulls $\pbar$ toward a target $\pdag$ through $\KL(\pbar \| \pdag)$, and because it depends on $\pbar$ alone, it still acts in the limit of Assumption~\ref{asm:a1}. The target must be chosen with care, because $\pbar$ mixes two signals. One is the class distribution of the stream, which the target should follow. The other is the corruption bias of Proposition~\ref{prop:sink}, which the target should reduce. A preset target ignores the first signal. The model's probabilities cannot serve as the target either. For example, the average of the model's outputs carries the corruption bias in full, and using it as the target lowers accuracy sharply (Section~\ref{sec:attribution}). CAMA therefore builds the target from a \emph{calibrated reference} $\bpi \in \Delta^{K-1}_\circ$, which estimates the class distribution of the stream with the bias reduced.

Before adaptation, CAMA runs a short \emph{calibration} step. On the first $N_{\mathrm{cal}}$ batches of the stream, without updating any parameter, it sets the reference and the alignment weight, and both then stay fixed. Because the softmax amplifies the corruption bias in the model's probabilities, whereas a rank does not grow with it, calibration estimates the reference using only the order in which each sample ranks the classes. Each sample $i$ gives class $k$ the harmonic weight $w_{ik} = 1/\mathrm{rank}_{ik}$, where $\mathrm{rank}_{ik} = 1$ for its top class, so that the runner-up classes, which a hard vote would discard, still count. Averaging the normalized weights over the $n$ calibration samples gives the rank score $s_k = \frac{1}{n}\sum_i w_{ik} / \sum_j w_{ij}$. The reference is then
\begin{equation}\label{eq:pi_def}
\pi_k = \frac{(s_k + \zeta)^\kappa}{\sum_j (s_j + \zeta)^\kappa},
\end{equation}
with a smoothing constant $\zeta > 0$ and a tempering exponent $\kappa \in (0, 1)$. Ranks make the estimate independent of the logit scale and of $\tau$, and both constants move every ratio between two classes toward one. The calibrated reference therefore keeps the class order of the rank scores but shrinks the corruption bias.

\subsection{Balancing Adaptation and Alignment}\label{sec:calibration}

The CAMA loss combines the two terms,
\begin{equation}\label{eq:lossB}
\mathcal{L} = -\Ibatch + (\lambda - 1) \cdot \KL(\pbar \| \pdag),
\end{equation}
where $\lambda \geq 1$ sets the strength of the alignment and $\pdag$ is the target that Eq.~\eqref{eq:pdag} builds from $\bpi$. Because maximizing $\Ibatch$ raises $H(\pbar)$, the MI term already spreads the batch marginal. To offset this spread, the target is a sharpened form of $\bpi$, called its \emph{escort},
\begin{equation}\label{eq:pdag}
p^\dagger_k = \frac{\pi_k^{\lambda/(\lambda-1)}}{\sum_j \pi_j^{\lambda/(\lambda-1)}}.
\end{equation}
Its exponent $\lambda/(\lambda-1)$ exceeds one, so $\pdag$ keeps the class order of $\bpi$ but is sharper. Since $\bpi$ gives every class positive mass, so does $\pdag$. With this exponent and $\lambda > 1$, Eq.~\eqref{eq:lossB} equals, up to a constant (Appendix~\ref{app:decomp_proof}),
\begin{equation}\label{eq:solver}
\mathcal{L}_{\mathrm{solver}} = \frac{1}{B}\sum_{i=1}^B H(\mathbf{p}_i) + \lambda \cdot \KL(\pbar \| \bpi).
\end{equation}
The first term is the loss of entropy minimization, and the second aligns $\pbar$ with $\bpi$ at the weight $\lambda$, which we call the \emph{alignment weight}. The target $\pdag$ of Eq.~\eqref{eq:lossB} is sharper than $\bpi$ only to offset the spread of its MI term, so the net effect is to align $\pbar$ with $\bpi$. The weight $\lambda$ thus sets how strongly the alignment resists the push for confidence.

In addition, the adaptation and alignment terms must be carefully balanced. If the alignment is too weak, the collapse gets through, and if it is too strong, it holds adaptation back. The two terms of Eq.~\eqref{eq:solver} differ in scale, so their values cannot be compared, and CAMA instead balances them by the size of their gradients, which measures how strongly each moves the model. In addition, the alignment is computed from the batch marginal of a single batch, which is inaccurate when the batch is small relative to the number of classes, so the weight also needs a correction for the batch size. CAMA sets $\lambda$ in the same calibration step as $\bpi$.

\noindent\textbf{Gradient-Balanced Weight.}~~Under compression, the entropy term raises the confidence of every prediction in the shared class $k^*$, while the alignment spreads $\pbar$ toward $\bpi$, so the two conflict. CAMA first sets a weight $\lambda_0$, before the batch-size correction below, to the ratio of the gradient norms of the two terms of Eq.~\eqref{eq:solver}, each summed over the $N_{\mathrm{cal}}$ calibration batches, with $\pbar_b$ the marginal of batch $b$,
\begin{equation*}
\lambda_0 = \frac{\sum_{b=1}^{N_{\mathrm{cal}}} \|\nabla_\theta \tfrac{1}{B}\sum_i H(\mathbf{p}_i)\|}{\sum_{b=1}^{N_{\mathrm{cal}}} \|\nabla_\theta \KL(\pbar_b \| \bpi)\|}.
\end{equation*}
At this ratio the two terms pull with equal strength, which keeps either from overriding the other. Unless the two gradients point in exactly opposite directions, one small update then lowers both terms (Appendix~\ref{app:lambda0_derivation}).

\noindent\textbf{Batch-Size Shrinkage.}~~When the batch is small relative to the number of classes, $\pbar$ is inaccurate, and the alignment then follows noise rather than the corruption bias. The reason is that $\pbar$ estimates the shares of $K$ classes from only $B$ predictions. This noise raises $\KL(\pbar \| \bpi)$ on average by up to about $(K-1)/(2B)$, the classical finite-sample error of an entropy estimate, even when there is no corruption bias to correct. CAMA therefore keeps only the part of the alignment weight that a batch resolves. The $K$ entries of $\pbar$ sum to one, so $\pbar$ has $K - 1$ degrees of freedom, and the $B$ samples must estimate all of them. Smoothing $\pbar$ toward $\bpi$ as a prior worth $K - 1$ samples, one per degree of freedom, puts the weight $B/(B + K - 1)$ on the batch. CAMA applies this factor, which we call the \emph{shrinkage}, to $\lambda_0 - 1$, the weight of the alignment term, so the MI term keeps its full weight,
\begin{equation}\label{eq:lambda}
\lambda = 1 + (\lambda_0 - 1) \cdot \frac{B}{B + K - 1}.
\end{equation}
The factor is close to one when $B \gg K$ and small when $B \ll K$, and it depends only on $B$ and $K$. The weight is floored at $1$, so that the weight $\lambda - 1$ of the alignment term in Eq.~\eqref{eq:lossB} is never negative.

\begin{table*}[t]
\centering
\caption{\textbf{CAMA has the best accuracy in every corruption family on all three benchmarks}. Online accuracy (\%) on CLIP ViT-B/16. \emph{Each column averages the corruptions of one family~\citep{hendrycks2019benchmarking}, and Appendix~\ref{app:additional} gives every corruption}. Dashed lines separate general from vision--language TTA methods, and $^*$ marks the online variants of WATT. Best per column in \textbf{bold}.}
\label{tab:main}
\small
\setlength{\tabcolsep}{3pt}
\renewcommand{\arraystretch}{0.86}
\resizebox{\textwidth}{!}{%
\begin{tabular}{l ccccc ccccc}
\toprule
 & \multicolumn{5}{c}{CIFAR-10-C} & \multicolumn{5}{c}{CIFAR-100-C} \\
\cmidrule(lr){2-6}\cmidrule(lr){7-11}
Method & Noise & Blur & Weather & Digital & Mean & Noise & Blur & Weather & Digital & Mean \\
\midrule
ZS~\citep{radford2021learning} & 44.68 & 63.55 & 76.48 & 55.81 & 61.16 & 22.10 & 38.42 & 49.17 & 30.03 & 35.79 \\
TENT~\citep{wang2020tent} & 23.96 & 65.50 & 84.81 & 64.34 & 62.03 & 8.05 & 41.89 & 51.72 & 42.71 & 37.96 \\
RoTTA~\citep{yuan2023rotta} & 45.83 & 63.88 & 77.00 & 56.82 & 61.89 & 23.01 & 38.38 & 49.38 & 30.44 & 36.12 \\
RPL~\citep{rusak2022rpl} & 24.54 & 63.67 & 85.31 & 63.32 & 61.52 & 6.87 & 42.95 & 52.77 & 43.40 & 38.47 \\
SAR~\citep{niu2023towards} & 54.05 & 67.86 & 80.06 & 64.17 & 67.37 & 28.42 & 42.35 & 53.68 & 37.12 & 41.19 \\
\noalign{\vskip 1.5pt}\cdashline{1-11}\noalign{\vskip 2pt}
TPT~\citep{shu2022test} & 46.85 & 65.69 & 78.15 & 59.68 & 63.64 & 21.53 & 38.72 & 49.52 & 31.18 & 36.15 \\
VTE~\citep{dobler2024vte} & 50.97 & 64.71 & 77.83 & 59.80 & 64.15 & 21.62 & 36.20 & 47.79 & 31.08 & 35.01 \\
WATT-P$^*$~\citep{osowiechi2024watt} & 51.61 & 67.29 & 80.43 & 61.64 & 66.15 & 23.35 & 37.80 & 48.80 & 30.19 & 35.81 \\
WATT-S$^*$~\citep{osowiechi2024watt} & 62.24 & 72.42 & 83.84 & 70.13 & 72.82 & 24.18 & 38.80 & 49.59 & 31.14 & 36.71 \\
BATCLIP~\citep{maharana2025batclip} & 63.66 & 74.52 & 84.26 & 70.41 & 73.85 & 28.77 & 44.93 & 53.91 & 37.41 & 42.09 \\
Mint~\citep{bao2025mint} & 64.89 & 74.76 & 84.19 & 68.64 & 73.67 & 34.87 & 45.89 & 55.59 & 39.36 & 44.53 \\
ClipTTA~\citep{lafon2025cliptta} & 55.00 & 69.11 & 80.92 & 64.57 & 68.23 & 28.37 & 42.38 & 52.94 & 35.39 & 40.53 \\
CAMA (Ours) & \textbf{71.27} & \textbf{79.47} & \textbf{87.04} & \textbf{78.03} & \textbf{79.46} & \textbf{39.21} & \textbf{49.50} & \textbf{58.28} & \textbf{45.71} & \textbf{48.77} \\
\bottomrule
\end{tabular}%
\hspace{10pt}%
\begin{tabular}{l ccccc}
\toprule
 & \multicolumn{5}{c}{ImageNet-C} \\
\cmidrule(lr){2-6}
Method & Noise & Blur & Weather & Digital & Mean \\
\midrule
ZS~\citep{radford2021learning} & 11.92 & 21.41 & 38.07 & 23.54 & 24.52 \\
TENT~\citep{wang2020tent} & 6.09 & 23.88 & 38.99 & 26.88 & 25.15 \\
RoTTA~\citep{yuan2023rotta} & 12.21 & 21.61 & 38.22 & 23.63 & 24.70 \\
RPL~\citep{rusak2022rpl} & 10.01 & 22.96 & 38.50 & 25.09 & 25.08 \\
SAR~\citep{niu2023towards} & 19.70 & 26.08 & 40.41 & 30.23 & 29.73 \\
\noalign{\vskip 1.5pt}\cdashline{1-6}\noalign{\vskip 2pt}
TPT~\citep{shu2022test} & 9.38 & 22.06 & 39.70 & 24.46 & 24.87 \\
VTE~\citep{dobler2024vte} & 10.24 & 22.15 & 40.39 & 25.77 & 25.60 \\
StatA ($\gamma$=0.1)~\citep{zanella2025stata} & 10.88 & 19.87 & 37.31 & 22.70 & 23.47 \\
StatA ($\gamma$=-1)~\citep{zanella2025stata} & 12.42 & 21.04 & 38.17 & 23.98 & 24.67 \\
BATCLIP~\citep{maharana2025batclip} & 20.10 & 27.10 & 41.54 & 31.47 & 30.72 \\
Mint~\citep{bao2025mint} & 23.09 & 28.22 & 43.06 & 32.92 & 32.40 \\
ClipTTA~\citep{lafon2025cliptta} & 19.17 & 25.51 & 40.33 & 28.58 & 29.01 \\
CAMA (Ours) & \textbf{24.59} & \textbf{30.88} & \textbf{44.96} & \textbf{36.70} & \textbf{34.93} \\
\bottomrule
\end{tabular}}
\vspace{-1em}
\end{table*}

\subsection{Alignment-Induced Collapse Prevention}\label{sec:theory}

Proposition~\ref{prop:amplify} showed that entropy minimization amplifies the bias into collapse. This section shows that CAMA, which uses the MI term and the alignment term together, does not. We analyze the worst case of Assumption~\ref{asm:a1}, in which the MI term gives no protection. All $\mathbf{p}_i$ then equal a single prediction $\mathbf{p} = \pbar$, so the loss of Eq.~\eqref{eq:solver} becomes a function of $\mathbf{p}$ alone,
\begin{equation*}
\begin{aligned}
\mathcal{L}_\lambda(\mathbf{p}) &= H(\mathbf{p}) + \lambda\,\KL(\mathbf{p} \,\|\, \bpi) \\
&= (1-\lambda)\,H(\mathbf{p}) - \lambda\,\langle \mathbf{p}, \log\bpi \rangle.
\end{aligned}
\end{equation*}
Theorem~\ref{thm:anticollapse} gives its minimizers.

\Needspace*{4\baselineskip}
\begin{theorem}[Alignment Prevents the Collapse]\label{thm:anticollapse}
Under Assumption~\ref{asm:a1}, for $\bpi \in \Delta^{K-1}_\circ$, any $\lambda > 1$, and $\mathbf{p} \in \Delta^{K-1}$:
\begin{equation*}
\mathcal{L}_\lambda(\mathbf{p}) = (\lambda - 1)\,\KL(\mathbf{p} \,\|\, \pdag) + C,
\end{equation*}
where $\pdag$ is the escort of Eq.~\eqref{eq:pdag} and the constant $C$ does not depend on $\mathbf{p}$. The unique minimizer is therefore $\pdag$. For $0 \leq \lambda < 1$, which includes entropy minimization at $\lambda = 0$, every minimizer puts all its mass on one class, and at $\lambda = 1$, on the classes with the largest $\pi_k$.
\end{theorem}

As a result, every optimum of entropy minimization is a batch collapsed onto one class, whereas the unique optimum of CAMA gives every class positive mass, because $\bpi$ does. Prediction collapse thus cannot be an optimum of CAMA, so unlike entropy minimization, the optimization of CAMA moves away from collapse as it proceeds. This holds even in the worst case, where the alignment term acts alone. When the predictions of a batch differ, the MI term protects as well, and a batch whose predictions are diverse enough has a lower loss than every collapsed batch. Appendices~\ref{app:proofs} and~\ref{app:calibration} give the derivations and proofs of this section.
\vspace{-0.5em}
\begin{figure*}[!t]
\centering
\includegraphics[width=0.872\textwidth]{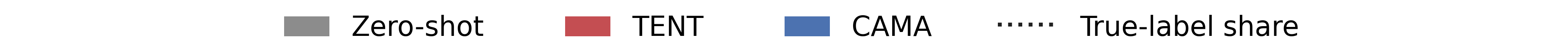}\\[1pt]
\begin{subfigure}[t]{0.32\textwidth}
\centering
\includegraphics[width=\linewidth]{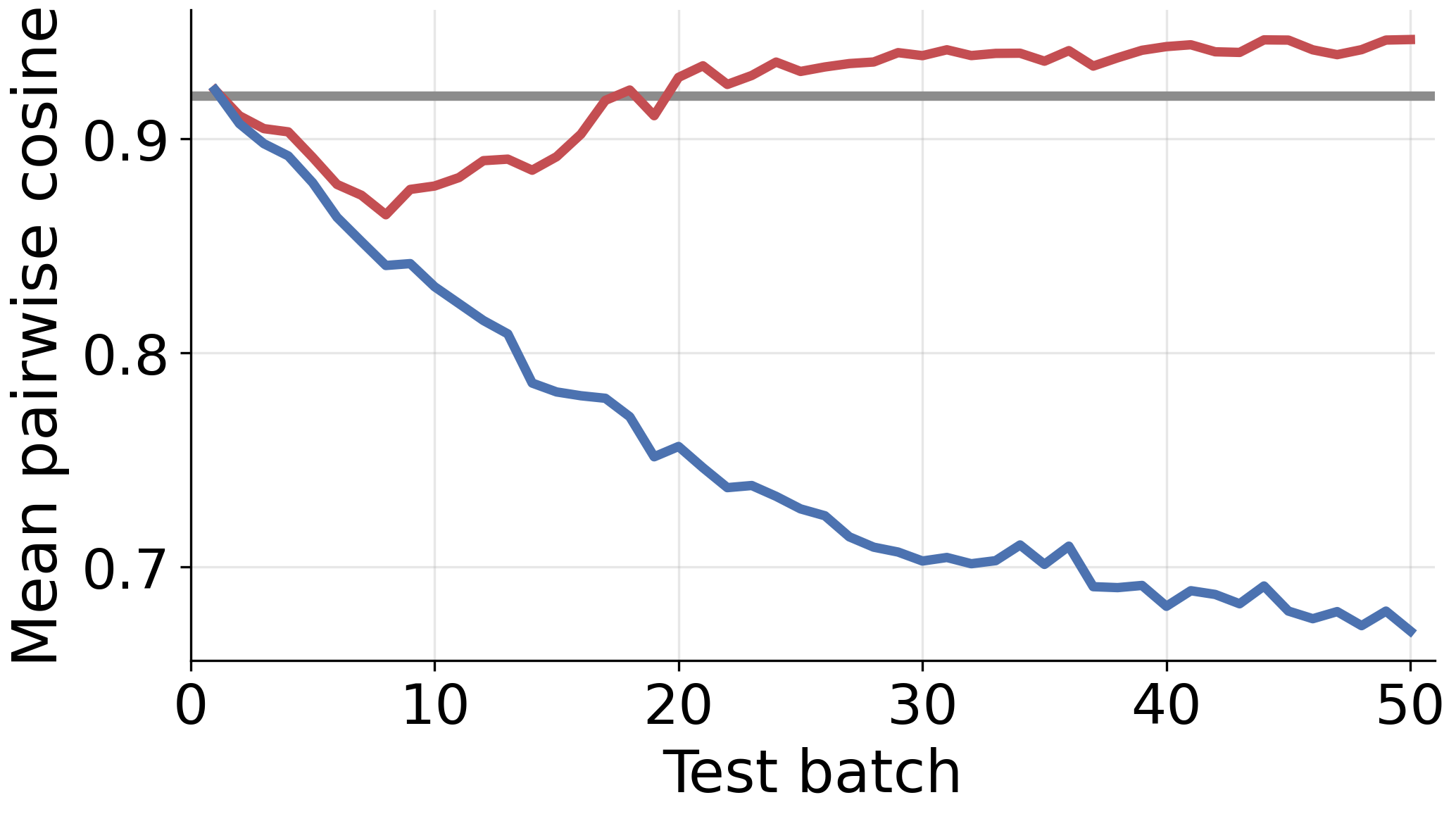}
\caption{Compression}
\end{subfigure}
\hfill
\begin{subfigure}[t]{0.32\textwidth}
\centering
\includegraphics[width=\linewidth]{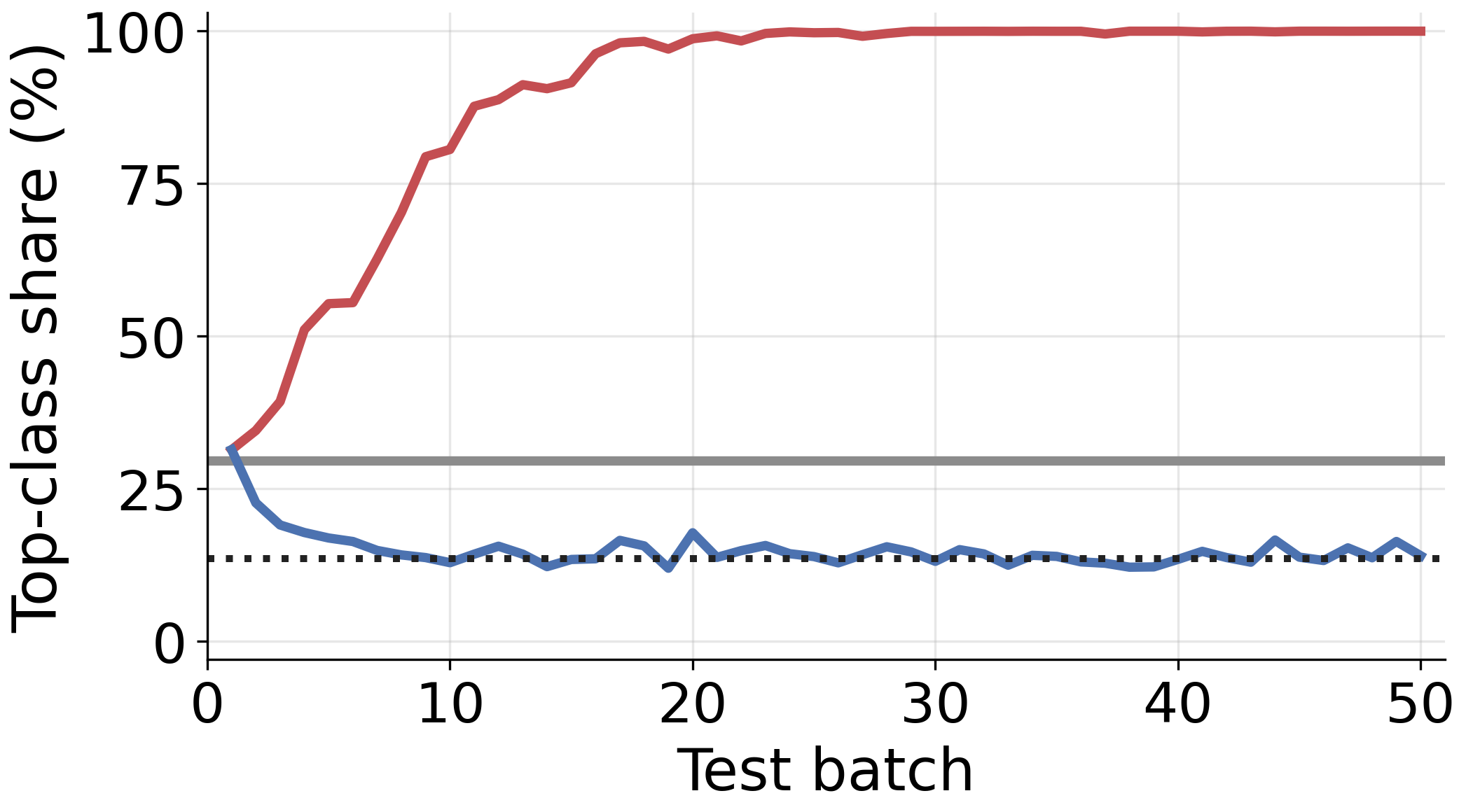}
\caption{Collapse}
\end{subfigure}
\hfill
\begin{subfigure}[t]{0.32\textwidth}
\centering
\includegraphics[width=\linewidth]{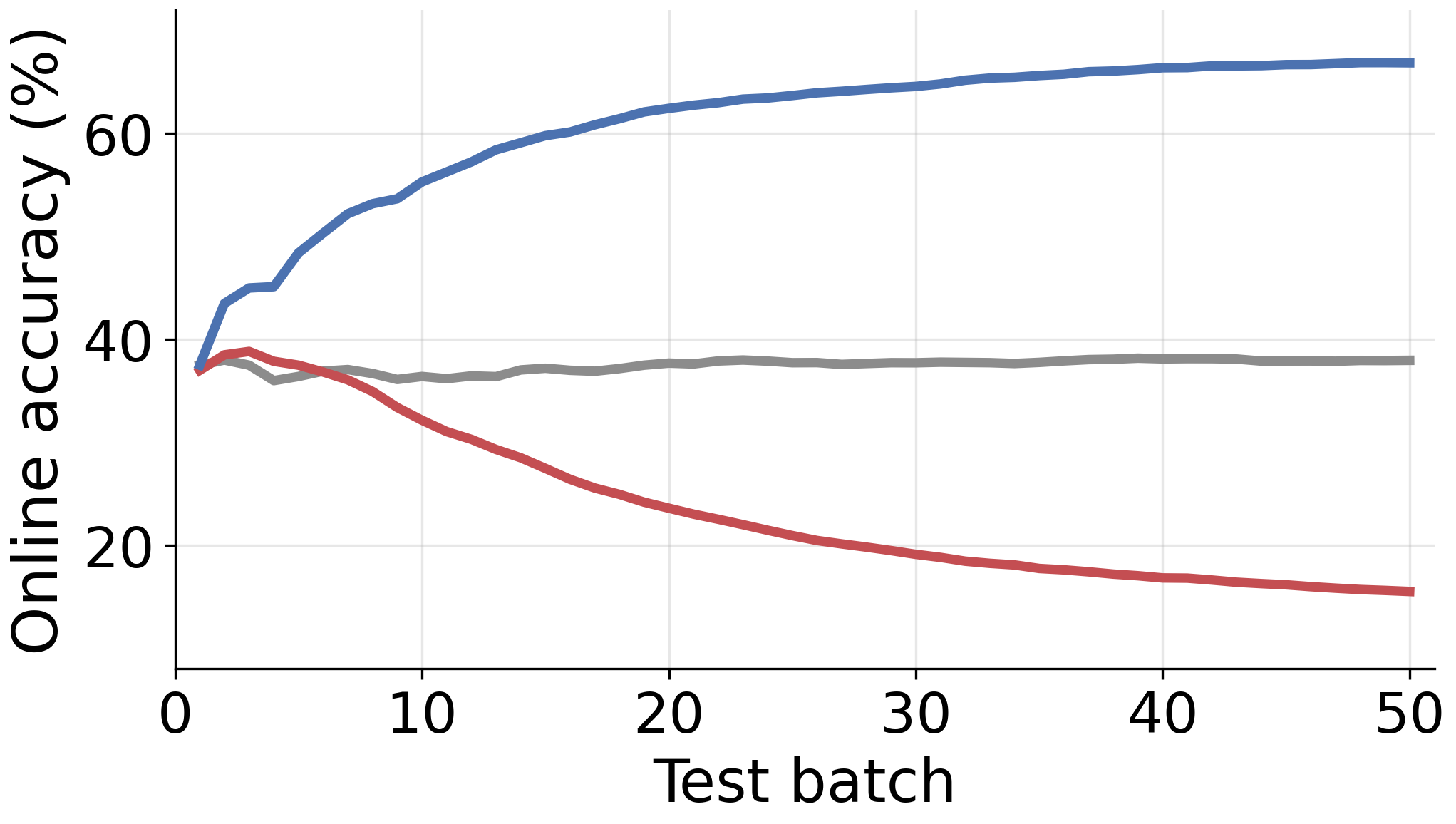}
\caption{Accuracy}
\end{subfigure}
\caption{As the test stream proceeds, compression, collapse, and accuracy loss build up together under TENT, whereas CAMA prevents them. CLIP ViT-B/16 on CIFAR-10-C Gaussian noise at severity 5. (a) Mean pairwise cosine of the image features. (b) Top-class share, fraction of a batch assigned to its most-predicted class, and true-label share, fraction in its most frequent true class. (c) Online accuracy.}
\vspace{-1em}
\label{fig:motivation}
\end{figure*}

\section{Experiments}

\subsection{Setup}

\noindent\textbf{Benchmarks.}~~We evaluate on CIFAR-10-C, CIFAR-100-C, and ImageNet-C, with $K = 10$, $100$, and $1000$ classes, each with 15 corruption types at severity 5~\citep{hendrycks2019benchmarking}. On ImageNet-C we use the 5,000-image RobustBench subset.

\noindent\textbf{Backbones.}~~We evaluate seven contrastive VLMs, CLIP ViT-B/16~\citep{radford2021learning}, SigLIP ViT-B/16~\citep{zhai2023sigmoid}, MetaCLIP ViT-B/16~\citep{xu2024demystifying}, EVA-CLIP EVA02-B/16~\citep{sun2023eva}, ConvNeXt-CLIP (ConvNeXt-Base, LAION-2B)~\citep{cherti2023reproducible}, and the ResNet CLIPs RN50 and RN101~\citep{radford2021learning}. Compared with CLIP ViT-B/16, SigLIP changes the training objective, MetaCLIP the data curation, EVA-CLIP the training recipe, and ConvNeXt-CLIP and the ResNet CLIPs the visual architecture.

\begin{table}[!t]
\centering
\caption{CAMA's accuracy lead carries over to the six contrastive VLMs other than CLIP ViT-B/16, where it is the best in all but one backbone--benchmark pair. Mean online accuracy (\%) over the 15 corruptions. Best per row in \textbf{bold}.}
\label{tab:backbones}
\footnotesize
\setlength{\tabcolsep}{3pt}
\resizebox{\columnwidth}{!}{%
\begin{tabular}{@{}ll ccccccc@{}}
\toprule
Backbone & Benchmark & ZS & TENT & SAR & BATCLIP & Mint & ClipTTA & CAMA \\
\midrule
\multirow{3}{*}{SigLIP~\citep{zhai2023sigmoid}} & C10-C & 59.04 & 59.93 & 65.70 & 72.80 & 70.01 & 67.46 & \textbf{78.43} \\
 & C100-C & 34.76 & 24.37 & 35.90 & 41.35 & 42.91 & 40.34 & \textbf{48.70} \\
 & IN-C & 27.91 & 28.67 & 30.29 & 29.89 & 33.26 & 31.12 & \textbf{37.20} \\
\midrule
\multirow{3}{*}{MetaCLIP~\citep{xu2024demystifying}} & C10-C & 75.94 & 84.50 & 80.14 & 85.17 & 83.25 & 80.58 & \textbf{87.33} \\
 & C100-C & 48.12 & 54.54 & 54.57 & 55.60 & 57.26 & 53.79 & \textbf{61.70} \\
 & IN-C & 31.49 & 32.95 & 35.07 & 36.02 & 36.26 & 33.73 & \textbf{39.89} \\
\midrule
\multirow{3}{*}{EVA-CLIP~\citep{sun2023eva}} & C10-C & 81.12 & 81.69 & 85.11 & 89.87 & 67.85 & 89.59 & \textbf{90.71} \\
 & C100-C & 58.34 & 57.34 & 63.13 & 61.98 & 43.25 & 67.98 & \textbf{71.01} \\
 & IN-C & 36.96 & 38.26 & 43.89 & 43.27 & 31.45 & 42.61 & \textbf{50.39} \\
\midrule
\multirow{3}{*}{ConvNeXt-CLIP~\citep{cherti2023reproducible}} & C10-C & 64.16 & 41.49 & 11.87 & 78.00 & 65.32 & 60.61 & \textbf{78.62} \\
 & C100-C & 38.84 & 14.18 & \phantom{0}2.71 & 47.37 & 33.87 & 33.75 & \textbf{53.11} \\
 & IN-C & 27.17 & 12.26 & \phantom{0}1.58 & 35.81 & 23.97 & 21.22 & \textbf{38.55} \\
\midrule
\multirow{3}{*}{RN50~\citep{radford2021learning}} & C10-C & 35.08 & 12.07 & 21.24 & 34.85 & 24.96 & 56.36 & \textbf{58.29} \\
 & C100-C & 15.41 & \phantom{0}2.05 & \phantom{0}6.34 & 15.76 & \phantom{0}7.50 & 26.85 & \textbf{31.75} \\
 & IN-C & 11.15 & \phantom{0}7.27 & \phantom{0}8.81 & 12.35 & \phantom{0}5.33 & 18.90 & \textbf{21.51} \\
\midrule
\multirow{3}{*}{RN101~\citep{radford2021learning}} & C10-C & 42.47 & 12.97 & 29.29 & 27.94 & 24.64 & \textbf{66.50} & 64.03 \\
 & C100-C & 20.59 & \phantom{0}2.62 & \phantom{0}8.90 & 15.07 & \phantom{0}7.27 & 34.63 & \textbf{37.18} \\
 & IN-C & 15.18 & \phantom{0}9.75 & 13.01 & 10.49 & \phantom{0}4.25 & 23.68 & \textbf{24.65} \\
\bottomrule
\end{tabular}}
\vspace{-1em}
\end{table}

\noindent\textbf{Baselines.}~~Besides zero-shot (ZS)~\citep{radford2021learning}, we compare against general TTA methods, namely TENT~\citep{wang2020tent}, the representative entropy minimization method, RoTTA~\citep{yuan2023rotta}, RPL~\citep{rusak2022rpl}, and SAR~\citep{niu2023towards}, and against vision--language TTA methods, namely TPT~\citep{shu2022test}, VTE~\citep{dobler2024vte}, WATT-P$^*$~\citep{osowiechi2024watt}, WATT-S$^*$~\citep{osowiechi2024watt}, BATCLIP~\citep{maharana2025batclip}, and StatA~\citep{zanella2025stata}. We report all results of entropy minimization with TENT. We also compare against two recent methods that target the compression and the collapse, Mint~\citep{bao2025mint} and ClipTTA~\citep{lafon2025cliptta}. All baselines use their published settings, including which parameters they adapt. WATT-P$^*$ and WATT-S$^*$ are the online variants of WATT defined by~\citet{maharana2025batclip}, adapting once per prompt template on each batch.

\noindent\textbf{Protocol.}~~We run all methods under BATCLIP's published protocol for each benchmark and re-tune no method per backbone, CAMA included. Following the standard practice of online TTA, every method makes one update per test batch and is scored by its online accuracy over the entire stream. CAMA calibrates on the first $N_{\mathrm{cal}} = \max(3, \lceil K/B \rceil)$ batches of each stream at the protocol batch size, so that the calibration samples number at least $K$. We measure compression by the mean pairwise cosine of the image features and collapse by the top-class share, the share of the most-predicted class, and a method collapses on a stream when this share exceeds one half. Appendices~\ref{app:implementation} and~\ref{app:sensitivity} give the implementation details and the sensitivity analyses.

\subsection{Main Results}\label{sec:results}

\noindent\textbf{Results on CLIP ViT-B/16.}~~CAMA has the best accuracy on all three benchmarks in Table~\ref{tab:main}. The strongest baseline is BATCLIP on CIFAR-10-C and Mint on CIFAR-100-C and ImageNet-C, and CAMA leads it by 5.6\%, 4.2\%, and 2.5\%. It is also first on all but one corruption of each CIFAR benchmark and on every ImageNet-C corruption.

\noindent\textbf{Generalization across Backbones.}~~CAMA is the best method in all but one row of Table~\ref{tab:backbones}, and that row is RN101 on CIFAR-10-C. On CIFAR-100-C and ImageNet-C, CAMA leads the strongest baseline on every backbone, by up to 5.8\% and 6.5\%.

\noindent\textbf{Robustness across Test Streams.}\label{sec:streams}~~Run continually across all fifteen corruptions with one calibration and no resets, CAMA remains the best method on both CIFAR benchmarks (Table~\ref{tab:streams} in Appendix~\ref{app:additional}). On CIFAR-10-C it stays within 0.6\% of its episodic accuracy, and on CIFAR-100-C its continual score of 50.68 even exceeds the episodic 48.77. BATCLIP, Mint, and TENT lose 10\% to 52\%. CAMA is also the best method on the streams that mix the corruptions, the severities, or both. On temporally correlated streams, CAMA with a replay buffer is the best method even when the baselines get the same buffer.

In summary, the strongest baseline changes with the benchmark and the backbone, whereas CAMA leads consistently across backbones, benchmarks, and stream types. CAMA is not fitted to one setting but targets the compression--collapse phenomenon itself, which contrastive VLMs share (Section~\ref{sec:analysis}).

\subsection{Mechanism Analysis}\label{sec:analysis}

\begin{figure}[!t]
\centering
\includegraphics[width=\linewidth]{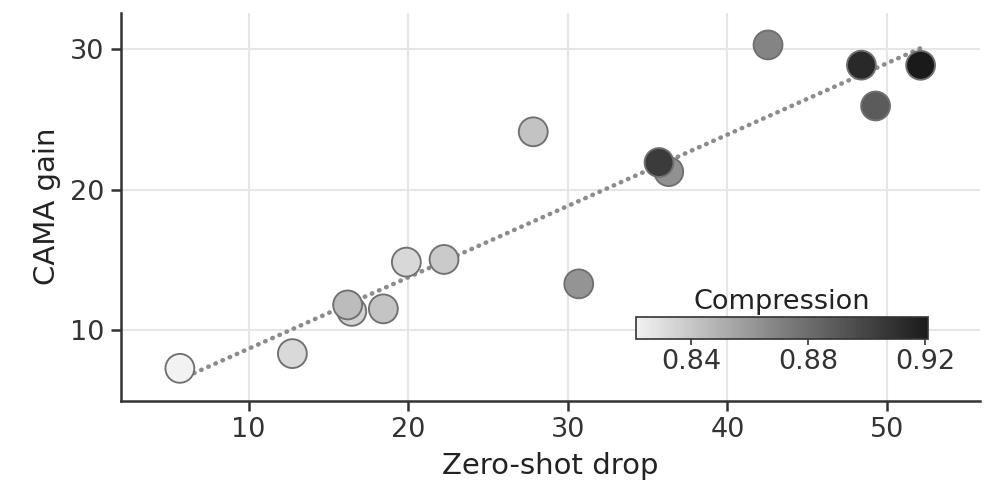}
\vspace{-2em}
\caption{Stronger compression comes with a larger zero-shot drop and a larger gain of CAMA. Each point is one CIFAR-10-C corruption on CLIP ViT-B/16. Compression is weakest for the weather and most blur corruptions (cosine 0.82--0.85), stronger for most digital ones (0.86--0.87), and strongest for noise (0.90--0.92). The zero-shot drop is the accuracy lost from clean data, and the CAMA gain is the accuracy CAMA adds over zero-shot.}
\vspace{-0.5em}
\label{fig:cone}
\end{figure}

\begin{table}[t]
\centering
\caption{The compression--collapse phenomenon and its prevention by CAMA appear consistently across backbones on the CIFAR-10-C corruptions where TENT collapses. TENT raises the feature cosine, which measures compression, raises the top share, which measures collapse, and lowers accuracy, whereas CAMA lowers the feature cosine, keeps the top share near its true 10\%, and raises accuracy. Means over these corruptions, best in \textbf{bold}.}
\label{tab:mechanism}
\footnotesize
\setlength{\tabcolsep}{1.9pt}
\begin{tabular}{@{}l ccc cc ccc@{}}
\toprule
 & \multicolumn{3}{c}{Feature cosine $\downarrow$} & \multicolumn{2}{c}{Top share (\%) $\downarrow$} & \multicolumn{3}{c}{Accuracy (\%) $\uparrow$} \\
\cmidrule(lr){2-4}\cmidrule(lr){5-6}\cmidrule(l){7-9}
Backbone & \textcolor{black!55}{ZS} & TENT & CAMA & TENT & CAMA & \textcolor{black!55}{ZS} & TENT & CAMA \\
\midrule
CLIP B/16 & \textcolor{black!55}{0.91} & 0.92 & \textbf{0.68} & 80.1 & \textbf{12.6} & \textcolor{black!55}{43.7} & 23.1 & \textbf{70.2} \\
SigLIP & \textcolor{black!55}{0.87} & 0.89 & \textbf{0.64} & 76.9 & \textbf{12.7} & \textcolor{black!55}{40.6} & 26.4 & \textbf{67.3} \\
EVA-CLIP & \textcolor{black!55}{0.70} & 0.93 & \textbf{0.28} & 60.7 & \textbf{10.4} & \textcolor{black!55}{73.8} & 44.2 & \textbf{86.2} \\
ConvNeXt & \textcolor{black!55}{0.68} & 0.89 & \textbf{0.39} & 72.8 & \textbf{11.0} & \textcolor{black!55}{58.8} & 31.4 & \textbf{76.3} \\
RN50 & \textcolor{black!55}{0.88} & 0.93 & \textbf{0.52} & 95.8 & \textbf{11.1} & \textcolor{black!55}{35.1} & 12.1 & \textbf{58.3} \\
RN101 & \textcolor{black!55}{0.91} & 0.98 & \textbf{0.66} & 94.6 & \textbf{11.0} & \textcolor{black!55}{42.5} & 13.0 & \textbf{64.0} \\
\bottomrule
\end{tabular}
\vspace{-0.5em}
\end{table}

\noindent\textbf{From Compression to Accuracy Loss.}~~Compression, collapse, and the loss of accuracy move together. Figure~\ref{fig:motivation} follows them over time on one stream. Corruption first compresses the features, raising their mean pairwise cosine from 0.79 on clean data to 0.92, and biases the predictions. Once TENT adapts, this bias grows into collapse until one class takes the whole stream, and accuracy falls below zero-shot. With CAMA, the compression is undone, the share of the most-predicted class stays at its true level, and accuracy stays above zero-shot throughout. The same relation holds across corruptions. Over the 15 CIFAR-10-C corruptions, which differ in how much they compress the features, stronger compression comes with a larger drop in zero-shot accuracy ($r = 0.91$), and a larger drop comes with a larger gain of CAMA ($r = 0.93$, Figure~\ref{fig:cone}).

\begin{figure}[!t]
\centering
\begin{subfigure}[t]{0.49\linewidth}
\centering
\includegraphics[width=\linewidth]{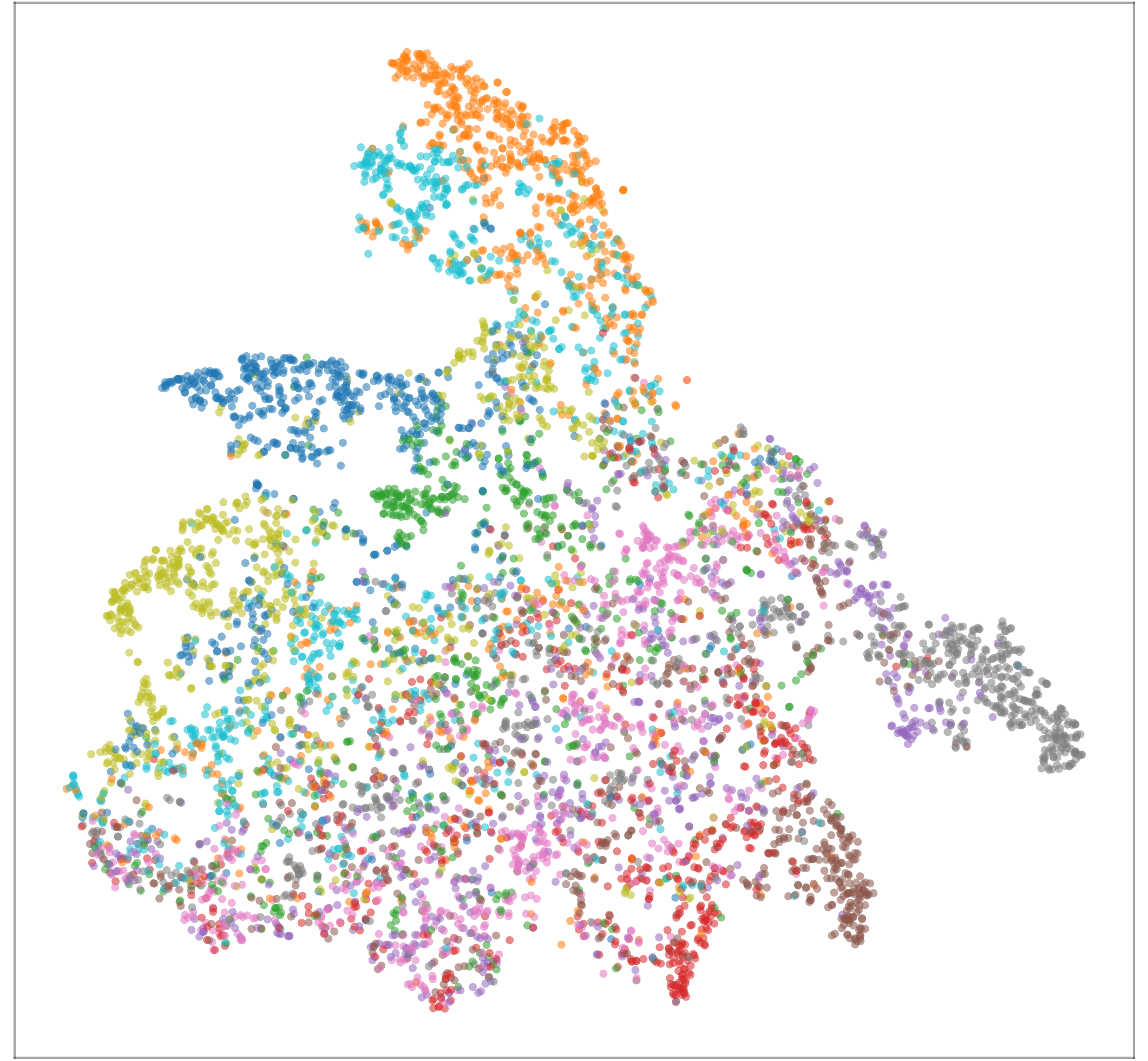}
\caption{Corrupted (zero-shot)}
\end{subfigure}
\hfill
\begin{subfigure}[t]{0.49\linewidth}
\centering
\includegraphics[width=\linewidth]{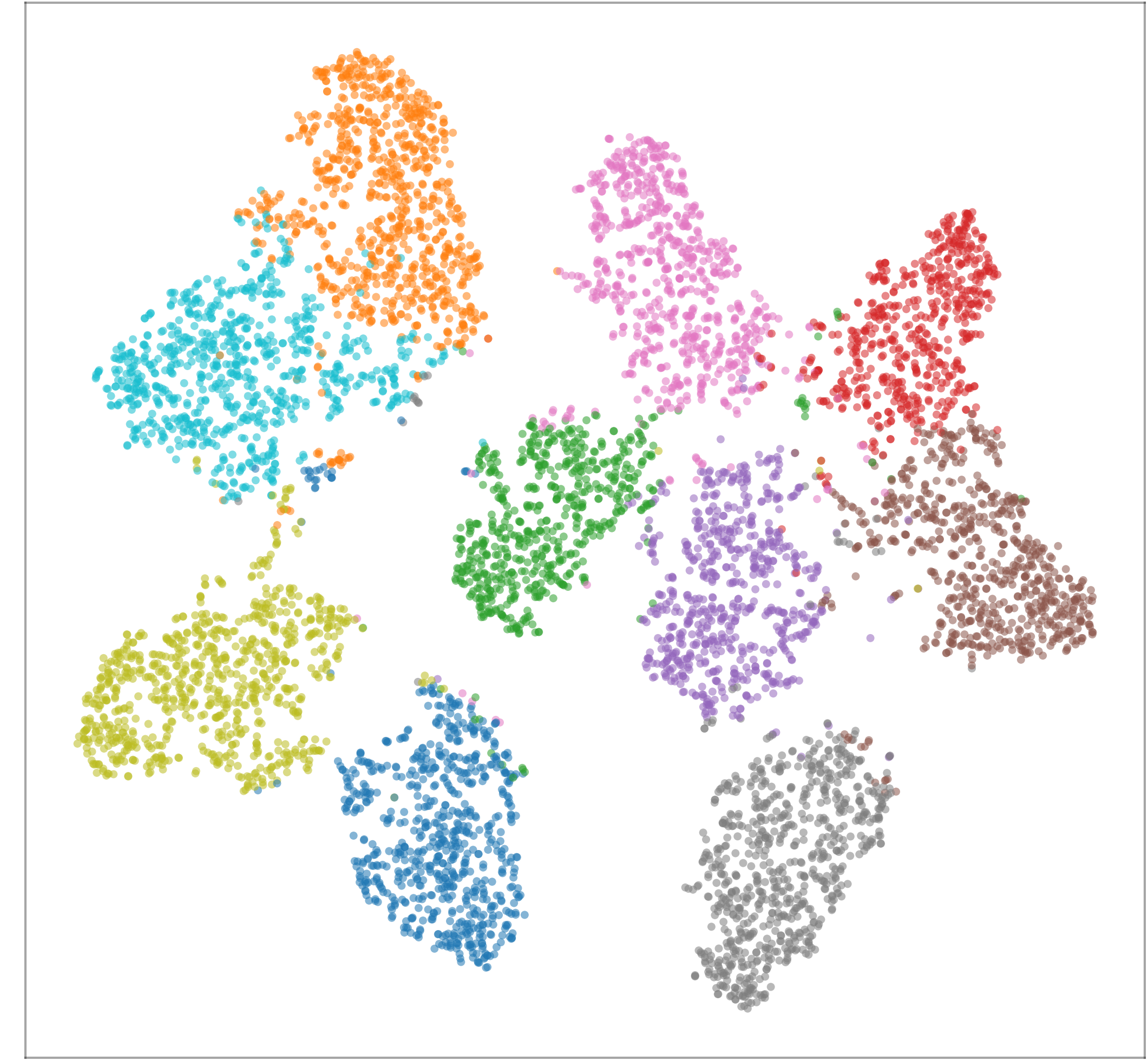}
\caption{After CAMA}
\end{subfigure}
\caption{Corruption merges class clusters, and CAMA separates them again. t-SNE of logits for correctly classified samples, colored by class, CLIP ViT-B/16 on CIFAR-10-C Gaussian noise.}
\label{fig:tsne_main}
\end{figure}

\noindent\textbf{Universality across Contrastive VLMs.}~~This phenomenon is not specific to CLIP but common to contrastive VLMs. On all seven backbones, which differ in architecture, training objective, and data, corruption compresses the features. TENT collapses on six of them, and where it collapses, it compresses the features further and assigns most of the stream, 61\% to 96\% on average, to one class (Table~\ref{tab:mechanism}). CAMA undoes the compression on every backbone and keeps the share of the most-predicted class near its true level. The compression--collapse phenomenon is thus a property of contrastive VLMs rather than of one model, and CAMA prevents it across them. Appendix~\ref{app:additional} gives every corruption.

\subsection{Ablation and Validation}\label{sec:attribution}

\begin{table}[!t]
\centering
\small
\caption{The calibrated reference of CAMA matches the oracle, whereas references that average the model's outputs lose accuracy. Online accuracy (\%) on CLIP ViT-B/16 with each reference, where the oracle is the true class distribution, mean-prob the mean prediction, and mean-logit the softmax of the mean logits.}
\label{tab:reference}
\setlength{\tabcolsep}{6pt}
\begin{tabular}{@{}lrrr@{}}
\toprule
Reference & CIFAR-10-C & CIFAR-100-C & ImageNet-C \\
\midrule
Oracle & 79.55 & 48.78 & 34.95 \\
\midrule
CAMA & \textbf{79.46} & \textbf{48.77} & \textbf{34.93} \\
Mean-prob & 68.00 & 39.49 & 23.41 \\
Mean-logit & 58.54 & 41.33 & 14.63 \\
\bottomrule
\end{tabular}
\vspace{-0.5em}
\end{table}

\begin{table}[!t]
\centering
\small
\caption{The shrinkage helps more for the smaller batches. Gain in online accuracy (\%) of the shrinkage over the unshrunk weight $\lambda_0$ on CLIP ViT-B/16. $^*$ marks the protocol batch size.}
\label{tab:shrinkage}
\setlength{\tabcolsep}{3.2pt}
\begin{tabular}{@{}lrrrrr@{}}
\toprule
& \multicolumn{5}{c}{Batch size $B$} \\
\cmidrule(l){2-6}
Benchmark ($K$) & $200$ & $64$ & $16$ & $8$ & $4$ \\
\midrule
CIFAR-10-C (10) & $0.0^*$ & $+0.2$ & $+0.5$ & $+2.3$ & $+7.0$ \\
CIFAR-100-C (100) & $+0.3^*$ & $+2.6$ & $+11.5$ & $+19.3$ & $+21.0$ \\
ImageNet-C (1000) & $+0.5$ & $+2.7^*$ & $+7.4$ & $+10.0$ & $+11.2$ \\
\bottomrule
\end{tabular}
\vspace{-1em}
\end{table}

We check that each component of CAMA does the job it was designed for.

\noindent\textbf{Batch Mutual Information.}~~The MI term keeps the adaptation of entropy minimization by making each prediction confident while keeping the predictions of a batch diverse. Without it, only the alignment term remains and the model receives no adaptation signal, so accuracy falls from 79.5 to 65.4 on CIFAR-10-C, from 48.8 to 33.1 on CIFAR-100-C, and from 34.9 to 16.4 on ImageNet-C. The drop grows with the number of classes, from 14.1\% to 15.7\% and 18.5\%. Preventing collapse is therefore not enough, and the adaptation of entropy minimization must be kept.

\noindent\textbf{Calibrated Reference.}~~The calibrated reference is the estimate of the class distribution to which the alignment term aligns the predictions, built from class ranks so that it carries little of the corruption bias. Table~\ref{tab:reference} shows that it serves this purpose. It is within 0.1\% of an oracle built from the true class distribution on all three benchmarks. References that average the model's outputs, mean-logit and mean-prob, instead inherit the corruption bias and lose 7.4\% to 20.9\%.

\noindent\textbf{Gradient-Balanced Weight.}~~Compared with the best weight for each corruption, chosen from test labels, the gradient-balanced weight of CAMA is within 1\% in accuracy on CIFAR-10-C (Appendix~\ref{app:sensitivity}). The shrinkage leaves accuracy unchanged where the batch is large enough and gains more as the batch shrinks, up to 21.0\% (Table~\ref{tab:shrinkage}).

\noindent\textbf{Recovery from Compression.}~~Finally, CAMA is designed to undo the compression that sets off the collapse. In Figure~\ref{fig:tsne_main}, the class clusters that merge under corruption separate again after CAMA. The mean pairwise cosine of the features also falls below its level on clean data (Table~\ref{tab:mechanism}). The image features show the same recovery in Appendix~\ref{app:visualizations}.

\section{Limitations and Conclusion}\label{sec:limitations}

\noindent\textbf{Limitations.}~~CAMA takes the average prediction of each batch to represent the class distribution of the stream. On streams whose classes are strongly correlated in time, it therefore needs a small memory buffer, with which it remains the best method (Appendix~\ref{app:additional}). Our analysis builds on contrastive VLMs, and extending it to models trained in other ways is left for future work.

\noindent\textbf{Conclusion.}~~Across contrastive VLMs, we showed that feature compression biases the predictions toward a few classes and that entropy minimization amplifies this bias into collapse. Following this analysis, CAMA keeps the adaptation of entropy minimization and corrects the bias by aligning the predictions with a calibrated reference. We proved that prediction collapse cannot be an optimum of CAMA, and in experiments CAMA prevented the collapse and achieved the best accuracy in all but one backbone--benchmark pair. We hope that this analysis serves as a starting point for designing TTA objectives for VLMs that consider the feature geometry and the class distribution of the predictions together.

\label{cama:references-start}
{\small
\bibliographystyle{ieeenat_fullname}
\bibliography{cama_references}
}

\clearpage
\onecolumn
\begin{center}
{\Large\textbf{\thetitle}}\\[0.6em]
{\large Appendix}
\end{center}
\vspace{1em}

\appendix
\label{cama:supplement-start}
\section{Proofs}\label{app:proofs}

This appendix proves Propositions~\ref{prop:sink} and~\ref{prop:amplify}, Theorem~\ref{thm:anticollapse}, and the stability of the escort to errors in $\bpi$. It also proves the other statements of Sections~\ref{sec:mechanism} and~\ref{sec:method}, namely the bound that compression places on $\Ibatch$ and the collapse thresholds for nonzero $\Ibatch$. It adds the optimal marginal between one-hot and identical predictions and a bound on the support size of the optimal marginal. Throughout this appendix all logarithms are natural, so entropies are in nats, and $z_{ik} = \mathrm{sim}(\mathbf{f}_i, \mathbf{t}_k)/\tau$ denotes the logits.

\subsection{Proof of Proposition~\ref{prop:sink} (Compression Concentrates the Predictions)}\label{app:prop_sink_proof}

\begin{proposition*}[Compression Concentrates the Predictions, restated]
If $\{\hat{\mathbf{f}}_i\}$ is $\varepsilon$-concentrated around $\tilde{\boldsymbol{\mu}}$, then for $k^* = \arg\max_k \hat{\mathbf{t}}_k^\top \tilde{\boldsymbol{\mu}}$ with margin $\rho = \min_{j \neq k^*} (\hat{\mathbf{t}}_{k^*} - \hat{\mathbf{t}}_j)^\top \tilde{\boldsymbol{\mu}}$, provided $\rho > 2\varepsilon$,
\begin{equation*}
\bar{p}_{k^*} \geq 1 - (K-1) \cdot \exp\!\big(\!-(\rho - 2\varepsilon)/\tau\big).
\end{equation*}
\end{proposition*}

\begin{proof}
Since $\hat{\mathbf{f}}_i$ and $\hat{\mathbf{t}}_k$ are $L^2$-normalized and $\|\hat{\mathbf{f}}_i - \tilde{\boldsymbol{\mu}}\| \leq \varepsilon$, Cauchy--Schwarz gives $|\hat{\mathbf{f}}_i^\top \hat{\mathbf{t}}_k - \tilde{\boldsymbol{\mu}}^\top \hat{\mathbf{t}}_k| \leq \varepsilon$ for all $i, k$. For any $j \neq k^*$:
\begin{equation*}
z_{ik^*} - z_{ij} = \frac{\hat{\mathbf{f}}_i^\top(\hat{\mathbf{t}}_{k^*} - \hat{\mathbf{t}}_j)}{\tau} \geq \frac{(\hat{\mathbf{t}}_{k^*} - \hat{\mathbf{t}}_j)^\top \tilde{\boldsymbol{\mu}} - 2\varepsilon}{\tau} \geq \frac{\rho - 2\varepsilon}{\tau}
\end{equation*}
Since $\rho > 2\varepsilon$, the softmax gives
\begin{equation*}
p_{ik^*} = \frac{1}{1 + \sum_{j \neq k^*} \exp(z_{ij} - z_{ik^*})} \geq \frac{1}{1 + (K-1)\exp\!\big(\!-\frac{\rho - 2\varepsilon}{\tau}\big)}
\end{equation*}
Since this holds for every $i$, the batch marginal $\bar{p}_{k^*} = \frac{1}{B}\sum_i p_{ik^*}$ satisfies the same bound. The stated form $\bar{p}_{k^*} \geq 1 - (K-1)\exp\!\big(\!-\frac{\rho-2\varepsilon}{\tau}\big)$ follows from the elementary inequality $\frac{1}{1+x} \geq 1-x$ for $x \geq 0$.
\end{proof}
The bound is strong when $\varepsilon$ is small, when $\rho$ is large, and when $\tau$ is small, as it is in contrastive VLMs. Since it holds for each prediction $p_{ik^*}$, under strong compression every prediction puts nearly all its probability on $k^*$. Under weaker compression the concentration is only partial.

\subsection{Proof of Proposition~\ref{prop:amplify} (Entropy Minimization Amplifies the Bias)}\label{app:entropy_amplification}

\begin{proposition*}[Entropy Minimization Amplifies the Bias, restated]
If every sample of a batch ranks $k^*$ first, the loss of entropy minimization $\mathcal{L}_{\mathrm{ent}}$ bounds the share of $k^*$ from below,
\begin{equation*}
\bar{p}_{k^*} \geq \exp(-\mathcal{L}_{\mathrm{ent}}).
\end{equation*}
Hence $\bar{p}_{k^*} \to 1$ as entropy minimization drives $\mathcal{L}_{\mathrm{ent}} \to 0$, and the batch collapses onto $k^*$.
\end{proposition*}

\begin{proof}
For any $\mathbf{p} \in \Delta^{K-1}$, $\log p_k \leq \log \max_j p_j$ for every $k$, so
\begin{equation*}
H(\mathbf{p}) = -\textstyle\sum_k p_k \log p_k \geq -\log \max_j p_j, \quad \text{that is,} \quad \max_j p_j \geq \exp(-H(\mathbf{p})).
\end{equation*}
Recall that $\mathcal{L}_{\mathrm{ent}} = \frac{1}{B}\sum_i H(\mathbf{p}_i)$. If sample $i$ ranks $k^*$ first, then $p_{ik^*} = \max_j p_{ij} \geq \exp(-H(\mathbf{p}_i))$. Averaging over the batch and applying Jensen's inequality to the convex function $x \mapsto \exp(-x)$ gives
\begin{equation*}
\bar{p}_{k^*} \geq \frac{1}{B}\sum_i \exp(-H(\mathbf{p}_i)) \geq \exp\!\Big(\!-\frac{1}{B}\sum_i H(\mathbf{p}_i)\Big) = \exp(-\mathcal{L}_{\mathrm{ent}}).
\end{equation*}
Since $\bar{p}_{k^*} \leq 1$, the bound forces $\bar{p}_{k^*} \to 1$ as $\mathcal{L}_{\mathrm{ent}} \to 0$.
\end{proof}
The bound is tight at both ends. It equals one for one-hot predictions and $1/K$ for uniform predictions.

The condition of the proposition persists under entropy minimization, because the entropy gradient with respect to the logits favors the top class.

\begin{lemma}[Entropy Descent Preserves the Top Class]\label{lem:topclass}
Let $\mathbf{p}$ be the softmax of logits $z \in \mathbb{R}^K$, and let $k^*$ be a class with the largest logit. Then
\begin{equation*}
-\frac{\partial H(\mathbf{p})}{\partial z_k} = p_k\big(\log p_k + H(\mathbf{p})\big),
\end{equation*}
and this descent direction is largest at $k^*$, strictly larger than at every class $j$ with $p_j < p_{k^*}$.
\end{lemma}

\begin{proof}
With $\partial p_j / \partial z_k = p_j(\mathbf{1}[j = k] - p_k)$,
\begin{equation*}
\frac{\partial H}{\partial z_k} = -\textstyle\sum_j (\log p_j + 1)\, p_j\,(\mathbf{1}[j = k] - p_k) = -p_k\big(\log p_k + H(\mathbf{p})\big).
\end{equation*}
Write $g_k = p_k(\log p_k + H(\mathbf{p}))$. Since $H(\mathbf{p}) \geq -\log p_{k^*}$, with equality only when $\mathbf{p}$ is uniform, $g_{k^*} \geq 0$. Take any $j$ with $p_j < p_{k^*}$, so that $\mathbf{p}$ is not uniform and $g_{k^*} > 0$. If $\log p_j + H(\mathbf{p}) \leq 0$, then $g_j \leq 0 < g_{k^*}$. Otherwise both factors of $g_j$ are positive and smaller than those of $g_{k^*}$, so $g_j < g_{k^*}$.
\end{proof}
A gradient step that lowers the entropy on the logits therefore widens every gap $z_{k^*} - z_j$, so $k^*$ remains the top class of the sample and $p_{k^*}$ grows. Applied to every sample of a batch that shares the top class $k^*$, the condition of Proposition~\ref{prop:amplify} continues to hold while $\mathcal{L}_{\mathrm{ent}}$ decreases.

\subsection{Decomposition and Loss Equivalence}\label{app:decomp_proof}

This subsection derives the decomposition used in the proofs below and then verifies that Eq.~\eqref{eq:lossB} and Eq.~\eqref{eq:solver} differ only by a constant, as Section~\ref{sec:method} states. Substituting $\frac{1}{B}\sum_i H(\mathbf{p}_i) = H(\pbar) - \Ibatch$ in Eq.~\eqref{eq:solver} gives the decomposition
\begin{equation}\label{eq:decomp}
\begin{aligned}
\mathcal{L}_{\mathrm{solver}} ={}& (1-\lambda)\, H(\pbar) \\
&{}- \lambda\, \langle \pbar, \log \bpi \rangle - \Ibatch.
\end{aligned}
\end{equation}

\begin{proof}
Expanding Eq.~\eqref{eq:solver} with $\Ibatch = H(\pbar) - \frac{1}{B}\sum_i H(\mathbf{p}_i)$ gives
\begin{equation*}
\mathcal{L}_{\mathrm{solver}} = \frac{1}{B}\textstyle\sum_i H(\mathbf{p}_i) + \lambda\,\KL(\pbar\|\bpi) = \big[H(\pbar) - \Ibatch\big] - \lambda\,H(\pbar) - \lambda\langle\pbar,\log\bpi\rangle
\end{equation*}
which simplifies to $(1-\lambda)\,H(\pbar) - \lambda\,\langle\pbar,\log\bpi\rangle - \Ibatch$, proving the decomposition.

\end{proof}

\textbf{Corollary (Loss Equivalence).}~~For $\lambda > 1$, Eq.~\eqref{eq:lossB} and Eq.~\eqref{eq:solver} differ by a constant and have the same gradient. \begin{proof} By Eq.~\eqref{eq:pdag}, $\log p^\dagger_k = \frac{\lambda}{\lambda-1}\log\pi_k - \log Z$, so $(\lambda-1)\KL(\pbar\|\pdag) = -(\lambda-1)H(\pbar) - \lambda\langle\pbar,\log\bpi\rangle + (\lambda-1)\log Z$. Hence
\begin{equation*}
-\Ibatch + (\lambda-1)\KL(\pbar\|\pdag) = \mathcal{L}_{\mathrm{solver}} + (\lambda-1)\log Z
\end{equation*}
Since $\bpi$ and $\pdag$ are held fixed, $Z = \sum_j \pi_j^{\lambda/(\lambda-1)}$ does not depend on $\theta$, and the two forms have the same gradient, $\nabla_\theta\mathcal{L} = \nabla_\theta\mathcal{L}_{\mathrm{solver}}$.
\end{proof}

\subsection{Proof of Theorem~\ref{thm:anticollapse} (Alignment Prevents the Collapse)}

\begin{theorem*}[Alignment Prevents the Collapse, restated]
Under Assumption~\ref{asm:a1}, for $\bpi \in \Delta^{K-1}_\circ$, any $\lambda > 1$, and $\mathbf{p} \in \Delta^{K-1}$:
\begin{equation*}
\mathcal{L}_\lambda(\mathbf{p}) = (\lambda - 1)\,\KL(\mathbf{p} \,\|\, \pdag) + C,
\end{equation*}
where $\pdag$ is the escort of Eq.~\eqref{eq:pdag} and the constant $C$ does not depend on $\mathbf{p}$. The unique minimizer is therefore $\pdag$. For $0 \leq \lambda < 1$, which includes entropy minimization at $\lambda = 0$, every minimizer puts all its mass on one class, and at $\lambda = 1$, on the classes with the largest $\pi_k$.
\end{theorem*}

The proof gives the constant explicitly. With $\alpha = \lambda/(\lambda-1)$, the escort is $\pdag = \mathcal{E}_\alpha(\bpi)$ with $\mathcal{E}_\alpha(\bpi)_k = \pi_k^\alpha / \sum_j \pi_j^\alpha$, and $C = H_\alpha(\bpi) = \frac{1}{1-\alpha}\log\sum_j \pi_j^\alpha$, the R\'enyi entropy of $\bpi$, which is also the minimum value of $\mathcal{L}_\lambda$.

\begin{proof}
Under Assumption~\ref{asm:a1}, $\pbar = \mathbf{p}$ and $\Ibatch = 0$. From Eq.~\eqref{eq:decomp}:
\begin{equation*}
\mathcal{L}_\lambda(\mathbf{p}) = \textstyle\sum_k p_k\big[(\lambda-1)\log p_k - \lambda \log \pi_k\big]
\end{equation*}
Let $\alpha = \lambda/(\lambda-1)$ and $Z_\alpha = \sum_j \pi_j^\alpha$. Since $\mathcal{E}_\alpha(\bpi)_k = \pi_k^\alpha / Z_\alpha$, we have $\log \mathcal{E}_\alpha(\bpi)_k = \alpha \log \pi_k - \log Z_\alpha$, hence $\lambda \log \pi_k = (\lambda-1)\log \mathcal{E}_\alpha(\bpi)_k + (\lambda-1)\log Z_\alpha$. Substituting:
\begin{align}
\mathcal{L}_\lambda(\mathbf{p}) &= (\lambda-1)\sum_k p_k \log \frac{p_k}{\mathcal{E}_\alpha(\bpi)_k} - (\lambda-1)\log Z_\alpha \notag \\
&= (\lambda-1)\,\KL(\mathbf{p} \| \mathcal{E}_\alpha(\bpi)) + H_\alpha(\bpi)
\end{align}
where $H_\alpha(\bpi) = \frac{1}{1-\alpha}\log Z_\alpha = -(\lambda-1)\log Z_\alpha$ (using $1-\alpha = -1/(\lambda-1)$). For $\lambda > 1$, $(\lambda-1) > 0$ and $\KL \geq 0$, so $\mathcal{L}_\lambda(\mathbf{p}) \geq H_\alpha(\bpi)$ with equality iff $\mathbf{p} = \mathcal{E}_\alpha(\bpi)$. For $0 \leq \lambda < 1$, $(\lambda-1) < 0$ reverses the inequality, so $\mathcal{L}_\lambda$ is concave on $\Delta^{K-1}_\circ$ and minimized at vertices. At $\lambda = 1$, $\mathcal{L}_1 = -\langle \mathbf{p}, \log\bpi\rangle$ is linear, minimized on $\arg\max_k \pi_k$. At $\lambda = 0$, $\alpha = 0$ and $\mathcal{E}_0(\bpi)$ is uniform, so $\mathcal{L}_0(\mathbf{p}) = H(\mathbf{p})$, the entropy-minimization objective, and the same conclusion holds.
\end{proof}

\subsection{Proof of Corollary~\ref{cor:collapse} (Concentration Near \texorpdfstring{$\lambda = 1$}{lambda = 1})}

\begin{corollary}[Concentration Near $\lambda = 1$]\label{cor:collapse}
Let $k_\pi = \arg\max_k \pi_k$ (assumed unique) and $\Delta_\pi = \log(\pi_{k_\pi}/\pi_{(2)})$. Then
$\TV(\pdag, \boldsymbol{\delta}_{k_\pi}) \leq (K-1)\exp(-\alpha\,\Delta_\pi)$.
As $\lambda \downarrow 1$, so that $\alpha \to \infty$, $\pdag$ concentrates on $k_\pi$ exponentially fast and collapse returns. The boundary $\lambda = 1$ is sharp. For $\lambda > 1$ the minimizer is interior, and at $\lambda = 1$ it is a vertex.
\end{corollary}

\begin{proof}
For $k \neq k_\pi$: $\mathcal{E}_\alpha(\bpi)_k / \mathcal{E}_\alpha(\bpi)_{k_\pi} = (\pi_k/\pi_{k_\pi})^\alpha \leq \exp(-\alpha\,\Delta_\pi)$. Hence $\sum_{k \neq k_\pi} \mathcal{E}_\alpha(\bpi)_k \leq (K-1)\exp(-\alpha\,\Delta_\pi)$. Since $\TV(\pdag, \boldsymbol{\delta}_{k_\pi}) = \sum_{k \neq k_\pi} p^\dagger_k$, the bound follows.
\end{proof}

\subsection{Softmax Contraction Lemma}

Proposition~\ref{prop:robustness} and Lemma~\ref{lem:ibatch_bound} both use the following bound on how far the softmax output moves when the logits move.

\begin{lemma}[Softmax contraction from $\ell_\infty$ to total variation]\label{lem:softmax}
For the standard softmax $\sigma: \mathbb{R}^K \to \Delta^{K-1}$,
\begin{equation*}
\TV(\sigma(z), \sigma(z')) \leq \frac{1}{2}\|z - z'\|_\infty \qquad \forall\, z, z' \in \mathbb{R}^K.
\end{equation*}
\end{lemma}

\begin{proof}
Let $s = \sigma(z)$. The Jacobian is $J_\sigma(z) = \mathrm{Diag}(s) - ss^\top$. For any $u \in \mathbb{R}^K$, $J_\sigma(z)u = s \odot (u - \langle s, u \rangle \mathbf{1})$, so $\|J_\sigma(z)u\|_1 = \sum_i s_i |u_i - \bar{u}|$ where $\bar{u} = \langle s, u \rangle$. Since $\bar{u}$ lies in $[\min_i u_i, \max_i u_i]$, the mean absolute deviation satisfies $\sum_i s_i |u_i - \bar{u}| \leq \|u\|_\infty$. Hence $\|J_\sigma(z)\|_{\infty \to 1} \leq 1$ for all $z$. Integrating along the segment from $z$ to $z'$: $\|\sigma(z) - \sigma(z')\|_1 \leq \|z - z'\|_\infty$. Dividing by $2$ gives the TV bound.
\end{proof}

\subsection{Escort Robustness}

Calibration estimates $\bpi$ from the first $N_{\mathrm{cal}}$ batches, so the estimate carries finite-sample error. The proposition below shows that this error moves the escort only at first order and raises the loss only at second order.

\begin{proposition}[Escort Robustness]\label{prop:robustness}
Let $\hat{\bpi}$ be an estimate of $\bpi$ with $\|\hat{\bpi} - \bpi\|_\infty \leq \eta < \mu := \min_k \pi_k$. Then (i) $\TV(\mathcal{E}_\alpha(\hat{\bpi}), \mathcal{E}_\alpha(\bpi)) \leq \alpha\eta / [2(\mu - \eta)]$, and (ii) the excess loss satisfies $\mathcal{L}_\lambda(\mathcal{E}_\alpha(\hat{\bpi}); \bpi) - \mathcal{L}_\lambda(\mathcal{E}_\alpha(\bpi); \bpi) \leq \frac{\lambda^2}{2(\lambda-1)} \cdot \frac{\eta^2}{(\mu - \eta)^2}$.
\end{proposition}

\begin{proof}
\textbf{Part (i).}~~The escort can be written as $\mathcal{E}_\alpha(\bpi) = \sigma(\alpha \log \bpi)$. By Lemma~\ref{lem:softmax}:
\begin{equation*}
\TV(\mathcal{E}_\alpha(\hat{\bpi}), \mathcal{E}_\alpha(\bpi)) \leq \frac{\alpha}{2}\|\log \hat{\bpi} - \log \bpi\|_\infty
\end{equation*}
Since $\|\hat{\bpi} - \bpi\|_\infty \leq \eta$ and $\pi_k \geq \mu > \eta$, the mean value theorem gives $|\log \hat{\pi}_k - \log \pi_k| \leq |\hat{\pi}_k - \pi_k|/(\pi_k - \eta) \leq \eta/(\mu - \eta)$. Combining yields the bound.

\textbf{Part (ii).}~~By Theorem~\ref{thm:anticollapse},
\begin{equation*}
\mathcal{L}_\lambda(\mathcal{E}_\alpha(\hat{\bpi});\bpi) - \mathcal{L}_\lambda(\mathcal{E}_\alpha(\bpi);\bpi) = (\lambda-1)\,\KL(\mathcal{E}_\alpha(\hat{\bpi}) \| \mathcal{E}_\alpha(\bpi))
\end{equation*}
Setting $z = \alpha\log\bpi$ and $\hat{z} = \alpha\log\hat{\bpi}$, we have $\mathcal{E}_\alpha(\bpi) = \sigma(z)$ and $\mathcal{E}_\alpha(\hat{\bpi}) = \sigma(\hat{z})$. The KL equals the Bregman divergence of the log-sum-exp function $A(u) = \log\sum_j e^{u_j}$:
\begin{equation*}
\KL(\sigma(\hat{z}) \| \sigma(z)) = B_A(z, \hat{z}) = \int_0^1 (1-t)\,\Delta^\top \nabla^2 A(\hat{z} + t\Delta)\,\Delta\,dt
\end{equation*}
where $\Delta = z - \hat{z}$. Since $\nabla^2 A(u) = \mathrm{Diag}(\sigma(u)) - \sigma(u)\sigma(u)^\top$, we have $\Delta^\top \nabla^2 A(u)\,\Delta = \mathrm{Var}_{\sigma(u)}(\Delta) \leq \|\Delta\|_\infty^2$. Integrating: $\KL \leq \frac{1}{2}\|\Delta\|_\infty^2 = \frac{\alpha^2}{2}\|\log\hat{\bpi} - \log\bpi\|_\infty^2 \leq \frac{\alpha^2\eta^2}{2(\mu-\eta)^2}$. Substituting $\alpha = \lambda/(\lambda-1)$:
\begin{equation*}
(\lambda-1) \cdot \frac{\alpha^2\eta^2}{2(\mu-\eta)^2} = \frac{\lambda^2}{2(\lambda-1)} \cdot \frac{\eta^2}{(\mu-\eta)^2} \qedhere
\end{equation*}
\end{proof}

\subsection{Sample Complexity}\label{app:sample_complexity}

The corollary below applies a standard concentration bound to Proposition~\ref{prop:robustness}. It is stated for a plain average of soft predictions rather than for the rank-based reference of Eq.~\eqref{eq:pi_def}.

\begin{corollary}[Generic plug-in bound]
Let $\hat{\bpi} = \frac{1}{n}\sum_{i=1}^n \mathbf{p}_i$ with $\mathbf{p}_i \in \Delta^{K-1}$ i.i.d.\ and $\mathbb{E}[\mathbf{p}_i] = \bpi$. By coordinatewise Hoeffding and a union bound, $\|\hat{\bpi} - \bpi\|_\infty \leq \sqrt{\log(2K/\delta)/(2n)}$ with probability at least $1-\delta$. If this quantity is below $\mu = \min_k \pi_k$, Proposition~\ref{prop:robustness} applies with $\eta = \sqrt{\log(2K/\delta)/(2n)}$.
\end{corollary}

\textbf{Remark on Large $K$.}~~At large $K$ the reference is kept robust by the tempering of Eq.~\eqref{eq:pi_def} rather than by this bound. The bound is a union bound over the $K$ coordinates, and for $K = 1000$, $n = N_{\mathrm{cal}}\cdot B = 16\cdot 64$, and $\delta = 0.05$ it gives $\eta \approx 0.071$. Harmonic weights normalized per sample give $s_k \in [1/(K H_K),\, 1/H_K]$, with $H_K$ the $K$-th harmonic number. Hence, for every calibration outcome, $\max_{k,j} |\log \pi_k - \log \pi_j| \leq \kappa \log \frac{1/H_K + \zeta}{1/(K H_K) + \zeta}$, which is $0.13$ nats at $K = 1000$ and at most $0.16$ nats for $K \geq 10$. The reference enters Eq.~\eqref{eq:solver} only through $\lambda \langle \pbar, \log \bpi \rangle$, so this term varies by at most $0.16\lambda$ nats, and the reference cannot become a peaked target for any calibration sample. Empirically the estimated reference matches the oracle on all three benchmarks, as Table~\ref{tab:reference} shows, and CAMA does not collapse on any of the $15$ corruptions at $K=1000$, as Section~\ref{sec:analysis} reports.

\subsection{Compression Bounds the Batch Mutual Information}\label{app:compression_a1}

We show that $\varepsilon$-concentration, as defined in Proposition~\ref{prop:sink}, bounds $\Ibatch$ from above. Assumption~\ref{asm:a1} is therefore the limit of compression, not a separate assumption.

\begin{lemma}[Compression Bounds $\Ibatch$]\label{lem:ibatch_bound}
If $\{\hat{\mathbf{f}}_i\}_{i=1}^B$ is $\varepsilon$-concentrated around $\tilde{\boldsymbol{\mu}}$, then:
\begin{equation*}
\Ibatch \;\leq\; \tfrac{\delta}{2}\log(K-1) + h_2(\tfrac{\delta}{2}), \qquad \delta = \frac{2\varepsilon}{\tau}
\end{equation*}
where $h_2(x) = -x\log x - (1-x)\log(1-x)$ is the binary entropy function, provided $\delta \leq 1$.
\end{lemma}

\begin{proof}
Since $\hat{\mathbf{f}}_i, \hat{\mathbf{t}}_k$ are $L^2$-normalized and $\|\hat{\mathbf{f}}_i - \tilde{\boldsymbol{\mu}}\| \leq \varepsilon$, for any two samples $i, j$ and any class $k$:
\begin{equation*}
|z_{ik} - z_{jk}| = \frac{|(\hat{\mathbf{f}}_i - \hat{\mathbf{f}}_j)^\top \hat{\mathbf{t}}_k|}{\tau} \leq \frac{2\varepsilon}{\tau} = \delta
\end{equation*}
using the triangle inequality $\|\hat{\mathbf{f}}_i - \hat{\mathbf{f}}_j\| \leq 2\varepsilon$. Hence $\|z_i - z_j\|_\infty \leq \delta$.

By Lemma~\ref{lem:softmax}, $\|\mathbf{p}_i - \mathbf{p}_j\|_1 \leq \|z_i - z_j\|_\infty \leq \delta$. Since $\pbar = \frac{1}{B}\sum_i \mathbf{p}_i$, convexity gives $\|\mathbf{p}_i - \pbar\|_1 \leq \delta$ for all $i$.

By the Fannes--Audenaert inequality~\citep{audenaert2007sharp}, for distributions $\mathbf{p}, \mathbf{q}$ on $K$ outcomes with $\|\mathbf{p} - \mathbf{q}\|_1 \leq 2t \leq 1$:
$|H(\mathbf{p}) - H(\mathbf{q})| \leq t\log(K-1) + h_2(t)$.

Applying this with $t = \delta/2$:
\begin{equation*}
|H(\mathbf{p}_i) - H(\pbar)| \leq \tfrac{\delta}{2}\log(K-1) + h_2(\tfrac{\delta}{2}) \quad \text{for all } i.
\end{equation*}
Since $\Ibatch = H(\pbar) - \frac{1}{B}\sum_i H(\mathbf{p}_i) \leq \max_i |H(\mathbf{p}_i) - H(\pbar)|$, the result follows.
\end{proof}

\textbf{Remark.}~~Proposition~\ref{prop:sink} concentrates the predictions on $k^*$ once $\Delta > 2\varepsilon$, and Lemma~\ref{lem:ibatch_bound} gives $\Ibatch \to 0$ as $\delta = 2\varepsilon/\tau \to 0$. Both follow from the same compression. Where collapse is most dangerous, the batch therefore approaches Assumption~\ref{asm:a1}, the setting of Theorem~\ref{thm:anticollapse}.

\subsection{Collapse Thresholds for Nonzero \texorpdfstring{$\Ibatch$}{I-batch}}

Theorem~\ref{thm:anticollapse} assumes identical predictions, $\Ibatch = 0$. We now drop this assumption and show that collapse stays suboptimal once $\Ibatch$ exceeds a threshold set by the reference, the guarantee away from the limit of Section~\ref{sec:theory}. The identity of Theorem~\ref{thm:anticollapse}, $\mathcal{L}_\lambda(\pbar) = H_\alpha(\bpi) + (\lambda-1)\KL(\pbar\|\pdag)$, is algebraic and holds for any $\pbar \in \Delta^{K-1}$ and any $\lambda \neq 1$, with or without Assumption~\ref{asm:a1}. With Eq.~\eqref{eq:decomp}, $\mathcal{L}_{\mathrm{solver}} = \mathcal{L}_\lambda(\pbar) - \Ibatch$, this gives the following identity.

\begin{corollary}[Identity for Any $\Ibatch$]\label{cor:finite_ibatch}
For any predictions $\{\mathbf{p}_i\}$:
$\mathcal{L}_{\mathrm{solver}} = \mathcal{L}_\lambda(\pdag) + (\lambda-1)\KL(\pbar\|\pdag) - \Ibatch$.
\end{corollary}

A batch whose predictions all equal $\pdag$ has $\Ibatch = 0$ and loss $\mathcal{L}_\lambda(\pdag)$. Hence at any minimizer $\mathcal{L}^*_{\mathrm{solver}} \leq \mathcal{L}_\lambda(\pdag)$, so $(\lambda-1)\KL(\pbar^*\|\pdag) \leq \Ibatch^*$. Thus $\pbar$ can move away from $\pdag$ only as far as $\Ibatch$ allows.

\begin{proposition}[Collapse Threshold for a Uniform Reference]\label{prop:threshold}
For uniform $\bpi = \mathbf{1}/K$, collapse is suboptimal whenever $\lambda > 1 - \Ibatch / \log K$.
\end{proposition}

\begin{proof}
A collapsed batch, $\pbar = \boldsymbol{\delta}_k$, has $\Ibatch = 0$ and loss $\mathcal{L}_\lambda(\boldsymbol{\delta}_k) = \lambda\log K$, since $\KL(\boldsymbol{\delta}_k\|\mathbf{1}/K) = \log K$. It suffices to find a batch with lower loss. Consider predictions $\{\mathbf{p}_i\}$ with $\Ibatch > 0$. By Corollary~\ref{cor:finite_ibatch}:
\begin{equation*}
\mathcal{L}_{\mathrm{solver}} = \mathcal{L}_\lambda(\pdag) + (\lambda-1)\KL(\pbar\|\pdag) - \Ibatch
\end{equation*}
For uniform $\bpi = \mathbf{1}/K$ and any $\lambda \neq 1$, $\pdag = \mathbf{1}/K$, so $\mathcal{L}_\lambda(\pdag) = \log K$ and $\KL(\pbar\|\pdag) \leq \log K$. For $\lambda > 1$ this gives $\mathcal{L}_{\mathrm{solver}} \leq \log K + (\lambda-1)\log K - \Ibatch = \lambda\log K - \Ibatch$. Collapse is then suboptimal whenever $\lambda\log K - \Ibatch < \lambda\log K$, that is, whenever $\Ibatch > 0$. For $\lambda < 1$, $\mathcal{L}_\lambda(\pdag) = \log K$ and $\KL(\pbar\|\pdag) \geq 0$, so $\mathcal{L}_{\mathrm{solver}} \leq \log K - \Ibatch$. Then $\log K - \Ibatch < \lambda\log K$ gives $\lambda > 1 - \Ibatch/\log K$.
\end{proof}

The next proposition covers any reference for $\lambda \geq 1$, with a threshold on $\Ibatch$ instead of on $\lambda$. \begin{proposition}[Collapse Threshold for Any Reference]\label{prop:threshold_general}
For any $\bpi \in \Delta^{K-1}_\circ$ and any $\lambda \geq 1$, every batch with $\Ibatch > \log(\max_k \pi_k / \min_k \pi_k)$ has lower loss than every collapsed batch. For a uniform reference the threshold is zero.
\end{proposition}

\begin{proof}
A collapsed batch has $\pbar = \boldsymbol{\delta}_k$ and $\Ibatch = 0$, so its loss is $\mathcal{L}_\lambda(\boldsymbol{\delta}_k) = -\lambda\log\pi_k$, smallest at $k_\pi = \arg\max_k \pi_k$. Let $D = \sum_k \bar p_k \log(\pi_{k_\pi}/\pi_k)$, so that $0 \leq D \leq \log(\max_k \pi_k / \min_k \pi_k)$. For any batch,
\begin{equation*}
\mathcal{L}_{\mathrm{solver}} + \lambda\log\pi_{k_\pi} = \lambda D - \Ibatch - (\lambda-1)H(\pbar) \leq \lambda(D - \Ibatch),
\end{equation*}
because $H(\pbar) \geq \Ibatch$ and $\lambda \geq 1$. The right side is negative whenever $\Ibatch > \log(\max_k \pi_k / \min_k \pi_k)$. The loss of Eq.~\eqref{eq:lossB} differs from $\mathcal{L}_{\mathrm{solver}}$ by a constant for $\lambda > 1$, as Appendix~\ref{app:decomp_proof} verifies, and equals $-\Ibatch$ at $\lambda = 1$, where any $\Ibatch > 0$ suffices.
\end{proof}

For the reference of Eq.~\eqref{eq:pi_def}, each normalized harmonic weight lies in $[1/(KH_K), 1/H_K]$ with $H_K = \sum_{r=1}^{K} 1/r$, and so does each score $s_k$. By the Remark on Large $K$ in Appendix~\ref{app:sample_complexity}, the threshold $\log(\max_k \pi_k / \min_k \pi_k)$ of Proposition~\ref{prop:threshold_general} is therefore at most $\kappa \log\frac{\zeta + 1/H_K}{\zeta + 1/(KH_K)}$, which is $0.15$, $0.15$, and $0.13$ for $K = 10$, $100$, and $1000$ at the default $\zeta$ and $\kappa$. The tempering thus keeps the guarantee close to that of a uniform reference.

\subsection{Lower Bound on the Support Size}

Theorem~\ref{thm:anticollapse} rules out collapse onto a single class. The corollary below also bounds the share of the target mass that the support of an optimal marginal must cover, which rules out marginals spread over too few classes when the target is spread over many.

\begin{corollary}[Support-Size Lower Bound]\label{cor:support}
For $\lambda > 1$, let $\{\mathbf{p}_i^*\}$ be any minimizer of $\mathcal{L}_{\mathrm{solver}}$ with batch marginal $\pbar^*$ and mutual information $\Ibatch^*$. Then:
\begin{equation*}
\|\pbar^* - \pdag\|_1 \;\leq\; \sqrt{\frac{2\,\Ibatch^*}{\lambda - 1}}
\end{equation*}
Moreover, if $\pbar^*$ has support $S \subseteq [K]$, then
\begin{equation*}
\sum_{k \in S} p^\dagger_k \;\geq\; \exp\!\Big(-\frac{\Ibatch^*}{\lambda-1}\Big)
\end{equation*}
For uniform $\pdag = \mathbf{1}/K$, this yields a support-size lower bound: $|S| \geq K \cdot \exp\!\big(-\Ibatch^*/(\lambda-1)\big)$.
\end{corollary}

\begin{proof}
From Corollary~\ref{cor:finite_ibatch}, at any minimizer $(\lambda-1)\KL(\pbar^*\|\pdag) \leq \Ibatch^*$. The $\ell_1$ bound follows from Pinsker's inequality, $\|\pbar^* - \pdag\|_1 \leq \sqrt{2\KL(\pbar^*\|\pdag)}$.

For the support bound, if $\mathrm{supp}(\pbar^*) \subseteq S$, then by Jensen's inequality:
\begin{equation*}
\KL(\pbar^*\|\pdag) = \sum_{k\in S} \bar{p}^*_k \log\frac{\bar{p}^*_k}{p^\dagger_k} \geq -\log\sum_{k\in S} p^\dagger_k
\end{equation*}
Combining this with $(\lambda-1)\KL(\pbar^*\|\pdag) \leq \Ibatch^*$ and exponentiating gives
$\sum_{k\in S} p^\dagger_k \geq \exp\!\big(-\Ibatch^*/(\lambda-1)\big)$.
For uniform $\pdag$, $\sum_{k\in S} p^\dagger_k = |S|/K$, giving $|S| \geq K \cdot \exp\!\big(-\Ibatch^*/(\lambda-1)\big)$.
\end{proof}

\textbf{Remark.}~~For uniform $\pdag = \mathbf{1}/K$ and $\Ibatch^* = 0$, the limit of Assumption~\ref{asm:a1}, the bound requires $\sum_{k\in S} p^\dagger_k \geq 1$, that is, $S = [K]$. The marginal must then cover every class, as in Theorem~\ref{thm:anticollapse}. For $\Ibatch^* > 0$ it may cover fewer classes, but not arbitrarily few.

\subsection{Between One-Hot and Identical Predictions}\label{app:eta_interp}

Theorem~\ref{thm:anticollapse} covers batches with $\Ibatch = 0$. The corollary below gives the optimal marginal for every batch, from one-hot predictions to identical ones.

\begin{corollary}[$\xi$-Interpolation]\label{cor:eta}
For a batch, define $\xi \in [0,1]$ by
\begin{equation*}
\frac{1}{B}\sum_{i=1}^B H(\mathbf{p}_i) = \xi\, H(\pbar)
\end{equation*}
Concavity of entropy gives $\xi \leq 1$, and $\xi$ is defined whenever $H(\pbar) > 0$, that is, whenever the predictions have not fully collapsed. Treating $\xi$ as fixed, the solver loss restricted to the marginal becomes
\begin{equation*}
\mathcal{L}_{\mathrm{solver}}\big|_\xi = (\xi - \lambda)\, H(\pbar) - \lambda\, \langle \pbar, \log\bpi \rangle
\end{equation*}
For $\lambda > \xi$, this is strictly convex on $\Delta^{K-1}$ with unique minimizer
\begin{equation*}
q^{(\xi)}_k = \frac{\pi_k^{\,\lambda/(\lambda-\xi)}}{\sum_j \pi_j^{\,\lambda/(\lambda-\xi)}}
\end{equation*}
\end{corollary}

\begin{proof}
From Eq.~\eqref{eq:decomp}, $\mathcal{L}_{\mathrm{solver}} = (1-\lambda)H(\pbar) - \lambda\langle\pbar,\log\bpi\rangle - \Ibatch$. Substituting $\Ibatch = H(\pbar) - \xi\, H(\pbar) = (1-\xi)H(\pbar)$:
\begin{equation*}
\mathcal{L}_{\mathrm{solver}} = (1-\lambda)H(\pbar) - \lambda\langle\pbar,\log\bpi\rangle - (1-\xi)H(\pbar) = (\xi-\lambda)H(\pbar) - \lambda\langle\pbar,\log\bpi\rangle.
\end{equation*}
For $\lambda > \xi$, the coefficient of $H(\pbar)$ is negative, making the objective strictly convex. Setting the gradient of the Lagrangian to zero gives $q_k \propto \pi_k^{\lambda/(\lambda-\xi)}$.
\end{proof}

\textbf{Remark.}~~In escort notation, $q^{(\xi)} = \mathcal{E}_{\lambda/(\lambda-\xi)}(\bpi)$. At $\xi = 1$, the case of Assumption~\ref{asm:a1}, $q^{(1)} = \pdag$ as in Theorem~\ref{thm:anticollapse}. At $\xi = 0$, where every $\mathbf{p}_i$ is one-hot, $q^{(0)}_k = \pi_k^{\lambda/\lambda}/Z = \pi_k$, so the optimal marginal is the calibrated reference itself. As corruption grows and $\xi$ moves from $0$ to $1$, the optimal marginal moves continuously from $\bpi$ to $\pdag$.

\section{Calibration Rule for \texorpdfstring{$\lambda$}{lambda}}\label{app:calibration}

\subsection{Gradient-Ratio Rule for \texorpdfstring{$\lambda_0$}{lambda0}}\label{app:lambda0_derivation}

Let $\mathbf{g}_E$ and $\mathbf{g}_K$ be the gradients of the entropy term and the KL term of Eq.~\eqref{eq:solver}. The descent direction $\mathbf{d} = -(\mathbf{g}_E + \lambda\,\mathbf{g}_K)$ decreases both the entropy term and the KL term if and only if $\mathbf{g}_E^\top \mathbf{d} < 0$ and $\mathbf{g}_K^\top \mathbf{d} < 0$. When the two gradients conflict, $\langle \mathbf{g}_E, \mathbf{g}_K \rangle < 0$, these conditions give the interval below, which is nonempty unless the two gradients point in exactly opposite directions.
\begin{equation*}
\frac{|\langle \mathbf{g}_E, \mathbf{g}_K \rangle|}{\|\mathbf{g}_K\|^2} < \lambda < \frac{\|\mathbf{g}_E\|^2}{|\langle \mathbf{g}_E, \mathbf{g}_K \rangle|}
\end{equation*}
We set $\lambda_0 = \|\mathbf{g}_E\|/\|\mathbf{g}_K\|$, the gradient-norm ratio, which lies inside this interval and equals the geometric mean of its endpoints. This standard choice balances the scales of the two terms and needs only the two gradient norms. In practice the two norms are computed per calibration batch with the batch-local marginal $\pbar_b$, and we sum each numerator and denominator over the $N_{\mathrm{cal}}$ batches before forming the ratio.

\subsection{Finite-Batch Scaling of \texorpdfstring{$\lambda$}{lambda}}\label{app:scaling}

This subsection derives the shrinkage factor $B/(B+K-1)$ of Eq.~\eqref{eq:lambda}. The batch marginal $\pbar = \frac{1}{B}\sum_i \mathbf{p}_i$ estimates the expected marginal $\boldsymbol{\mu} = \mathbb{E}[\mathbf{p}_i]$ of the model from $B$ samples. Even when $\boldsymbol{\mu} = \bpi$, the KL computed from $\pbar$ is biased upward. A second-order Taylor expansion of $\KL(\pbar \| \bpi)$ around $\boldsymbol{\mu}$ gives:
\begin{equation*}
\mathbb{E}[\KL(\pbar \| \bpi)] = \KL(\boldsymbol{\mu} \| \bpi) + \frac{d_{\mathrm{eff}}}{2B} + o(B^{-1}),
\qquad d_{\mathrm{eff}} := \mathrm{tr}\!\big(\mathrm{diag}(\boldsymbol{\mu})^{-1} \Sigma\big)
\end{equation*}
where $\Sigma = \mathrm{Cov}(\mathbf{p}_i)$. For one-hot predictions $d_{\mathrm{eff}} = K - 1$, which gives the classical Miller--Madow bias $(K{-}1)/(2B)$. For soft predictions $d_{\mathrm{eff}} \leq K - 1$.

This bias means that part of the gradient $\nabla_\theta [\lambda \cdot \KL(\pbar \| \bpi)]$ comes from sampling noise rather than from the true KL. A standard fix is Bayesian smoothing, which replaces $\pbar$ with the Dirichlet posterior mean $\pbar_{\mathrm{smooth}} = (B \cdot \pbar + A \cdot \bpi) / (B + A)$, with $A$ pseudo-counts placed at $\bpi$. To first order, this replacement scales the gradient signal of the KL term, $\log(\pbar/\bpi)$, by $\frac{B}{B+A}$, and the shrinkage below applies this factor to the alignment weight.

Setting $A = d_{\mathrm{eff}}$ matches the number of pseudo-counts to the noise level. The exact $d_{\mathrm{eff}}$ depends on the unknown $\boldsymbol{\mu}$ and on the covariance of the predictions. Both would have to be estimated from the calibration batches, that is, from the same $N_{\mathrm{cal}} \cdot B$ samples whose noise the shrinkage is meant to absorb. The plug-in estimate $\mathrm{tr}(\mathrm{diag}(\bpi)^{-1}\hat{\Sigma}_{\mathrm{cal}})$ also depends on the prediction distribution. Under soft predictions it is smaller, so the weight is shrunk less, even where a 64-sample batch cannot resolve a 1000-class marginal. We therefore use the worst-case value $d_{\mathrm{eff}} = K - 1$, which bounds the plug-in estimate from above for every prediction distribution and needs no covariance estimate. Since the alignment term of Eq.~\eqref{eq:lossB} carries the weight $\lambda - 1$, we apply the shrinkage factor to $\lambda - 1$,
\begin{equation*}
\lambda = 1 + (\lambda_0 - 1) \cdot \frac{B}{B + K - 1}.
\end{equation*}
The factor is $0.96$ on CIFAR-10-C, where $B = 200$ and $K = 10$, $0.67$ on CIFAR-100-C, where $K = 100$, and $0.06$ on ImageNet-C, where $B = 64$ and $K = 1000$. The weight thus stays close to $\lambda_0$ when a batch resolves the class marginal and is scaled down when it cannot. Sampling noise in the batch marginal therefore does not drive the adaptation.

\subsection{Calibrated \texorpdfstring{$\lambda$}{lambda} per Corruption Family}\label{app:lambda_stats}

Table~\ref{tab:lambda} lists the calibrated weight $\lambda$ for each of the four corruption families of~\citet{hendrycks2019benchmarking} on the three benchmarks. On the CIFAR benchmarks it varies by more than a factor of five across families, from $1.6$ on noise to $9.1$ on weather. On ImageNet-C it lies between $1.00$ and $1.15$, because the shrinkage of Eq.~\eqref{eq:lambda} is strongest there, as Appendix~\ref{app:scaling} shows. The gradient-ratio rule of Section~\ref{sec:calibration} and the shrinkage of Eq.~\eqref{eq:lambda} thus set a different weight for each corruption and benchmark.

\begin{table}[!t]
\caption{The calibrated weight $\lambda$ differs across corruption families and benchmarks. Range and mean of $\lambda$ within each corruption family on CIFAR-10-C, CIFAR-100-C, and ImageNet-C, CLIP ViT-B/16, seed 1.}
\label{tab:lambda}
\centering
\small
\begin{tabular}{l cc cc cc}
\toprule
 & \multicolumn{2}{c}{$K$=10}
 & \multicolumn{2}{c}{$K$=100}
 & \multicolumn{2}{c}{$K$=1000} \\
Family & Range & Mean & Range & Mean & Range & Mean \\
\midrule
Noise   & 1.6--2.9 & 2.1 & 1.9--2.4 & 2.1 & 1.00--1.01 & 1.00 \\
Blur    & 2.6--6.2 & 4.7 & 1.9--5.8 & 4.5 & 1.14--1.15 & 1.14 \\
Weather & 5.1--6.3 & 5.6 & 6.6--9.1 & 8.1 & 1.05--1.09 & 1.07 \\
Digital & 2.3--5.4 & 3.9 & 2.5--5.3 & 4.2 & 1.06--1.15 & 1.11 \\
\bottomrule
\end{tabular}
\end{table}

\section{Additional Visualizations}\label{app:visualizations}

\begin{figure*}[!t]
\centering
\begin{subfigure}[t]{0.30\textwidth}
\centering
\includegraphics[width=\linewidth]{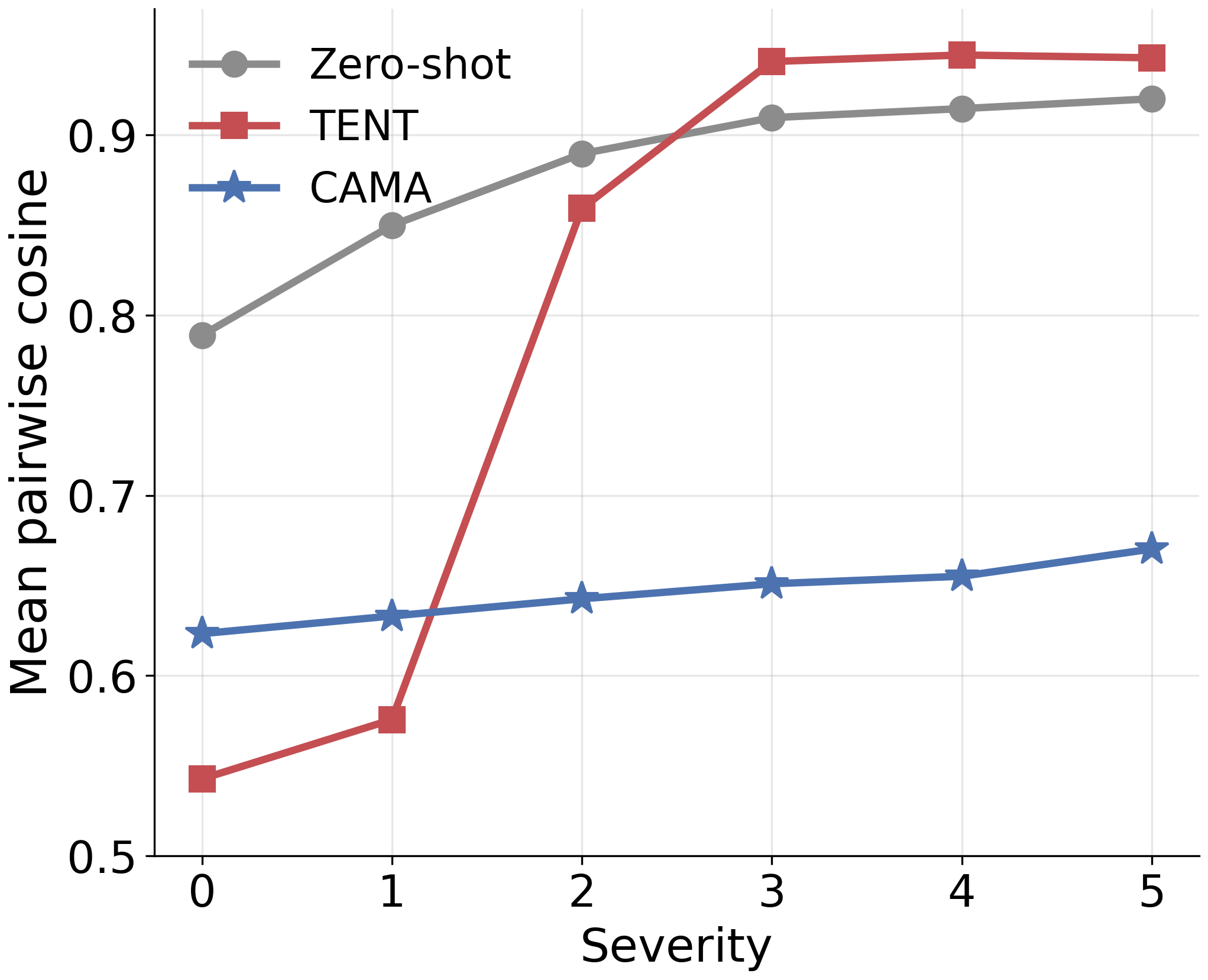}
\caption{Compression}
\end{subfigure}
\hfill
\begin{subfigure}[t]{0.30\textwidth}
\centering
\includegraphics[width=\linewidth]{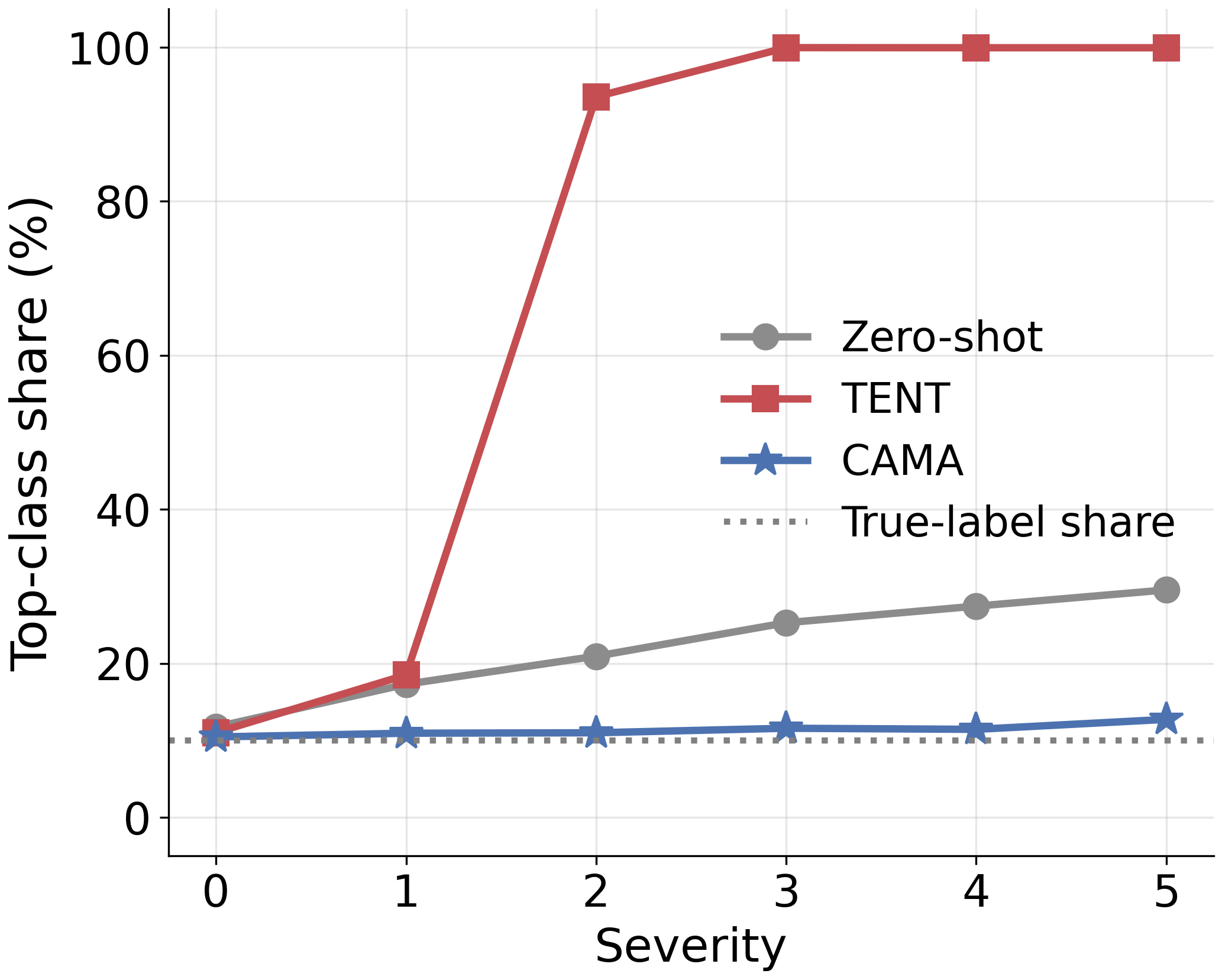}
\caption{Collapse}
\end{subfigure}
\hfill
\begin{subfigure}[t]{0.30\textwidth}
\centering
\includegraphics[width=\linewidth]{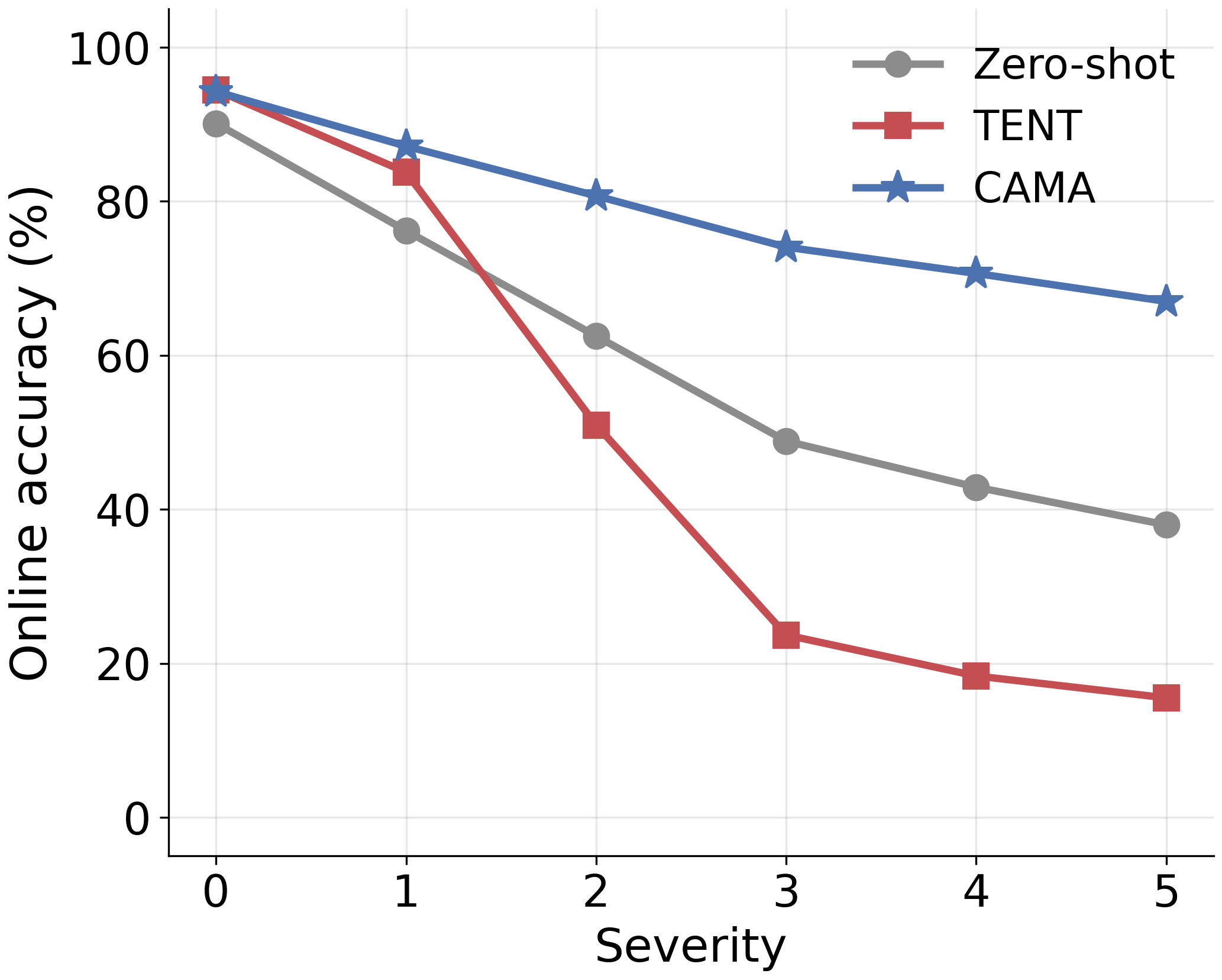}
\caption{Accuracy}
\end{subfigure}
\caption{Compression and collapse grow with severity, TENT amplifies both and falls below zero-shot accuracy once the corruption is strong enough, and CAMA removes the collapse at every severity. CLIP ViT-B/16 on CIFAR-10-C Gaussian noise after adaptation on the full stream, with severity 0 the clean test set. Panels show feature compression (a), prediction collapse (b), and online accuracy (c). The dotted line in (b) is the true-label share.}
\label{fig:severity}
\end{figure*}

\textbf{Severity Sweep.}~~Figure~\ref{fig:severity} repeats the three measurements of Figure~\ref{fig:motivation} after adaptation on the full stream for severities 0 to 5 of CIFAR-10-C Gaussian noise, where severity 0 is the clean test set. Corruption compresses the image features and concentrates the predictions toward one class, and both effects grow with severity. TENT amplifies both, and its accuracy falls below zero-shot from severity 2 on. CAMA removes the collapse at each severity, keeping the features spread and the marginal balanced.

\begin{figure}[p]
\centering
\includegraphics[width=1.0\textwidth]{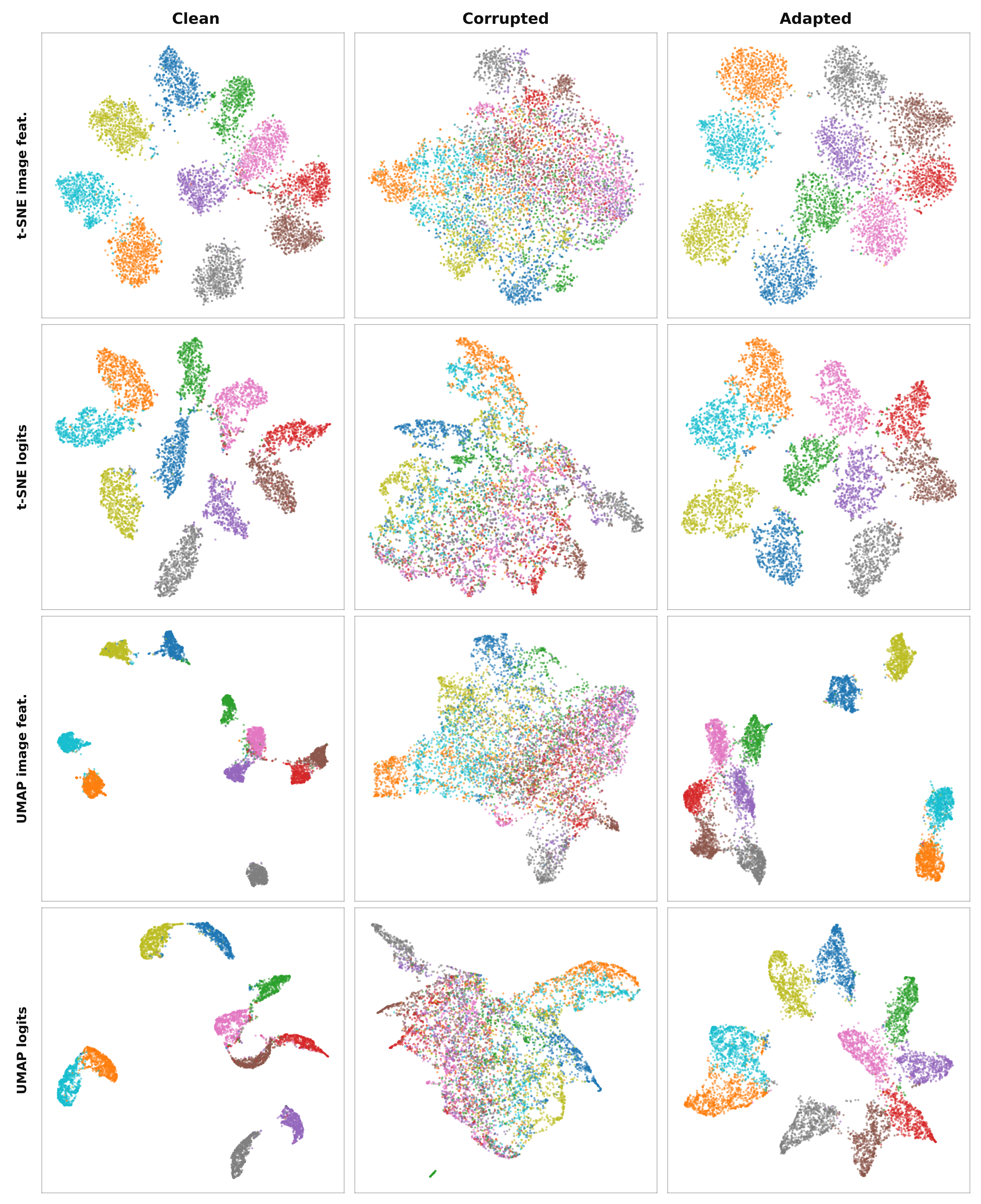}
\caption{Corruption mixes the class clusters in all four views, and CAMA separates them again. CIFAR-10-C ($K{=}10$), Gaussian noise, severity 5. Rows are t-SNE of the image features, t-SNE of the logits, UMAP of the image features, and UMAP of the logits. Columns are clean, corrupted, and after CAMA.}
\label{fig:vis_k10}
\end{figure}

\begin{figure}[p]
\centering
\includegraphics[width=1.0\textwidth]{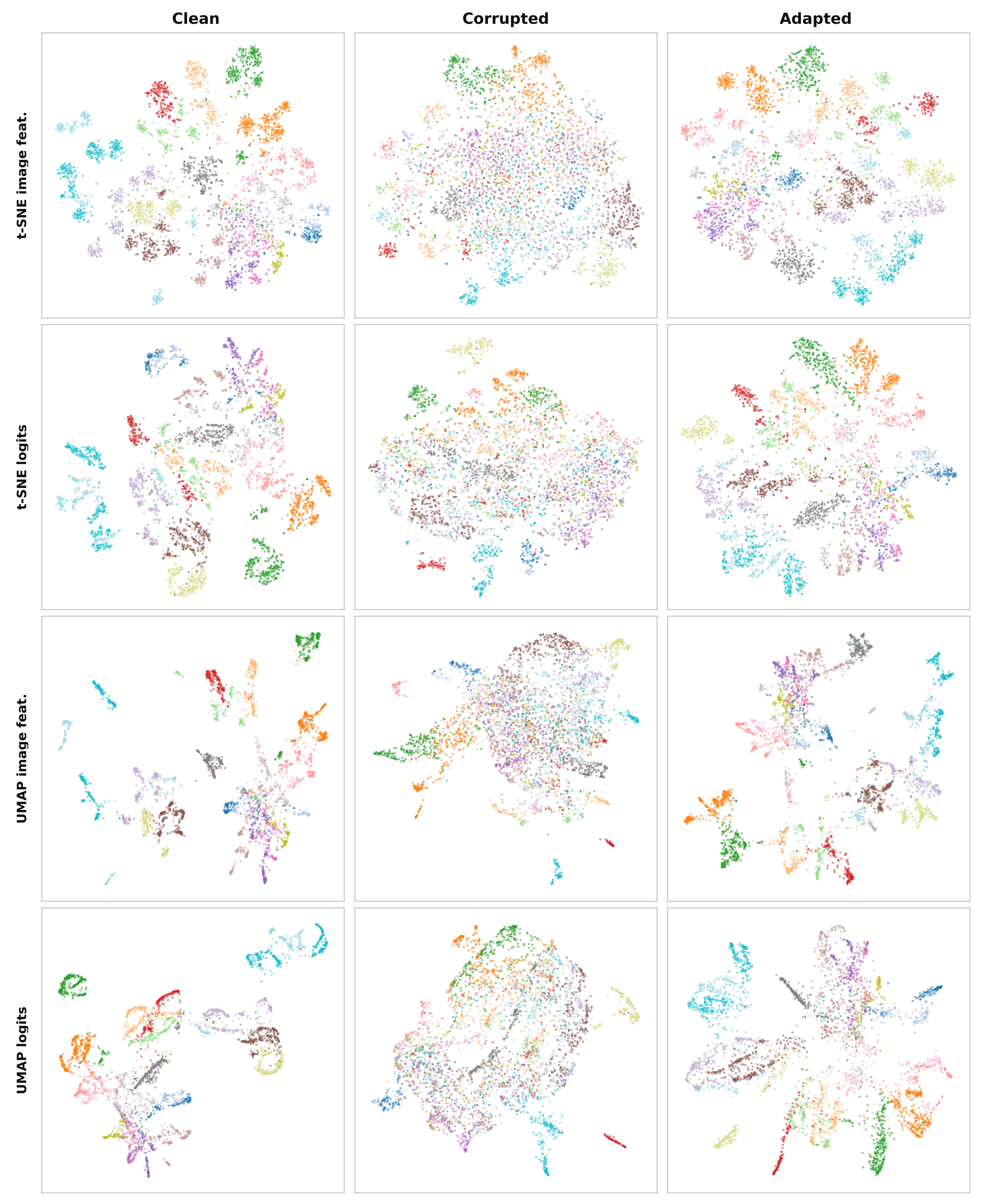}
\caption{On CIFAR-100-C ($K{=}100$) too, corruption mixes the class clusters and CAMA separates them again. Gaussian noise, severity 5, with the layout of Figure~\ref{fig:vis_k10}. Colors are the $20$ CIFAR-100 super-classes.}
\label{fig:vis_k100}
\end{figure}

\begin{figure}[p]
\centering
\includegraphics[width=1.0\textwidth]{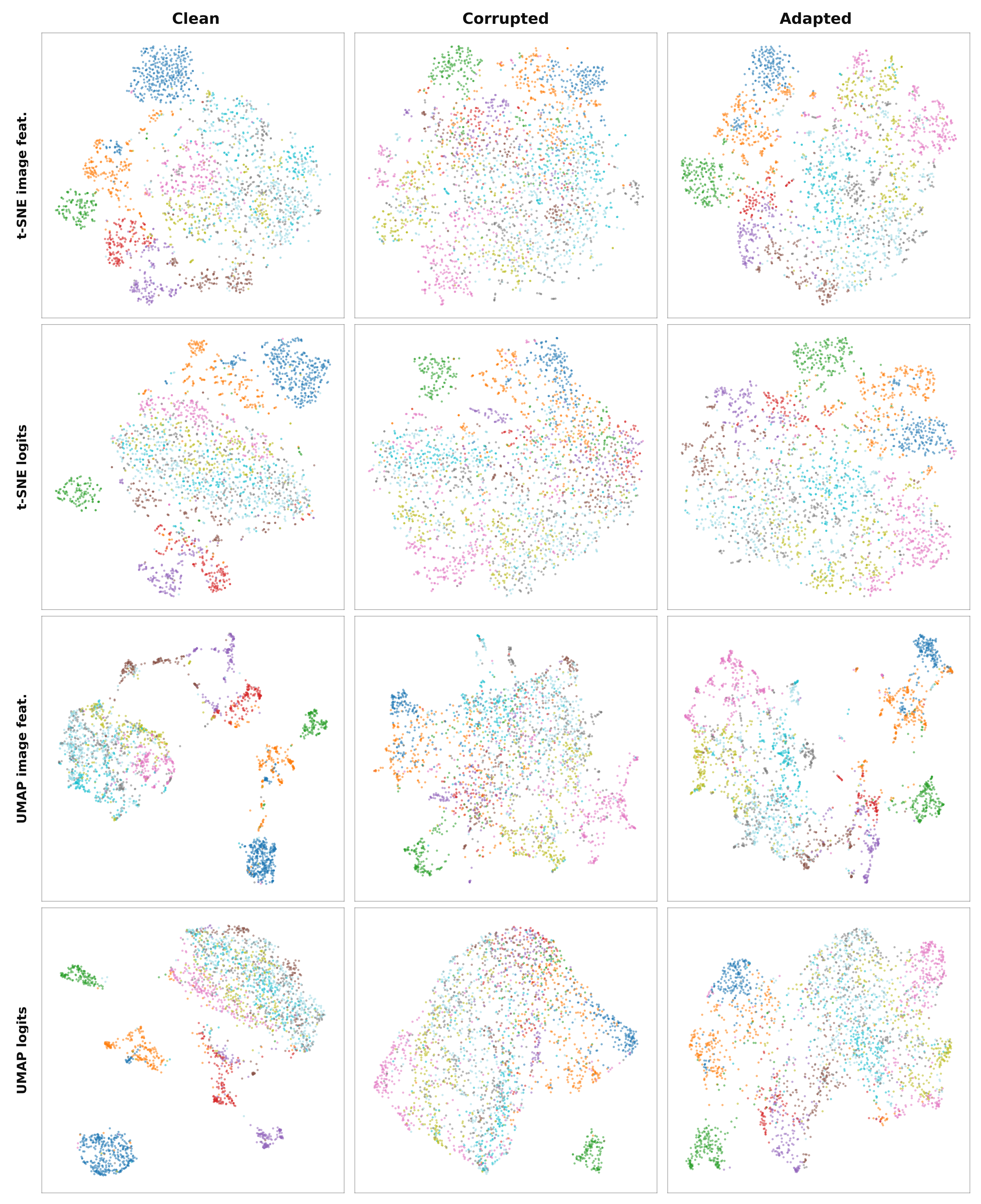}
\caption{On ImageNet-C ($K{=}1000$) too, corruption mixes the class clusters and CAMA separates them again. Gaussian noise, severity 5, with the layout of Figure~\ref{fig:vis_k10}. Colors are $11$ high-level WordNet super-classes.}
\label{fig:vis_k1000}
\end{figure}

\textbf{Feature and Logit Projections.}~~Figures~\ref{fig:vis_k10}, \ref{fig:vis_k100}, and~\ref{fig:vis_k1000} extend Figure~\ref{fig:tsne_main} to all three benchmarks on Gaussian noise at severity 5. On every benchmark and in every view, the classes form separate clusters on clean data, mix under corruption, and separate again after CAMA. For readability, colors mark the official $20$ CIFAR-100 super-classes for $K{=}100$ and $11$ high-level WordNet super-classes for $K{=}1000$.

\section{Implementation Details}\label{app:implementation}

\textbf{Backbones and Adapted Parameters.}~~We load all seven vision--language models through OpenCLIP~\citep{cherti2023reproducible}, namely CLIP ViT-B/16, RN50, and RN101 with the OpenAI weights, SigLIP ViT-B/16 (\texttt{webli}), MetaCLIP ViT-B/16 (\texttt{metaclip\_fullcc}), EVA02-B/16 (\texttt{merged2b\_s8b\_b131k}), and ConvNeXt-Base-w (\texttt{laion2b\_s13b\_b82k}). CAMA adapts the LayerNorm affine parameters of both the image and the text encoder, so that the text features move with the adaptation, and freezes all other parameters, as BATCLIP~\citep{maharana2025batclip} does. The image encoders of the ResNet CLIPs use BatchNorm instead of LayerNorm. On them CAMA adapts the BatchNorm affine parameters of the image encoder and the LayerNorm affine parameters of the text encoder, and it keeps the source BatchNorm statistics. Every baseline that adapts BatchNorm layers keeps the source statistics as well, except RoTTA, whose robust normalization updates them as published. VTE updates no parameter. CAMA and every baseline run in our framework use the prompt ``a photo of a \{\}.'' on all benchmarks, except TPT and VTE, which tune and ensemble prompts as published. Stochastic depth is inactive during adaptation for every method on ConvNeXt-CLIP. ImageNet-C is distributed at 224 pixels and is fed to every model without resampling. The convolutional tower of ConvNeXt-CLIP, pre-trained at 256 pixels, accepts this input directly. TPT is evaluated on CLIP ViT-B/16 only.

\textbf{Hyperparameters and Protocol.}~~Table~\ref{tab:hp} summarizes the per-benchmark hyperparameters. CAMA uses the optimizer, learning rate, weight decay, and batch size of BATCLIP's published setup for each benchmark. Every baseline runs with the configuration that this setup assigns to it, which follows the baseline's official implementation. On ImageNet-C this setup evaluates the 5,000-image RobustBench subset of each corruption, and every ImageNet-C row of Tables~\ref{tab:main}, \ref{tab:main_full}, and~\ref{tab:backbones} uses this subset. Each batch gets one gradient update, the online budget of every compared method since TENT. The method-specific constants $(\zeta, \kappa) = (0.05, 0.1)$ are round defaults, fixed once and used unchanged across all benchmarks, backbones, corruption types, and severities. The small $\kappa$ keeps $\bpi$ from inheriting single-class peaks from corrupted predictions. The number of calibration batches is $N_{\mathrm{cal}} = 3$ on the CIFAR benchmarks and $16$ on ImageNet-C, that is, 600 and 1{,}024 samples. No hyperparameter was selected on test accuracy, and Appendix~\ref{app:sensitivity} reports sweeps of the batch size and of $(\zeta, \kappa)$ as sensitivity checks.

\textbf{Baseline Configurations.}~~As in BATCLIP~\citep{maharana2025batclip}, WATT-P$^*$ and WATT-S$^*$ run on the CIFAR benchmarks and StatA on ImageNet-C, the benchmarks closest to those of their own papers. Mint and ClipTTA run from their official code with one update per batch and their published hyperparameters otherwise unchanged. Mint's released batch size of 20, with the model restored after every batch, gives 73.9, 44.0, and 32.4 on the three benchmarks, within 0.6\% of its result at the protocol batch size. Mint uses learning rate 0.007 on all three benchmarks, the value its paper assigns to ViT-B backbones. ClipTTA's released configuration makes ten updates per batch and keeps a class-wise memory of confident past samples. Table~\ref{tab:main} runs it with one update per batch at the protocol batch size and without the memory. Keeping the memory changes its means by $-0.46\%$ and $+0.06\%$ on CIFAR-10-C and CIFAR-100-C. On the six other backbones and on the streams of Section~\ref{sec:streams}, Mint and ClipTTA run in our framework with the update rules of their official code and the same hyperparameters.

\begin{table}[!t]
\caption{CAMA reuses BATCLIP's published setup on each benchmark and keeps its own constants $\zeta$ and $\kappa$ fixed across benchmarks. Hyperparameters per benchmark.}
\label{tab:hp}
\centering
\small
\begin{tabular}{lccc}
\toprule
 & CIFAR-10-C & CIFAR-100-C & ImageNet-C \\
\midrule
Optimizer & AdamW & Adam & AdamW \\
Learning rate & $10^{-3}$ & $5 \times 10^{-4}$ & $5 \times 10^{-4}$ \\
Weight decay & 0.01 & 0.0 & 0.01 \\
Batch size $B$ & 200 & 200 & 64 \\
$N_{\mathrm{cal}}$ & 3 & 3 & 16 \\
$\zeta$ (Eq.~\eqref{eq:pi_def}) & 0.05 & 0.05 & 0.05 \\
$\kappa$ (Eq.~\eqref{eq:pi_def}) & 0.1 & 0.1 & 0.1 \\
\bottomrule
\end{tabular}
\end{table}

\textbf{Run Setup.}~~All CAMA runs and all runs on the six backbones other than CLIP ViT-B/16 used NVIDIA H100, H200, and A30 GPUs. On CLIP ViT-B/16, the Mint and ClipTTA rows of Table~\ref{tab:main} also used H100 and H200 GPUs, and the other baseline rows used NVIDIA RTX A5000 GPUs. CLIP ViT-B/16 runs in mixed precision, with \texttt{torch.cuda.amp.autocast} and a gradient scaler initialized at $1000$, and adaptation on the other six backbones runs in float32. Adaptation in the standard benchmark is episodic, with model parameters reset to the pre-trained weights and the optimizer re-initialized at the start of each corruption type. The sample order of each stream is drawn once from seed $1$ when the dataset is built, and the data loader reads it in that order without shuffling, so the seed fixes both the sample order and the composition of the calibration batches. Tables~\ref{tab:main} and~\ref{tab:backbones} report seed 1. Over three seeds on CLIP ViT-B/16, the standard deviation is at most 0.1\% for CAMA and Mint and 1.0\% for TENT and BATCLIP. Online accuracy is scored over the entire stream for all methods, including the calibration batches. Calibrating on one corruption family and adapting on another changes the accuracy of CAMA by at most 0.7\% on CIFAR-10-C.

\textbf{Calibrated Reference.}~~Each choice in Eq.~\eqref{eq:pi_def} helps the reference follow the class marginal of the stream without copying the corruption bias of Proposition~\ref{prop:sink}. Ranks make the estimate independent of the logit scale and of $\tau$. Harmonic weights keep the runner-up classes that a hard vote would discard. The constant $\zeta$ lifts the small scores of classes that rarely rank high, which limits how strongly $\bpi$ can favor one class over another. Tempering keeps the class ordering but moves every ratio between two classes toward one, which shrinks the size of the corruption bias. At the default constants these choices keep every ratio $\pi_k/\pi_j$ close to one for any calibration outcome, as the Remark on Large $K$ in Appendix~\ref{app:sample_complexity} shows.

\textbf{Calibration Phase.}~~Calibration runs forward and backward passes without optimizer updates on the first $N_{\mathrm{cal}}$ batches of the stream. The reference $\bpi$ is estimated from the logits pooled across all of these batches. $\lambda_0$ follows Appendix~\ref{app:lambda0_derivation} at batch size $B$. The shrinkage of Eq.~\eqref{eq:lambda} uses the worst-case value $d_{\mathrm{eff}} = K - 1$ of Appendix~\ref{app:scaling} and therefore depends only on $B$ and $K$. The output of Eq.~\eqref{eq:lambda} is floored at $\lambda = 1$, where the alignment term of Eq.~\eqref{eq:lossB} vanishes, $\pdag$ is not formed, and Proposition~\ref{prop:threshold_general} still makes collapse suboptimal for any batch with $\Ibatch > 0$. The floor applies only when the calibration returns $\lambda_0 \leq 1$, in nine of the 315 combinations of backbone, benchmark, and corruption in Tables~\ref{tab:main} and~\ref{tab:backbones} on seed 1, all with $\lambda_0$ within $0.07$ of $1$. The tempering exponent $\kappa$ also limits how sharp the escort becomes. Forming $\pdag$ raises $s_k + \zeta$ to the power $\kappa\lambda/(\lambda-1)$, which stays small at the weights calibrated on the CIFAR benchmarks and is $0.5$ even at $\lambda = 1.25$ with $\kappa = 0.1$. Hence $\pdag$ keeps the observed class ordering without becoming a single-class target. As $\lambda \to 1$ the escort exponent $\alpha$ grows while the weight $\lambda - 1$ vanishes. Their product $(\lambda-1)\alpha = \lambda$ gives the term $-\lambda\langle\pbar, \log\bpi\rangle$ of Eq.~\eqref{eq:solver}, which the tempering of Eq.~\eqref{eq:pi_def} keeps within $0.16\lambda$ nats, as Appendix~\ref{app:sample_complexity} shows.

\textbf{Online Accuracy Convention.}~~Online accuracy is computed over the full test stream for every method, scoring each batch with the prediction made before the update on that batch. Calibration makes no update, and the stream is then adapted from its first batch as for the baselines, so the only extra cost of calibration is the wall-clock time in Appendix~\ref{app:sensitivity}.

\textbf{Seeds and Environments.}~~Table~\ref{tab:seeds} reports means and standard deviations over three seeds on CLIP ViT-B/16 for CAMA, TENT, BATCLIP, and Mint, the strongest baseline on CIFAR-100-C and ImageNet-C. CAMA's standard deviations are far below its margins on all three benchmarks. Three seeds also confirm the two closest results of Table~\ref{tab:backbones}, EVA-CLIP and ConvNeXt-CLIP on CIFAR-10-C. CAMA scores $90.72 \pm 0.11$ against $89.77 \pm 0.16$ for BATCLIP on EVA-CLIP and $78.82 \pm 0.18$ against $77.88 \pm 0.10$ for BATCLIP on ConvNeXt-CLIP, and it leads on every seed. All other entries in the paper are single runs with seed 1, except the Mint and ClipTTA rows. These use the default seeds of their official code, $0$ for Mint and $42$ for ClipTTA, and Mint's three runs in Table~\ref{tab:seeds} add seeds $2$ and $3$ to its default. Each of the six backbones other than CLIP ViT-B/16 was run in a single environment for all methods.

\begin{table}[htbp]
\centering
\caption{CAMA's standard deviations are far below its margins on all three benchmarks. Mean $\pm$ standard deviation of online accuracy (\%) over three seeds on CLIP ViT-B/16, 15 corruptions at severity 5.}
\label{tab:seeds}
\begin{tabular}{lrrr}
\toprule
Method & CIFAR-10-C & CIFAR-100-C & ImageNet-C \\
\midrule
TENT & $63.06\pm0.95$ & $37.80\pm0.14$ & $25.13\pm0.02$ \\
BATCLIP & $73.90\pm0.07$ & $42.06\pm0.12$ & $30.58\pm0.14$ \\
Mint & $73.73\pm0.07$ & $44.61\pm0.08$ & $32.34\pm0.06$ \\
CAMA & $79.48\pm0.05$ & $48.75\pm0.10$ & $34.97\pm0.04$ \\
\bottomrule
\end{tabular}
\end{table}

\section{Sensitivity Analysis}\label{app:sensitivity}

All results in this appendix use CLIP ViT-B/16 and report mean online accuracy over the 15 corruptions at severity 5. Entries are single runs with seed 1 unless a standard deviation is given. The first three paragraphs vary the settings that CAMA inherits or fixes, namely the batch size, the smoothing constants, and the number of calibration batches. The next three paragraphs test the calibration itself, and the last reports the cost.

\textbf{Batch Size.}~~The protocol batch sizes, $B = 200$ on the CIFAR benchmarks and $B = 64$ on ImageNet-C, are BATCLIP's published values. Every method in Tables~\ref{tab:main} and~\ref{tab:backbones} uses them. Table~\ref{tab:batch_sweep} varies $B$ with all other settings unchanged, including $N_{\mathrm{cal}}$ in batches, so that the calibration batches shrink with $B$. Since one update is made per batch, a smaller $B$ also means more updates on the same stream. Smaller batches also make the batch marginal noisier, which hurts any objective that matches the batch marginal. On the CIFAR benchmarks accuracy stays within $2.8\%$ of the protocol value across the sweep, and every setting remains above the strongest baseline of Table~\ref{tab:main} at the protocol batch size, $73.85$ and $44.53$. On the 5,000-image ImageNet-C stream the gain from more updates outweighs the noisier marginal, with $313$ updates at $B = 16$ against $25$ at $B = 200$. Accuracy therefore rises as the batch shrinks, from $31.75$ at $B = 200$ to $36.61$ at $B = 16$, and stays above the strongest baseline, $32.40$, at every batch size up to $128$. As the batch shrinks, the shrinkage lowers the calibrated weight, on CIFAR-10-C from $4.20$ at $B = 200$ to $2.2$ at $B = 16$, on CIFAR-100-C from $4.92$ to $1.2$, and on ImageNet-C from $1.43$ to $1.01$.

\begin{table}[htbp]
\centering
\caption{The protocol batch size, fixed before any CAMA run, is not CAMA's best on any benchmark. Mean online accuracy (\%) for each batch size $B$ over 15 corruptions at severity 5, seed 1. $^{\mathrm{s}}$ marks the protocol batch size of Table~\ref{tab:main}.}
\label{tab:batch_sweep}
\begin{tabular}{lrrrrr}
\toprule
$B$ & 16 & 32 & 64 & 128 & 200 \\
\midrule
CIFAR-10-C & 76.69 & 79.01 & 80.08 & \textbf{80.30} & 79.46$^{\mathrm{s}}$ \\
CIFAR-100-C & 51.37 & \textbf{51.56} & 51.17 & 50.21 & 48.77$^{\mathrm{s}}$ \\
ImageNet-C & \textbf{36.61} & 36.14 & 34.93$^{\mathrm{s}}$ & 33.15 & 31.75 \\
\bottomrule
\end{tabular}
\end{table}

\textbf{Smoothing Constants.}~~Table~\ref{tab:zeta_kappa} varies $(\zeta, \kappa)$ on a $3 \times 3$ grid around the default $(0.05, 0.1)$ on all 15 corruptions. The spread is $0.19\%$ on CIFAR-10-C and $0.04\%$ on CIFAR-100-C. The constants only smooth the rank scores, and the reference depends on the class ordering rather than on exact values, so moderate changes leave the reference, and with it the target $\pdag$, nearly unchanged.

\begin{table}[htbp]
\centering
\caption{Accuracy barely changes across the smoothing constants $\zeta$ and $\kappa$. Mean online accuracy (\%) over 15 corruptions at severity 5, seed 1, with $\zeta$ in rows and $\kappa$ in columns. The center entry is the default used everywhere else.}
\label{tab:zeta_kappa}
\begin{tabular}{l rrr @{\qquad} rrr}
\toprule
& \multicolumn{3}{c}{CIFAR-10-C} & \multicolumn{3}{c}{CIFAR-100-C} \\
\cmidrule(lr){2-4}\cmidrule(lr){5-7}
$\zeta \backslash \kappa$ & 0.05 & 0.1 & 0.2 & 0.05 & 0.1 & 0.2 \\
\midrule
0.01 & 79.50 & 79.48 & 79.35 & 48.76 & 48.75 & 48.74 \\
0.05 & 79.52 & 79.46 & 79.46 & 48.76 & 48.77 & 48.76 \\
0.10 & 79.53 & 79.51 & 79.48 & 48.77 & 48.77 & 48.78 \\
\bottomrule
\end{tabular}
\end{table}

\textbf{Number of Calibration Batches.}~~The default is $N_{\mathrm{cal}} = 3$ batches on the CIFAR benchmarks and $16$ on ImageNet-C. With $1$, $2$, $3$, or $6$ calibration batches on CIFAR and $4$, $8$, $16$, or $32$ on ImageNet-C, and everything else unchanged, the mean stays within $0.1\%$ of the default on every benchmark. It ranges from $79.43$ to $79.50$ on CIFAR-10-C, from $48.69$ to $48.80$ on CIFAR-100-C, and from $34.93$ to $34.95$ on ImageNet-C. The number of calibration batches therefore needs no tuning.

\begin{figure}[htbp]
\centering
\includegraphics[width=0.5\textwidth]{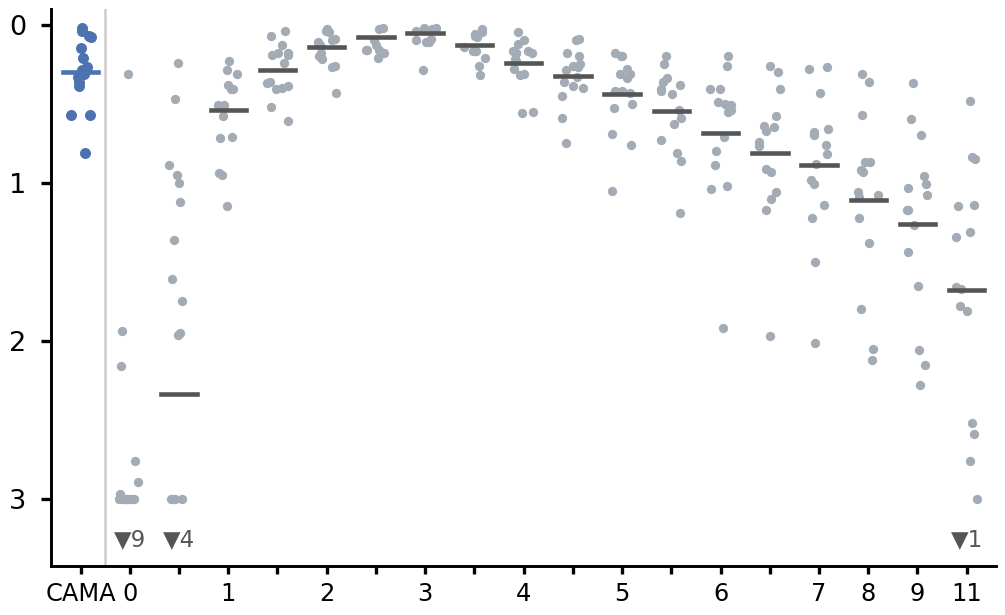}
\caption{Without labels, the calibrated weight comes within 1\% of the accuracy of an oracle that picks the best fixed weight for each corruption with test labels, whereas fixed weights below one fall far behind on many corruptions and, beyond the best ones, accuracy falls steadily as the weight grows. CLIP ViT-B/16 on the 15 CIFAR-10-C corruptions, seed 1. The horizontal axis lists the calibrated weight of CAMA, then fixed weights $\lambda$ from $0$ to $11$. The vertical axis is the accuracy gap (\%) to the best fixed weight of each corruption. Each marker is one corruption and each bar is the mean over the 15, without the zero gap of the best weight itself. Gaps beyond 3\% are drawn at 3\% and counted below the axis.}
\label{fig:lambda_gap}
\end{figure}

\textbf{Fixed Alignment Weight.}~~Figure~\ref{fig:lambda_gap} compares the calibrated weight with 18 fixed weights from $0$ to $11$ on CIFAR-10-C, corruption by corruption. Without labels, the calibrated weight comes within 1\% of the accuracy of an oracle that picks the best fixed weight for each corruption with test labels. Fixed weights below one lose up to $54\%$, and beyond the best fixed weights the loss grows steadily with the weight. The calibrated weights average $4.20$ on CIFAR-10-C and $4.92$ on CIFAR-100-C, and Appendix~\ref{app:lambda_stats} lists their variation per corruption family.

\textbf{Calibration Mismatch.}~~In the continual stream of Section~\ref{sec:streams}, one calibration serves all 15 corruptions, so CAMA must tolerate a calibration taken on a different corruption. To test this, on CIFAR-10-C we calibrate on one of the four corruption families and stream another, for all twelve ordered pairs. Online accuracy changes by at most $0.7\%$, with a mean absolute change of $0.3\%$. This holds even when the mismatched weight is $3.2$ times the matched one. Calibrating on fog and streaming Gaussian noise gives $\lambda = 5.07$, where the matched value is $1.58$. Two results explain this tolerance. An error in the reference enters the excess loss only at second order, by Proposition~\ref{prop:robustness}. An error in the weight costs little, since accuracy varies little across the fixed weights near the calibrated one in Figure~\ref{fig:lambda_gap}. Where a batch cannot resolve the class marginal, the weight that the shrinkage sets does matter, as Table~\ref{tab:shrinkage} shows.

\textbf{Finite-Batch Shrinkage.}~~Table~\ref{tab:shrinkage_small_batch} gives the accuracies behind the small-batch columns of Table~\ref{tab:shrinkage}. It compares runs with and without the shrinkage of Eq.~\eqref{eq:lambda}, with the same seeds and sample order in both. The benefit of the shrinkage grows as the batch shrinks, from $0.5\%$ at $B = 16$ to $7.0\%$ at $B = 4$ on CIFAR-10-C, from $11.5$ to $21.0$ on CIFAR-100-C, and from $7.4$ to $11.2$ on ImageNet-C. Without the shrinkage the runs also become unstable across seeds, with standard deviations of $4.09$ on CIFAR-10-C and $2.79$ on CIFAR-100-C at $B = 4$ against $0.20$ and $0.37$ with it.

\begin{table}[htbp]
\centering
\caption{The shrinkage helps more as the batch shrinks, and without it the runs vary more across seeds. Online accuracy (\%) with and without the shrinkage of Eq.~\eqref{eq:lambda} on CLIP ViT-B/16, mean $\pm$ std over three seeds on the CIFAR benchmarks and seed 1 on the 5,000-image ImageNet-C protocol.}
\label{tab:shrinkage_small_batch}
\begin{tabular}{llrrr}
\toprule
Benchmark & $B$ & With shrinkage & Without shrinkage & Gain \\
\midrule
\multirow{3}{*}{CIFAR-10-C} & 16 & $77.12\pm0.37$ & $76.61\pm0.61$ & $+0.5$ \\
 & 8 & $73.63\pm0.43$ & $71.33\pm0.72$ & $+2.3$ \\
 & 4 & $68.65\pm0.20$ & $61.70\pm4.09$ & $+7.0$ \\
\midrule
\multirow{3}{*}{CIFAR-100-C} & 16 & $51.27\pm0.05$ & $39.78\pm1.36$ & $+11.5$ \\
 & 8 & $49.92\pm0.22$ & $30.63\pm1.08$ & $+19.3$ \\
 & 4 & $46.49\pm0.37$ & $25.50\pm2.79$ & $+21.0$ \\
\midrule
\multirow{3}{*}{ImageNet-C} & 16 & $36.61$ & $29.23$ & $+7.4$ \\
 & 8 & $35.96$ & $25.91$ & $+10.0$ \\
 & 4 & $34.51$ & $23.33$ & $+11.2$ \\
\bottomrule
\end{tabular}
\end{table}

\textbf{Cost.}~~We ran all methods back-to-back on one fixed A30 GPU on identical CIFAR-10-C streams, with CAMA and BATCLIP in float32 and TENT in mixed precision. The full 15-corruption benchmark takes $409.3$~s for CAMA, $345.0$~s for TENT, and $367.0$~s for BATCLIP, so CAMA's total test-time cost is $1.19\times$ TENT and $1.12\times$ BATCLIP. Calibration accounts for $5.4$~s per corruption, about $20\%$ of CAMA's per-corruption time on CIFAR. On ImageNet-C we timed one A30 with the stream preloaded, so that only computation is counted. Adaptation of the 5,000-image stream takes $51$~s per corruption. Cutting the calibration from 16 batches, $37$~s, to 4 batches, $9$~s, leaves accuracy unchanged, $34.95$ with 4 batches against $34.93$ with 16, as the sweep over calibration batches above shows.

\section{Additional Results}\label{app:additional}

This appendix gives the full results behind Sections~\ref{sec:results} to~\ref{sec:streams}, in that order. Table~\ref{tab:main_full} lists every corruption behind the family means of Table~\ref{tab:main}.

\begin{table*}[!t]
\caption{CAMA is first on all but one corruption of each CIFAR benchmark and on every ImageNet-C corruption. Per-corruption online accuracy (\%) on CLIP ViT-B/16 at severity 5, the full version of Table~\ref{tab:main}. Dashed lines separate general from vision--language TTA methods, and $^*$ marks the online variants of WATT. Best per column in \textbf{bold}.}
\label{tab:main_full}
\centering
\setlength{\tabcolsep}{3pt}
\renewcommand{\arraystretch}{0.92}
\resizebox{\textwidth}{!}{%
\begin{tabular}{cl ccccc ccccc ccccc | c}
\toprule
& Method
& \rotatebox{70}{Gaussian} & \rotatebox{70}{Shot} & \rotatebox{70}{Impulse}
& \rotatebox{70}{Defocus} & \rotatebox{70}{Glass}
& \rotatebox{70}{Motion} & \rotatebox{70}{Zoom}
& \rotatebox{70}{Snow} & \rotatebox{70}{Frost} & \rotatebox{70}{Fog} & \rotatebox{70}{Brightness}
& \rotatebox{70}{Contrast} & \rotatebox{70}{Elastic} & \rotatebox{70}{Pixelate} & \rotatebox{70}{JPEG}
& Mean \\
\midrule
\parbox[t]{3.5mm}{\multirow{13}{*}{\rotatebox[origin=c]{90}{CIFAR-10-C}}}
& ZS & 37.92 & 41.70 & 54.42 & 71.75 & 40.89 & 67.93 & 73.62 & 73.89 & 77.35 & 70.22 & 84.45 & 62.36 & 53.81 & 47.65 & 59.43 & 61.16 \\
& TENT & 15.49 & 18.28 & 38.12 & 81.59 & 21.73 & 76.32 & 82.35 & 84.62 & 82.19 & 80.60 & \textbf{91.83} & 80.55 & 63.52 & 58.57 & 54.71 & 62.03 \\
& RoTTA & 39.17 & 42.90 & 55.41 & 72.18 & 41.32 & 68.02 & 74.01 & 74.38 & 78.01 & 70.80 & 84.80 & 63.19 & 54.62 & 49.32 & 60.15 & 61.89 \\
& RPL & 15.47 & 17.43 & 40.73 & 81.76 & 20.08 & 69.89 & 82.93 & 84.43 & 83.19 & 81.84 & 91.80 & 79.42 & 64.89 & 54.07 & 54.90 & 61.52 \\
& SAR & 47.98 & 53.60 & 60.56 & 74.30 & 47.56 & 73.15 & 76.43 & 77.91 & 79.88 & 75.66 & 86.79 & 71.62 & 58.34 & 62.03 & 64.71 & 67.37 \\
\noalign{\vskip 1.5pt}\cdashline{2-18}\noalign{\vskip 2pt}
& TPT & 37.74 & 42.24 & 60.57 & 72.88 & 44.80 & 69.69 & 75.37 & 75.96 & 78.84 & 72.12 & 85.68 & 62.04 & 58.90 & 55.14 & 62.64 & 63.64 \\
& VTE & 42.42 & 46.26 & 64.23 & 71.10 & 45.58 & 68.50 & 73.66 & 76.75 & 78.27 & 71.02 & 85.28 & 57.24 & 59.54 & 60.59 & 61.85 & 64.15 \\
& WATT-P* & 44.70 & 49.66 & 60.48 & 74.35 & 46.33 & 72.21 & 76.28 & 78.18 & 80.97 & 74.90 & 87.66 & 68.63 & 58.35 & 56.13 & 63.47 & 66.15 \\
& WATT-S* & 57.22 & 61.86 & 67.63 & 78.61 & 53.07 & 77.85 & 80.14 & 81.84 & 83.46 & 80.38 & 89.67 & 78.45 & 65.76 & 68.63 & 67.67 & 72.81 \\
& BATCLIP & 61.13 & 64.09 & 65.76 & 80.51 & 54.96 & 80.65 & 81.94 & 83.04 & 84.19 & 80.84 & 88.95 & 82.15 & 69.16 & 62.68 & 67.64 & 73.85 \\
& Mint & 59.59 & 65.51 & 69.56 & 79.40 & 58.05 & 80.34 & 81.25 & 82.78 & 82.61 & 80.84 & 90.53 & 79.84 & 66.95 & 61.48 & 66.27 & 73.67 \\
& ClipTTA & 49.97 & 53.95 & 61.08 & 75.73 & 49.11 & 73.64 & 77.95 & 78.82 & 81.09 & 75.83 & 87.94 & 72.04 & 61.44 & 61.08 & 63.73 & 68.23 \\
& CAMA (Ours) & \textbf{66.88} & \textbf{70.60} & \textbf{76.34} & \textbf{83.18} & \textbf{66.80} & \textbf{82.90} & \textbf{84.98} & \textbf{85.70} & \textbf{85.69} & \textbf{85.07} & 91.69 & \textbf{86.41} & \textbf{75.08} & \textbf{77.89} & \textbf{72.72} & \textbf{79.46} \\
\midrule
\parbox[t]{3.5mm}{\multirow{13}{*}{\rotatebox[origin=c]{90}{CIFAR-100-C}}}
& ZS & 19.64 & 21.40 & 25.26 & 42.54 & 20.03 & 43.17 & 47.95 & 48.35 & 49.74 & 41.57 & 57.02 & 34.58 & 29.15 & 23.96 & 32.43 & 35.79 \\
& TENT & 7.60 & 8.21 & 8.33 & 51.81 & 7.95 & 52.45 & 55.34 & 54.16 & 36.17 & 50.92 & 65.63 & \textbf{54.51} & 36.52 & 43.99 & 35.81 & 37.96 \\
& RoTTA & 20.65 & 22.22 & 26.17 & 42.48 & 20.26 & 42.90 & 47.88 & 48.75 & 49.92 & 41.86 & 57.00 & 34.52 & 29.27 & 25.08 & 32.88 & 36.12 \\
& RPL & 6.44 & 7.09 & 7.09 & 52.16 & 11.81 & 52.33 & 55.50 & 54.20 & 38.83 & 51.99 & 66.07 & 54.45 & 36.86 & 42.83 & 39.45 & 38.47 \\
& SAR & 25.30 & 27.19 & 32.78 & 47.12 & 23.42 & 47.16 & 51.70 & 51.94 & 52.48 & 48.77 & 61.54 & 44.50 & 32.26 & 33.67 & 38.06 & 41.19 \\
\noalign{\vskip 1.5pt}\cdashline{2-18}\noalign{\vskip 2pt}
& TPT & 17.95 & 19.51 & 27.13 & 43.53 & 20.08 & 42.65 & 48.63 & 49.11 & 49.48 & 42.14 & 57.35 & 33.26 & 31.13 & 27.59 & 32.75 & 36.15 \\
& VTE & 17.96 & 18.72 & 28.17 & 40.38 & 19.60 & 39.50 & 45.33 & 48.24 & 46.87 & 40.73 & 55.31 & 30.04 & 32.47 & 30.35 & 31.45 & 35.01 \\
& WATT-P* & 20.53 & 22.22 & 27.30 & 43.14 & 17.51 & 42.37 & 48.17 & 47.31 & 49.34 & 41.49 & 57.07 & 35.29 & 27.75 & 25.83 & 31.89 & 35.81 \\
& WATT-S* & 21.20 & 23.11 & 28.23 & 44.16 & 18.45 & 43.44 & 49.14 & 48.16 & 50.05 & 42.35 & 57.82 & 36.43 & 28.36 & 26.85 & 32.93 & 36.71 \\
& BATCLIP & 24.91 & 27.73 & 33.66 & 50.11 & 26.27 & 48.49 & 54.85 & 52.35 & 51.62 & 48.38 & 63.27 & 45.21 & 34.74 & 32.38 & 37.31 & 42.09 \\
& Mint & 30.33 & 33.90 & 40.38 & 51.20 & 26.88 & 50.09 & 55.37 & 53.16 & 52.02 & 51.07 & 66.09 & 49.52 & 36.78 & 34.52 & 36.61 & 44.53 \\
& ClipTTA & 25.53 & 27.46 & 32.12 & 46.54 & 24.82 & 46.50 & 51.64 & 51.54 & 52.13 & 46.46 & 61.62 & 40.78 & 33.30 & 31.13 & 36.35 & 40.53 \\
& CAMA (Ours) & \textbf{35.91} & \textbf{37.90} & \textbf{43.81} & \textbf{53.26} & \textbf{34.29} & \textbf{52.54} & \textbf{57.89} & \textbf{56.37} & \textbf{56.84} & \textbf{52.99} & \textbf{66.92} & 53.83 & \textbf{41.90} & \textbf{44.75} & \textbf{42.38} & \textbf{48.77} \\
\midrule
\parbox[t]{3.5mm}{\multirow{13}{*}{\rotatebox[origin=c]{90}{ImageNet-C}}}
& ZS & 11.18 & 12.54 & 12.04 & 23.36 & 15.18 & 24.50 & 22.58 & 32.32 & 29.88 & 35.88 & 54.18 & 17.20 & 12.72 & 30.96 & 33.26 & 24.51 \\
& TENT & 5.14 & 5.70 & 7.44 & 25.22 & 19.34 & 26.80 & 24.16 & 33.56 & 30.42 & 37.74 & 54.24 & 22.50 & 13.90 & 35.02 & 36.08 & 25.15 \\
& RoTTA & 11.34 & 12.96 & 12.32 & 23.38 & 15.50 & 24.66 & 22.90 & 32.56 & 30.02 & 35.98 & 54.32 & 17.20 & 12.80 & 31.06 & 33.46 & 24.78 \\
& RPL & 9.04 & 10.04 & 10.96 & 24.40 & 17.40 & 26.28 & 23.76 & 32.70 & 30.62 & 36.64 & 54.04 & 19.38 & 13.24 & 33.14 & 34.60 & 25.08 \\
& SAR & 17.96 & 20.46 & 20.68 & 25.72 & 23.04 & 29.52 & 26.04 & 34.92 & 32.74 & 39.00 & 55.00 & 27.14 & 19.64 & 36.66 & 37.50 & 29.73 \\
\noalign{\vskip 1.5pt}\cdashline{2-18}\noalign{\vskip 2pt}
& TPT & 8.48 & 9.46 & 10.20 & 23.98 & 15.16 & 25.10 & 24.00 & 33.94 & 32.12 & 37.08 & 55.64 & 16.54 & 13.68 & 34.06 & 33.58 & 24.87 \\
& VTE & 9.18 & 10.76 & 10.78 & 24.72 & 14.30 & 24.36 & 25.24 & 35.38 & 32.46 & 38.16 & 55.56 & 16.14 & 14.26 & 38.72 & 33.98 & 25.60 \\
& StatA ($\gamma$=0.1) & 10.56 & 11.22 & 10.86 & 22.11 & 14.12 & 22.39 & 20.85 & 31.39 & 29.90 & 34.63 & 53.32 & 16.00 & 12.46 & 29.55 & 32.79 & 23.47 \\
& StatA ($\gamma$=-1) & 11.83 & 13.02 & 12.41 & 23.71 & 15.02 & 23.64 & 21.79 & 31.80 & 30.22 & 36.52 & 54.15 & 17.56 & 13.00 & 32.01 & 33.37 & 24.67 \\
& BATCLIP & 19.32 & 21.38 & 19.60 & 26.58 & 21.94 & 30.88 & 29.02 & 36.48 & 32.00 & 40.98 & 56.72 & 26.14 & 23.74 & 37.67 & 38.34 & 30.72 \\
& Mint & 22.50 & 24.06 & 22.70 & 28.44 & 24.24 & 32.22 & 27.96 & 37.28 & 33.60 & 42.44 & 58.92 & 27.04 & 22.10 & 41.58 & 40.98 & 32.40 \\
& ClipTTA & 18.79 & 20.23 & 18.49 & 26.54 & 21.09 & 28.36 & 26.04 & 34.66 & 32.70 & 39.54 & 54.43 & 24.01 & 18.33 & 36.84 & 35.14 & 29.01 \\
& CAMA (Ours) & \textbf{23.72} & \textbf{25.08} & \textbf{24.98} & \textbf{30.68} & \textbf{27.34} & \textbf{33.86} & \textbf{31.62} & \textbf{39.50} & \textbf{36.66} & \textbf{44.56} & \textbf{59.12} & \textbf{32.12} & \textbf{29.48} & \textbf{43.34} & \textbf{41.88} & \textbf{34.93} \\
\bottomrule
\end{tabular}
}
\end{table*}

\textbf{All Baselines on the Seven Backbones.}~~Table~\ref{tab:backbone_full} adds RoTTA, RPL, and VTE to the baselines of Table~\ref{tab:backbones} on the CIFAR benchmarks for all seven backbones. On ConvNeXt-CLIP, SAR concentrates its predictions on one class on every corruption of the three benchmarks, with a top-class share of at least $43\%$, brightness included. Its recovery scheme, which restores the source weights when the moving-average entropy falls below $0.2$, is triggered $11$ times on ImageNet-C. Mint on EVA-CLIP falls below zero-shot on all three benchmarks. On ImageNet-C, RoTTA reaches $28.09$, $31.69$, and $37.54$ on SigLIP, MetaCLIP, and EVA-CLIP, RPL reaches $28.42$, $32.24$, and $37.89$, and VTE reaches $27.73$, $34.81$, and $37.30$, all below SAR.

\begin{table*}[htbp]
\centering
\small
\caption{CAMA is the best method in all but one row against every baseline in the table. Mean online accuracy (\%) over 15 corruptions at severity 5, seed 1. Best per row in \textbf{bold}.}
\label{tab:backbone_full}
\setlength{\tabcolsep}{5pt}
\begin{tabular}{llrrrrrrrrrr}
\toprule
Benchmark & Backbone & ZS & TENT & RoTTA & RPL & SAR & VTE & BATCLIP & Mint & ClipTTA & CAMA \\
\midrule
CIFAR-10-C & CLIP ViT-B/16 & 61.16 & 62.03 & 61.89 & 61.52 & 67.37 & 64.15 & 73.85 & 73.67 & 68.23 & \textbf{79.46} \\
& SigLIP & 59.04 & 59.93 & 59.81 & 59.38 & 65.70 & 60.78 & 72.80 & 70.01 & 67.46 & \textbf{78.43} \\
& MetaCLIP & 75.94 & 84.50 & 76.70 & 83.87 & 80.14 & 78.22 & 85.17 & 83.25 & 80.58 & \textbf{87.33} \\
& EVA-CLIP & 81.12 & 81.69 & 83.35 & 76.70 & 85.11 & 79.40 & 89.87 & 67.85 & 89.59 & \textbf{90.71} \\
& ConvNeXt-CLIP & 64.16 & 41.49 & 64.63 & 36.49 & 11.87 & 64.49 & 78.00 & 65.32 & 60.61 & \textbf{78.62} \\
& RN50 & 35.08 & 12.07 & 22.32 & 11.78 & 21.24 & 36.58 & 34.85 & 24.96 & 56.36 & \textbf{58.29} \\
& RN101 & 42.47 & 12.97 & 20.78 & 12.82 & 29.29 & 41.93 & 27.94 & 24.64 & \textbf{66.50} & 64.03 \\
\midrule
CIFAR-100-C & CLIP ViT-B/16 & 35.79 & 37.96 & 36.12 & 38.47 & 41.19 & 35.01 & 42.09 & 44.53 & 40.53 & \textbf{48.77} \\
& SigLIP & 34.76 & 24.37 & 35.29 & 26.54 & 35.90 & 31.87 & 41.35 & 42.91 & 40.34 & \textbf{48.70} \\
& MetaCLIP & 48.12 & 54.54 & 48.76 & 56.40 & 54.57 & 50.23 & 55.60 & 57.26 & 53.79 & \textbf{61.70} \\
& EVA-CLIP & 58.34 & 57.34 & 60.37 & 59.19 & 63.13 & 56.78 & 61.98 & 43.25 & 67.98 & \textbf{71.01} \\
& ConvNeXt-CLIP & 38.84 & 14.18 & 39.44 & 15.47 & 2.71 & 35.49 & 47.37 & 33.87 & 33.75 & \textbf{53.11} \\
& RN50 & 15.41 & 2.05 & 4.91 & 2.06 & 6.34 & 13.41 & 15.76 & 7.50 & 26.85 & \textbf{31.75} \\
& RN101 & 20.59 & 2.62 & 6.18 & 2.49 & 8.90 & 17.98 & 15.07 & 7.27 & 34.63 & \textbf{37.18} \\
\bottomrule
\end{tabular}
\end{table*}

\textbf{Compression and Collapse on the Seven Backbones.}~~Table~\ref{tab:mechanism_backbones} gives, for each backbone on CIFAR-10-C, the mean pairwise cosine of the image features on clean and corrupted data. It splits the 15 corruptions by whether TENT collapses on them, that is, whether TENT assigns more than half of the stream to one class. The after-TENT columns come from a separate pass with the final adapted model under the protocol of the corresponding table. The other columns are read from the CAMA runs of those tables. Corruption raises the cosine above its clean value on all 15 corruptions of every backbone. Entropy minimization hurts only where compression is strong enough for the bound of Proposition~\ref{prop:sink} to take effect. On the collapsing corruptions TENT compresses the features further, and its top-class share averages $61\%$ to $96\%$. On the other corruptions its cosine falls below the corrupted value, and its share stays between $14\%$ and $34\%$ on average. CAMA brings the cosine below its clean value and the share to within $5.4$ points of the true-label share on every corruption of every backbone. Its gain over TENT follows the collapse, $41\%$ to $51\%$ on the collapsing corruptions and $3\%$ to $17\%$ on the rest. Within each backbone other than the ResNet CLIPs, the Spearman correlation between TENT's top-class share and the gain is $0.87$ to $0.99$. On the ResNet CLIPs TENT collapses on every corruption, and CAMA leads it by $32\%$ to $64\%$ on each. MetaCLIP, on which TENT never assigns more than half of the stream to one class, is the mildest case, and CAMA still has a higher mean than TENT on both CIFAR benchmarks in Table~\ref{tab:backbones}.

\begin{figure*}[t]
\centering
\includegraphics[width=0.663\textwidth]{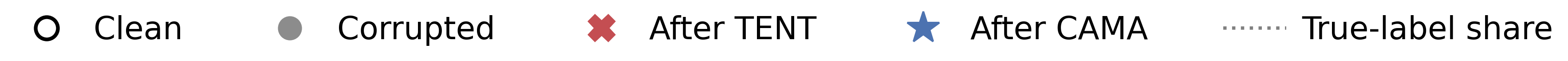}\\[1pt]
\begin{subfigure}[t]{0.26\textwidth}\centering\includegraphics[width=\linewidth]{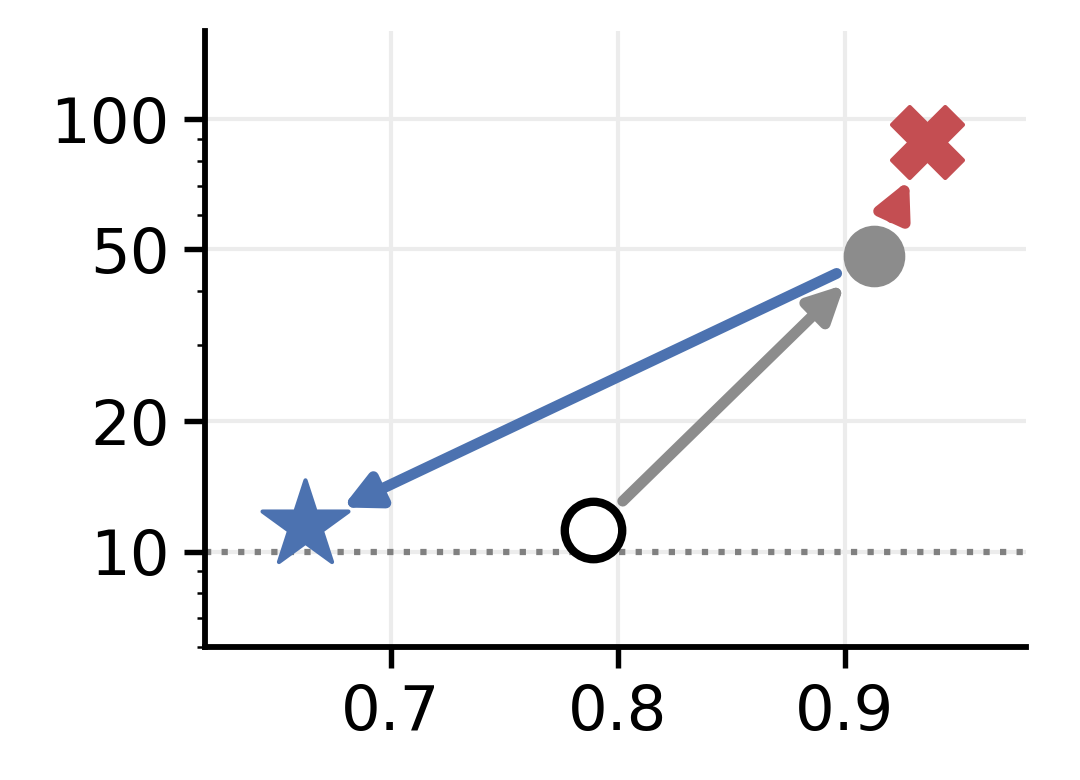}\caption{CLIP ViT-B/16}\end{subfigure}\hspace{0.03\textwidth}
\begin{subfigure}[t]{0.26\textwidth}\centering\includegraphics[width=\linewidth]{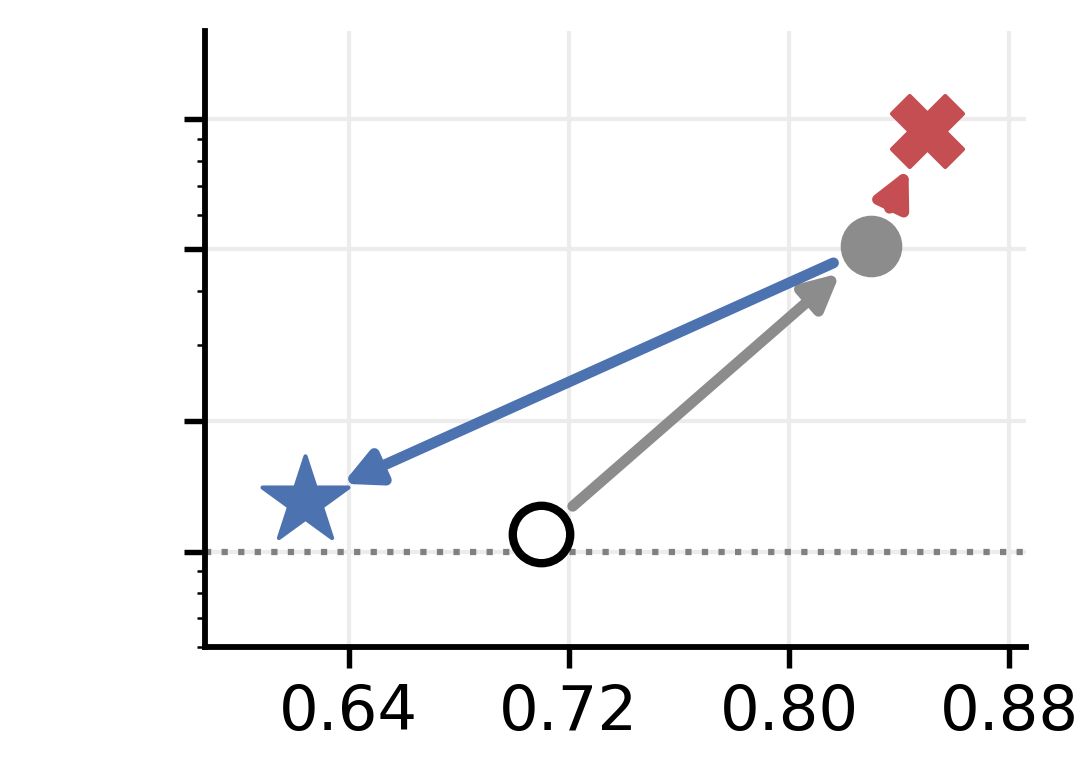}\caption{SigLIP}\end{subfigure}\hspace{0.03\textwidth}
\begin{subfigure}[t]{0.26\textwidth}\centering\includegraphics[width=\linewidth]{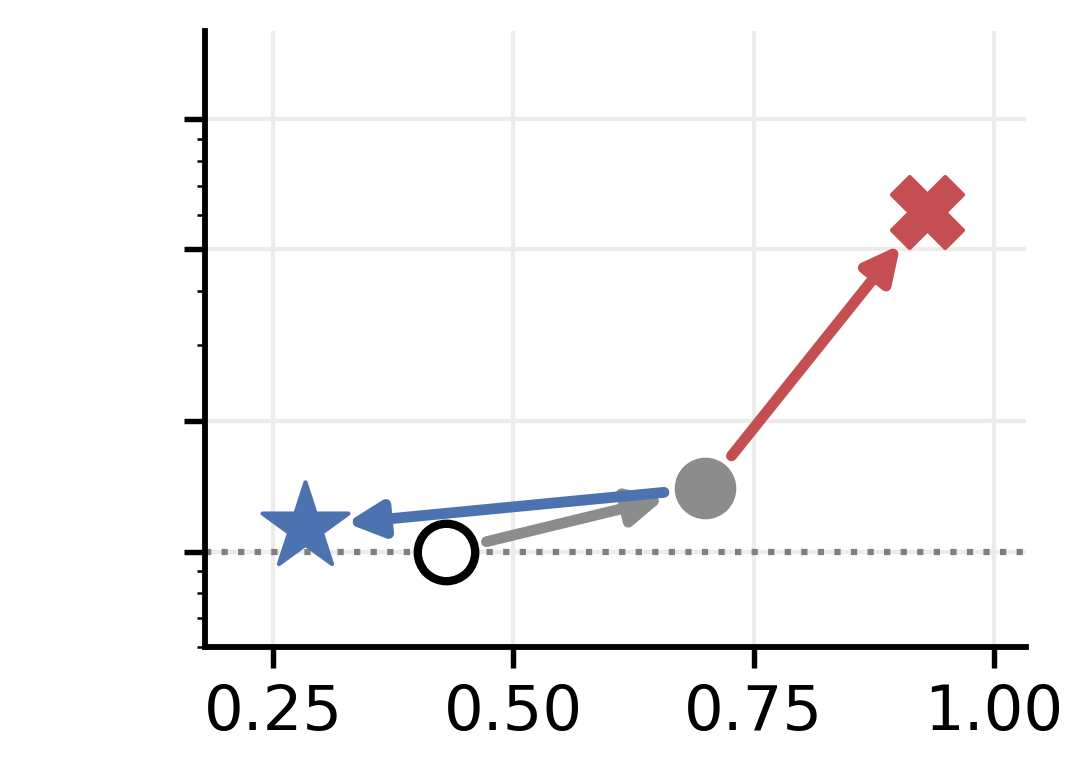}\caption{EVA-CLIP}\end{subfigure}\\[2pt]
\begin{subfigure}[t]{0.26\textwidth}\centering\includegraphics[width=\linewidth]{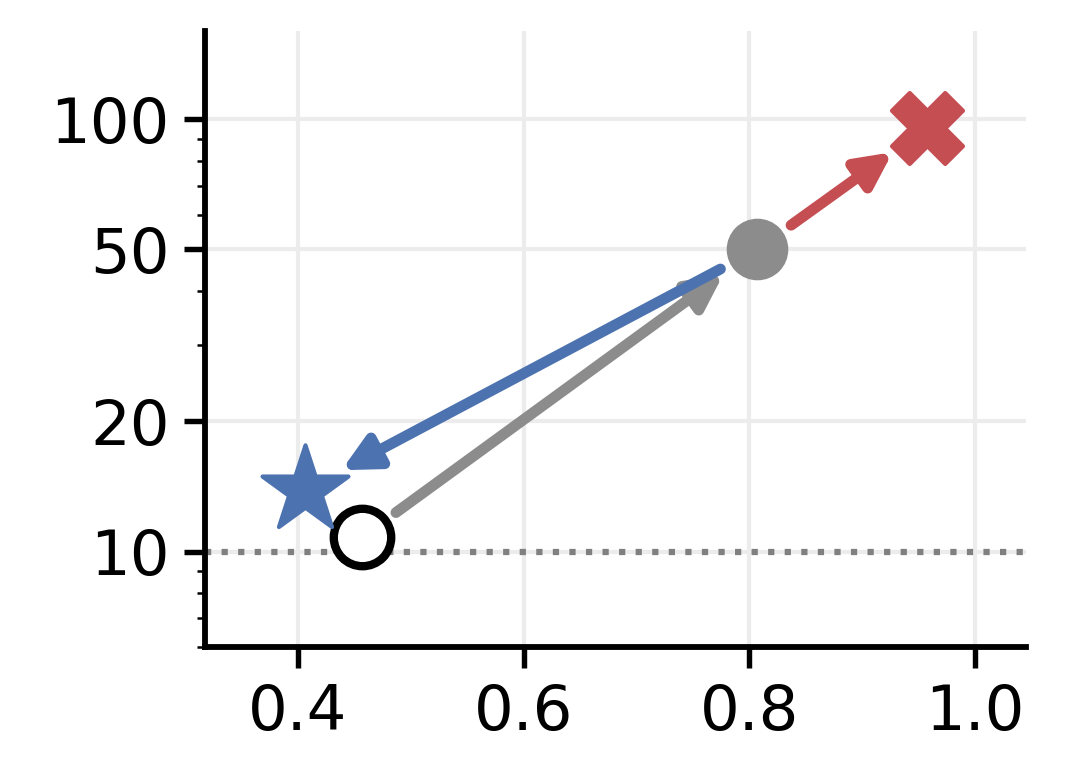}\caption{ConvNeXt-CLIP}\end{subfigure}\hspace{0.03\textwidth}
\begin{subfigure}[t]{0.26\textwidth}\centering\includegraphics[width=\linewidth]{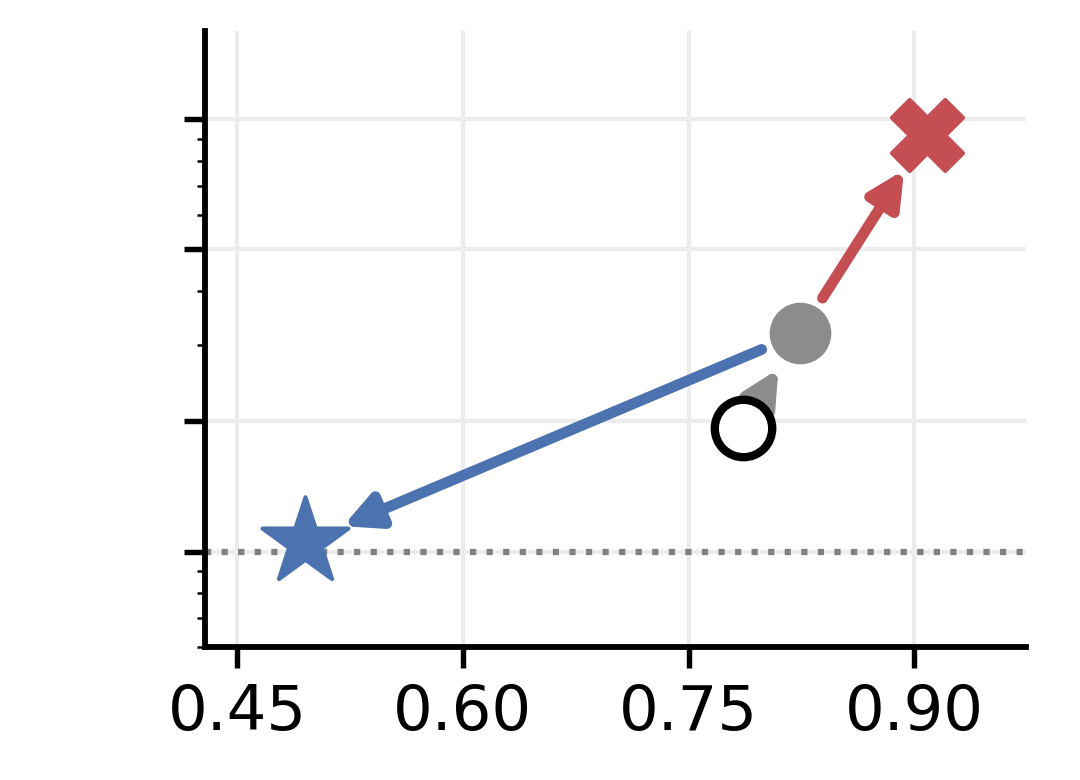}\caption{RN50}\end{subfigure}\hspace{0.03\textwidth}
\begin{subfigure}[t]{0.26\textwidth}\centering\includegraphics[width=\linewidth]{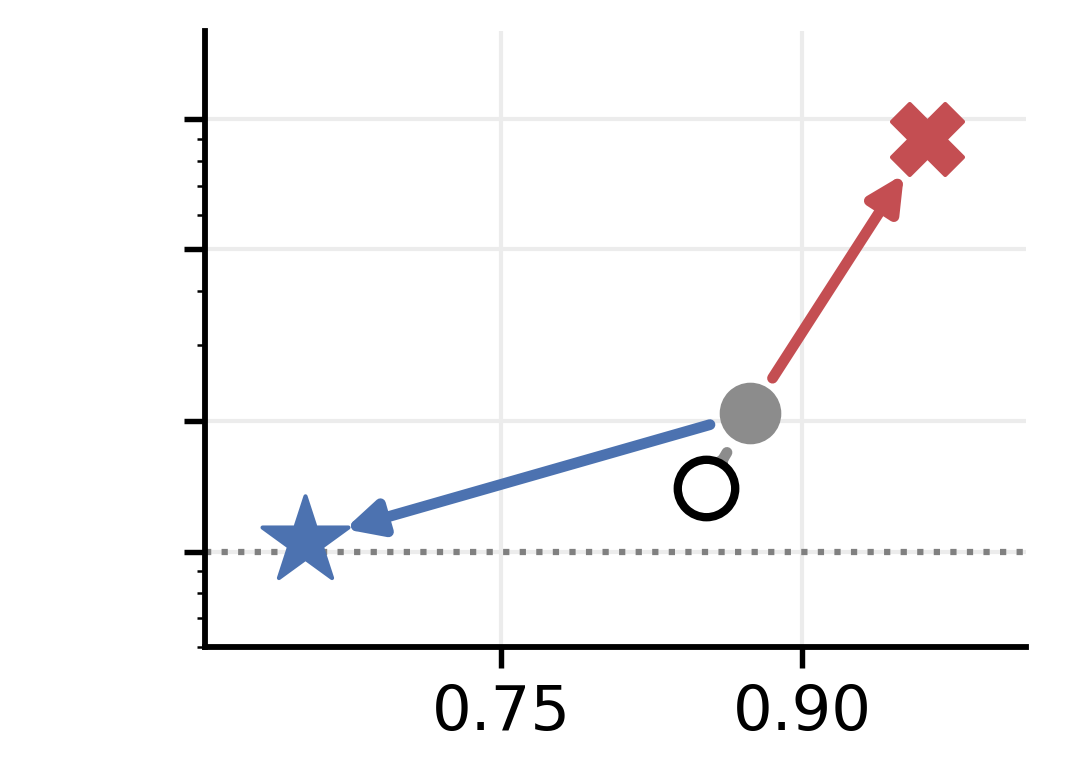}\caption{RN101}\end{subfigure}
\caption{On every backbone shown, corruption raises compression and the top-class share, TENT drives both into collapse, and CAMA reverses both. The horizontal axis is the mean pairwise cosine of the image features, and the vertical axis is the top-class share (\%) on a log scale, with the true-label share dotted. Each backbone is shown on the CIFAR-10-C corruption where TENT loses most.}
\label{fig:chain}
\end{figure*}

Corruption also raises the cosine on all 15 CIFAR-100-C corruptions of every backbone. The clean cosine is $0.791$, $0.697$, $0.690$, $0.442$, $0.459$, $0.791$, and $0.856$ for CLIP ViT-B/16, SigLIP, MetaCLIP, EVA-CLIP, ConvNeXt-CLIP, RN50, and RN101, and the corrupted mean is $0.862$, $0.805$, $0.783$, $0.621$, $0.656$, $0.875$, and $0.911$. On CIFAR-100-C, TENT collapses on 9, 3, and 12 of the 15 corruptions of SigLIP, EVA-CLIP, and ConvNeXt-CLIP. CAMA's gain over TENT there is $34.5\%$, $48.0\%$, and $43.0\%$, and $9.0\%$, $5.0\%$, and $22.6\%$ on the rest. On RN50 and RN101 TENT collapses on all 15 corruptions, and CAMA leads it by $29.7\%$ and $34.6\%$. On CIFAR-100-C and ImageNet-C as well, CAMA's top-class share stays within $4.9$ points of the true-label share on every corruption of every backbone. Figure~\ref{fig:chain} shows the chain from compression to collapse on each backbone where TENT collapses, on the CIFAR-10-C corruption where TENT loses most.

\begin{table}[htbp]
\centering
\small
\caption{On every backbone, CAMA brings the cosine below its clean value and the top-class share close to the true-label share, and it leads TENT by 41\% to 51\% where TENT collapses and by 3\% to 17\% on the other corruptions, where TENT's top-class share is still above the true-label share. CIFAR-10-C at severity 5, seed 1. The 15 corruptions are split by whether TENT collapses on them, that is, whether its top-class share exceeds one half. The columns give the number of corruptions and the group means of the pairwise cosine of the image features under corruption, after TENT, and after CAMA. The remaining columns give the group means of the top-class share (\%) after TENT and after CAMA and of CAMA's accuracy gain over TENT (\%). The clean cosine of each backbone is in parentheses, and the true-label share is $10\%$.}
\label{tab:mechanism_backbones}
\setlength{\tabcolsep}{4pt}
\begin{tabular}{ll r rrr rr r}
\toprule
& & & \multicolumn{3}{c}{Mean pairwise cosine} & \multicolumn{2}{c}{Top-class share (\%)} & \\
\cmidrule(lr){4-6} \cmidrule(lr){7-8}
Backbone (clean cosine) & Corruptions & $n$ & Corrupted & TENT & CAMA & TENT & CAMA & CAMA $-$ TENT \\
\midrule
\multirow{2}{*}{CLIP ViT-B/16 (0.789)} & TENT collapses & 4 & 0.907 & 0.925 & 0.679 & 80.1 & 12.6 & $+47.0$ \\
& others & 11 & 0.844 & 0.601 & 0.655 & 19.2 & 10.6 & $+6.6$ \\
\midrule
\multirow{2}{*}{SigLIP (0.713)} & TENT collapses & 6 & 0.868 & 0.895 & 0.641 & 76.9 & 12.7 & $+40.9$ \\
& others & 9 & 0.785 & 0.567 & 0.617 & 16.7 & 10.6 & $+3.6$ \\
\midrule
\multirow{2}{*}{MetaCLIP (0.704)} & TENT collapses & 0 & --- & --- & --- & --- & --- & --- \\
& others & 15 & 0.792 & 0.548 & 0.629 & 14.3 & 10.6 & $+2.8$ \\
\midrule
\multirow{2}{*}{EVA-CLIP (0.431)} & TENT collapses & 1 & 0.698 & 0.930 & 0.283 & 60.7 & 10.4 & $+41.9$ \\
& others & 14 & 0.611 & 0.372 & 0.299 & 17.2 & 10.4 & $+6.1$ \\
\midrule
\multirow{2}{*}{ConvNeXt-CLIP (0.457)} & TENT collapses & 11 & 0.679 & 0.891 & 0.388 & 72.8 & 11.0 & $+44.9$ \\
& others & 4 & 0.598 & 0.529 & 0.378 & 33.8 & 10.4 & $+17.1$ \\
\midrule
\multirow{2}{*}{RN50 (0.786)} & TENT collapses & 15 & 0.879 & 0.933 & 0.522 & 95.8 & 11.1 & $+46.2$ \\
& others & 0 & --- & --- & --- & --- & --- & --- \\
\midrule
\multirow{2}{*}{RN101 (0.852)} & TENT collapses & 15 & 0.912 & 0.978 & 0.660 & 94.6 & 11.0 & $+51.1$ \\
& others & 0 & --- & --- & --- & --- & --- & --- \\
\bottomrule
\end{tabular}
\end{table}

On CLIP ViT-B/16, CAMA shows no collapse on any corruption of the three benchmarks, and its top-class share over the stream stays within 4.6 points of the true-label share. On every CIFAR-10-C corruption of every backbone it stays within 5.4 points, and where TENT collapses, CAMA leads it by 41\% to 51\%. On MetaCLIP, TENT collapses on no CIFAR-10-C corruption. Still, on the five corruptions where it assigns 17\% to 23\% of the stream to one class, CAMA leads it by 5.5\% to 12.8\%. Over all three benchmarks, TENT ends below zero-shot in 11 of the 18 rows of Table~\ref{tab:backbones}. On ConvNeXt-CLIP and the ResNet CLIPs, where TENT collapses on most corruptions, it falls far below zero-shot on all three benchmarks, and SAR, which also minimizes entropy, ends below zero-shot there too.

\textbf{Reference Construction and Label Imbalance.}~~Let $z_i\in\mathbb{R}^K$ denote the logits of calibration sample $i$, and $N$ the number of calibration samples. The mean-logit and mean-prob references of Table~\ref{tab:reference} are
\begin{equation}
\pi^{\mathrm{logit}}_k = \frac{\exp(\bar z_k)}{\sum_j\exp(\bar z_j)},\quad \bar z = \frac{1}{N}\sum_i z_i ,\qquad \pi^{\mathrm{prob}}_k = \frac{1}{N}\sum_i \frac{\exp(z_{ik})}{\sum_j\exp(z_{ij})} .
\end{equation}
CAMA uses the rank-based reference in Eq.~\eqref{eq:pi_def}. In Table~\ref{tab:reference} each reference is used in both calibration and adaptation, and $\lambda$ is recomputed for each variant with the same calibration rule, so the accuracy drops of the averaged references reflect both the reference and the weight it leads to. All other settings follow Table~\ref{tab:hp}.

\textbf{Component Controls.}~~Table~\ref{tab:component_controls} removes the escort, the shrinkage, and the mutual-information term one at a time on CLIP ViT-B/16 with the protocol batch sizes. On the balanced benchmarks removing the escort leaves accuracy nearly unchanged, because the escort of a nearly uniform reference nearly equals the reference. The shrinkage has little to correct when $B \gg K$, as on CIFAR with $B = 200$, and matters when $B \ll K$. On ImageNet-C, with $B = 64$ and $K = 1000$, removing it leaves the weight at $\lambda_0$ and costs $2.7\%$. The mutual-information term is essential everywhere.

\begin{table}[htbp]
\centering
\caption{The MI term is essential on every benchmark, and the shrinkage matters on ImageNet-C, where the batch is small relative to the label set. Mean online accuracy (\%) on CLIP ViT-B/16, seed 1. Without the escort, $\bpi$ is the alignment target. Without the shrinkage, the weight is $\lambda_0$. Without MI, the alignment term is kept.}
\label{tab:component_controls}
\begin{tabular}{lrrr}
\toprule
Variant & CIFAR-10-C & CIFAR-100-C & ImageNet-C \\
\midrule
Full CAMA & 79.46 & 48.77 & 34.93 \\
Without escort & 79.52 & 48.77 & 34.96 \\
Without shrinkage & 79.48 & 48.42 & 32.26 \\
Without MI & 65.36 & 33.11 & 16.42 \\
\bottomrule
\end{tabular}
\end{table}

\textbf{Continual and Mixed Streams.}~~We follow the continual sequence of~\citet{wang2022cotta}, which concatenates the 15 corruptions at severity 5 into one stream, and calibrate CAMA once at its start. Adaptation continues without model resets, recalibration, boundary labels, or shift detection. On CIFAR-10-C, CAMA retains $78.86\%$ accuracy, against $79.46\%$ in the episodic setting, while ClipTTA reaches $75.90\%$ and Mint $63.04\%$. BATCLIP drops from $73.85\%$ to $56.65\%$ and TENT from $62.03\%$ to $10.40\%$. On CIFAR-100-C, continual accuracy is $50.68\%$ for CAMA, $46.48\%$ for ClipTTA, $32.46\%$ for Mint, $27.05\%$ for BATCLIP, and $1.49\%$ for TENT. Corruption mixing follows the mixed-domains protocol of~\citet{marsden2024universal}, which pools and shuffles all 15 corruptions at severity 5 and draws a 10,000-sample stream from the pool with the protocol seed. Severity mixing shuffles severities 1--5 within each corruption type, and joint mixing shuffles both, again into a 10,000-sample stream. All methods receive the same samples in the same order, and CAMA calibrates on the first batches of the mixed stream. CAMA leads in all six mixed settings of Table~\ref{tab:streams}, by $2.6\%$ to $6.0\%$ over the strongest baseline.

\textbf{Temporally Correlated Streams.}~~Temporal class correlation reduces the class diversity within each arriving batch, so the batch marginal no longer represents the class marginal of the stream. We follow the Dirichlet-ordered protocol of RoTTA~\citep{yuan2023rotta} at its default concentration, $\gamma = 0.1$, and at a milder one, $\gamma = 1.0$. A replay buffer adds past samples to the adaptation batch, and CAMA's objective stays unchanged. This evaluation resets the model and buffer for each corruption. The FIFO buffer holds 600 raw inputs and starts with the calibration batches. For each update, up to 200 inputs sampled uniformly from the buffer join the current batch of 200. The current inputs enter the FIFO after the update. Online accuracy is scored on the arriving samples before the update. The update thus processes up to twice as many samples as the base method. The buffer size equals the number of calibration samples and was not tuned.

Table~\ref{tab:replay_streams} compares CAMA, BATCLIP, Mint, and ClipTTA with and without the buffer over three seeds on both CIFAR benchmarks. The baselines get the same buffer through a wrapper around their update. Their buffer starts empty at each corruption, since they have no calibration batches. The buffer makes CAMA the best method at both correlation levels on both benchmarks and leaves its i.i.d.\ accuracy on CIFAR-10-C unchanged, $79.73\pm0.09\%$ against $79.48\pm0.05\%$. On CIFAR-10-C the same buffer gives the baselines $1.1\%$ to $4.7\%$, and they stay below CAMA, which gains $13.8\%$ and $11.2\%$. What a correlated stream takes away is thus the class diversity of each batch rather than anything specific to CAMA's objective. On CIFAR-100-C CAMA leads at both levels even without the buffer, and the buffer widens its lead over the strongest baseline from $4.5\%$ and $2.8\%$ to $5.1\%$ and $4.6\%$. RoTTA itself, which has a memory bank but no anti-collapse objective, stays at zero-shot level because its robust-BatchNorm component finds no BatchNorm to update in these encoders.

\begin{table}[!t]
\centering
\small
\caption{CAMA is the best method on every stream and keeps its accuracy without resets. Online accuracy (\%) on CLIP ViT-B/16 with one calibration per stream. Continual runs the 15 corruptions in sequence, and the mixed streams shuffle the corruptions, the severities, or both. Best per row in \textbf{bold}.}
\label{tab:streams}
\footnotesize
\setlength{\tabcolsep}{3pt}
\resizebox{\columnwidth}{!}{%
\begin{tabular}{ll rrrrrr}
\toprule
Stream & Benchmark & ZS & TENT & BATCLIP & Mint & ClipTTA & CAMA \\
\midrule
\multirow{2}{*}{Episodic} & CIFAR-10-C & 61.16 & 62.03 & 73.85 & 73.67 & 68.23 & \textbf{79.46} \\
 & CIFAR-100-C & 35.79 & 37.96 & 42.09 & 44.53 & 40.53 & \textbf{48.77} \\
\midrule
\multirow{2}{*}{Continual} & CIFAR-10-C & 61.16 & 10.40 & 56.65 & 63.04 & 75.90 & \textbf{78.86} \\
 & CIFAR-100-C & 35.79 & 1.49 & 27.05 & 32.46 & 46.48 & \textbf{50.68} \\
\midrule
\multirow{2}{*}{Mixed corr.} & CIFAR-10-C & 60.96 & 58.95 & 68.09 & 67.63 & 64.95 & \textbf{74.11} \\
 & CIFAR-100-C & 35.77 & 37.66 & 37.97 & 37.75 & 38.23 & \textbf{42.71} \\
\midrule
\multirow{2}{*}{Mixed sev.} & CIFAR-10-C & 73.69 & 75.01 & 82.22 & 81.38 & 78.67 & \textbf{85.28} \\
 & CIFAR-100-C & 48.36 & 51.32 & 52.89 & 55.22 & 52.31 & \textbf{57.82} \\
\midrule
\multirow{2}{*}{Mixed both} & CIFAR-10-C & 74.18 & 79.69 & 80.07 & 77.73 & 77.25 & \textbf{83.30} \\
 & CIFAR-100-C & 48.65 & 51.71 & 50.97 & 50.80 & 51.06 & \textbf{55.01} \\
\bottomrule
\end{tabular}}
\end{table}

\begin{table}[htbp]
\centering
\caption{With a replay buffer, CAMA is the best method at both correlation levels on both benchmarks. Online accuracy (\%) on temporally correlated CIFAR-10-C and CIFAR-100-C streams with CLIP ViT-B/16, mean $\pm$ std over three seeds, averaged over 15 corruptions at severity 5. A smaller Dirichlet concentration $\gamma$ gives stronger class correlation.}
\label{tab:replay_streams}
\setlength{\tabcolsep}{3pt}
\resizebox{\columnwidth}{!}{%
\begin{tabular}{l rr @{\quad} rr}
\toprule
& \multicolumn{2}{c}{CIFAR-10-C} & \multicolumn{2}{c}{CIFAR-100-C} \\
\cmidrule(lr){2-3}\cmidrule(l){4-5}
Method & Mild ($\gamma=1.0$) & Strong ($\gamma=0.1$) & Mild ($\gamma=1.0$) & Strong ($\gamma=0.1$) \\
\midrule
BATCLIP & $69.96\pm1.06$ & $62.93\pm2.53$ & $42.07\pm0.07$ & $41.87\pm0.24$ \\
BATCLIP + replay & $72.05\pm1.03$ & $64.06\pm2.17$ & $42.79\pm0.04$ & $42.56\pm0.26$ \\
Mint & $69.71\pm0.54$ & $64.48\pm2.99$ & $44.21\pm0.07$ & $43.88\pm0.19$ \\
Mint + replay & $70.85\pm0.14$ & $65.84\pm3.58$ & $43.79\pm0.09$ & $43.59\pm0.09$ \\
ClipTTA & $62.03\pm1.07$ & $59.48\pm0.22$ & $41.04\pm0.06$ & $40.80\pm0.08$ \\
ClipTTA + replay & $66.69\pm0.42$ & $62.25\pm0.23$ & $41.18\pm0.08$ & $41.07\pm0.06$ \\
CAMA & $60.83\pm1.64$ & $56.01\pm0.77$ & $48.67\pm0.09$ & $46.64\pm0.22$ \\
CAMA + replay & $\mathbf{74.60}\pm0.49$ & $\mathbf{67.16}\pm0.91$ & $\mathbf{49.32}\pm0.15$ & $\mathbf{48.53}\pm0.22$ \\
\bottomrule
\end{tabular}}
\end{table}

When the classes are temporally correlated, the batch marginal no longer represents the class marginal of the stream. The replay buffer holds 600 past inputs, from which up to 200 join each batch, and it leaves the accuracy on i.i.d.\ streams unchanged. Under correlation, what is missing is therefore memory rather than the objective.

\section{Supervised Classifiers}\label{app:regime}

The analysis of Section~\ref{sec:mechanism} concerns contrastive vision--language encoders with class text features. To test whether the mechanism and the objective extend beyond this model class, we apply CAMA unchanged to supervised classifiers with a linear head. CAMA adapts their LayerNorm affine parameters and calibrates $\lambda$ per corruption as usual. For both TENT and CAMA, the LayerNorm modules of the ConvNeXt implementation are found by their interface rather than by their class, since they do not subclass the standard module. BATCLIP is not applicable without a text encoder. Table~\ref{tab:supervised} reports CIFAR-10-C with a ConvNeXt-Base and a ViT-B/16 fine-tuned on CIFAR-10, and CIFAR-100-C with a ViT-B/16 fine-tuned on CIFAR-100, all at severity 5 with $B = 200$ and seed 1. Here too corruption concentrates the predictions. On CIFAR-10-C the top-class share rises from $10\%$ on clean data to $25\%$ for ConvNeXt-Base and $16\%$ for ViT-B/16 under corruption. After TENT it is $20\%$ on both models, and CAMA brings it back to $11\%$. Across the three settings of Table~\ref{tab:supervised}, CAMA gains 7\% to 10\% over TENT.

\begin{table}[htbp]
\centering
\caption{CAMA also improves supervised classifiers without a text encoder and leads TENT on every backbone. Mean online accuracy (\%) over 15 corruptions at severity 5, seed 1. Source is the unadapted model.}
\label{tab:supervised}
\begin{tabular}{llrrr}
\toprule
Benchmark & Backbone & Source & TENT & CAMA \\
\midrule
CIFAR-10-C & ConvNeXt-Base & 69.11 & 81.85 & \textbf{88.96} \\
CIFAR-10-C & ViT-B/16 & 80.97 & 82.81 & \textbf{90.48} \\
CIFAR-100-C & ViT-B/16 & 63.94 & 65.94 & \textbf{75.84} \\
\bottomrule
\end{tabular}
\end{table}

\end{document}